\documentclass{article}

\usepackage{arxiv}

\usepackage[utf8]{inputenc}
\usepackage[T1]{fontenc}
\usepackage{microtype}

\usepackage{amsmath,amssymb,amsthm}

\usepackage{graphicx}
\usepackage{subcaption}
\usepackage{booktabs}
\usepackage{multirow}
\usepackage{tabularx}

\usepackage[numbers,sort&compress]{natbib}

\usepackage[colorlinks=true,linkcolor=blue,citecolor=blue,urlcolor=blue]{hyperref}
\usepackage{url}

\usepackage{tcolorbox}
\newtheorem{definition}{Definition}

\renewcommand{\headeright}{Preprint}
\renewcommand{\undertitle}{Preprint}
\renewcommand{\shorttitle}{Knowledge Activation}

\hypersetup{
  pdftitle={Knowledge Activation: AI Skills as the Institutional Knowledge Primitive for Agentic Software Development},
  pdfauthor={Gal Bakal},
  pdfkeywords={knowledge management, AI agents, skills, platform engineering, context window, enterprise software},
}

\title{Knowledge Activation: AI Skills as the Institutional Knowledge Primitive\\
for Agentic Software Development}
\author{%
  Gal Bakal\\
  Yahoo Inc.\thanks{The views expressed are the author's own and do not represent an official position of Yahoo Inc.}\\
  \texttt{gal.bakal@yahooinc.com}
}
\date{Submitted 16 March 2026; revised 4 June 2026}

\begin{document}

\maketitle

\begin{abstract}
Enterprise software organizations accumulate critical institutional knowledge---architectural decisions, deployment procedures, compliance policies, incident playbooks---yet this knowledge remains trapped in formats designed for human interpretation. The bottleneck to effective agentic software development is not model capability but knowledge architecture. When any knowledge consumer---an autonomous AI agent, a newly onboarded engineer, or a senior developer navigating an unfamiliar codebase---encounters an enterprise task without institutional context, the result is the same: guesswork, correction cascades, and a disproportionate tax on the senior engineers who must manually supply what others cannot infer.

This paper introduces \emph{Knowledge Activation}, a framework that addresses this bottleneck by specializing AI Skills---the open standard for agent-consumable knowledge---into structured, governance-aware \emph{Atomic Knowledge Units} (AKUs) designed for institutional knowledge delivery. Rather than retrieving documents for interpretation, AKUs deliver action-ready specifications---encoding what to do, which tools to use, what constraints to respect, and where to go next---so that agents act correctly and the engineers working with them receive institutionally grounded guidance without reconstructing organizational context from scratch.

AKUs are designed not as isolated artifacts but as interconnected nodes. Each AKU declares its relationships to others, forming a composable knowledge graph that agents traverse at runtime---compressing onboarding, reducing cross-team friction, and eliminating the correction cascades that currently degrade both agent performance and developer productivity. The paper formalizes the resource constraints that make this architecture necessary, specifies the AKU schema and deployment architecture, and grounds long-term maintenance in knowledge commons practice. We additionally report an enterprise deployment of the framework at Yahoo, where an anonymous survey of 67 engineers establishes statistically significant developer-experience improvements (large to very large effect sizes on all four perceived-experience drivers, a mean of 2.6 hours per week saved per engineer, a Net Promoter Score of $+35$, and no significant difference detected across business units) consistent with the framework's structural claims. For practitioners, the core implication is direct: the organizations that architect their institutional knowledge for the agentic era will outperform those that invest solely in model capability.
\end{abstract}

\section{Introduction}
\label{sec:introduction}

\subsection{The Agentic Shift in Software Engineering}

The role of artificial intelligence in software engineering is undergoing a fundamental
transformation. Where earlier generations of AI tooling served as passive assistants---offering
code completions, suggesting refactorings, or answering developer queries---a new class
of autonomous AI agents is emerging that can independently navigate codebases, formulate
multi-step plans, execute tool-mediated actions, and resolve complex engineering tasks with
minimal human oversight. The SWE-bench benchmark demonstrated that language model agents
can resolve real-world GitHub issues drawn from popular open-source repositories, establishing
a rigorous evaluation framework for autonomous software engineering
capability~\cite{jimenez2024swebench}. Building on this foundation, SWE-agent introduced
agent-computer interfaces specifically designed to enable language models to interact with
software repositories, achieving substantial improvements in autonomous issue
resolution~\cite{yang2024sweagent}. Complementary platforms such as
OpenHands~\cite{wang2024openhands} and multi-agent frameworks including
AutoGen~\cite{wu2023autogen} and MetaGPT~\cite{hong2023metagpt} further illustrate the
rapid maturation of agent architectures capable of sustained, goal-directed software
engineering work. Empirical analysis of GitHub activity confirms that this shift is
already underway at scale: coding agents are increasingly visible as contributors
across public repositories, moving from research prototypes to production
tools~\cite{robbes2026agentic}.

These advances rest on a broader foundation of research into language model reasoning and
tool use. Chain-of-thought prompting demonstrated that large language models can perform
complex multi-step reasoning when guided by appropriate intermediate
steps~\cite{wei2022chain}. The ReAct paradigm unified reasoning traces with action
execution, enabling agents to interleave deliberation with environment
interaction~\cite{yao2023react}. Toolformer showed that language models can learn to
invoke external tools autonomously, expanding their effective capability
set~\cite{schick2023toolformer}. Reflexion introduced verbal reinforcement learning,
allowing agents to improve through self-reflection on prior
attempts~\cite{shinn2023reflexion}. Collectively, these developments have moved the field
from language models as text generators to language models as autonomous actors embedded
in computational environments. Hassan et al.~\cite{hassan2025agentic} formalize this
trajectory as \emph{agentic software engineering}---an emerging research frontier
encompassing agent architectures, evaluation frameworks, and the organizational
infrastructure required to deploy agents effectively in practice.

\subsection{The Institutional Knowledge Problem}

Yet a critical gap separates the demonstrated capabilities of AI agents on open-source
benchmarks from the requirements of enterprise software development. In the open-source
setting, the relevant knowledge is largely available to the agent through two
complementary channels: the repository itself---where codebases are self-documenting
by convention, contribution guidelines are explicit, and the problem scope is bounded
by the issue description---and the agent's parametric knowledge, which is trained
predominantly on open-source code and thus already encodes the conventions, APIs, and
idioms of the open-source ecosystem. Enterprise software development
operates under fundamentally different conditions. Organizations accumulate decades of
institutional knowledge---architectural decisions and their rationales, deployment
procedures tailored to specific infrastructure configurations, compliance policies
mandated by regulatory frameworks, incident response playbooks refined through operational
experience, and team-specific conventions that exist only as tribal
knowledge~\cite{davenport1998, argote2000knowledge}.

This institutional knowledge is precisely what distinguishes a productive engineer who has
been embedded in an organization for years from a technically skilled newcomer---and the
difficulty of acquiring it is well documented. Cohen and Levinthal's theory of
\emph{absorptive capacity}~\cite{cohen1990absorptive} established that an organization's
ability to assimilate new knowledge depends on prior related knowledge; a newcomer lacking
organizational context faces a structural deficit that no amount of technical skill can
compensate for. Szulanski~\cite{szulanski1996stickiness} demonstrated that internal
knowledge transfer is impeded not primarily by motivational factors but by the
\emph{stickiness} of knowledge itself---causal ambiguity, lack of absorptive capacity in
the recipient, and arduous relationships between knowledge source and
recipient. Empirical studies of software developer onboarding confirm the pattern: the
core barrier is not technical setup but the transfer of tacit
knowledge~\cite{ju2021onboarding, balali2024newcomer}.

Polanyi's foundational distinction between tacit and explicit
knowledge~\cite{polanyi1966} frames the underlying cause: much of what makes an
enterprise developer effective is tacit, embedded in practices and judgment rather than
documented in any retrievable artifact. Nonaka and Takeuchi's knowledge creation
theory~\cite{nonaka1995} further emphasizes that organizational knowledge exists in a
dynamic cycle between tacit and explicit forms, and that the value of knowledge management
lies in enabling the conversion between these states. Alavi and Leidner's comprehensive
review of knowledge management systems~\cite{alavi2001} established that the central
challenge is not the storage of knowledge but its transfer and application---the very
problem that both new engineers and AI agents face, at human and machine scale
respectively.

This transfer problem manifests in AI agents as what this paper terms the
\emph{Institutional Impedance Mismatch}: the structural disconnect between a
knowledge consumer's existing knowledge and the specific institutional patterns of
the organization it serves. From the deployment side, we refer to this same
dynamic as the \emph{last-mile problem of agentic software development}:
foundation models arrive at the enterprise boundary having mastered the
general case---public APIs, open-source idioms, canonical patterns, broadly
documented frameworks---but with no exposure to the institutional patterns
and architecture of any specific organization, and no amount of additional
pretraining on public corpora will close a gap that is by definition
organization-private. The two terms describe the same phenomenon from
different angles---the former emphasizing the structural mismatch a
knowledge consumer encounters, the latter the delivery gap a practitioner
must close---and we use them interchangeably in what follows. When an agent
encounters an enterprise task, it attempts to act on the basis of its
generic training
knowledge---but this knowledge does not encode the organization's particular naming
conventions, architectural guardrails, deployment constraints, or compliance
requirements. The agent's initial attempt violates an organization-specific convention;
the user corrects the error; the agent adjusts but encounters another unknown
constraint; a further correction follows. Each iteration consumes context tokens with
failed attempts, error descriptions, and correction instructions. Recent empirical
research has identified this progressive accumulation of irrelevant and redundant content
as \emph{context rot}---the measurable degradation of LLM performance as the context
window fills~\cite{chromaresearch2025contextrot}. The Institutional Impedance Mismatch
thus drives a failure cascade that systematically erodes the agent's effective reasoning
capacity through mismatch-induced context rot. Section~\ref{sec:context-window-economy}
formalizes this dynamic as the \emph{productivity paradox} and identifies its
sociotechnical consequences, including what this paper terms the \emph{institutional
knowledge tax}---the disproportionate overhead imposed on senior engineers who must
manually supply the organizational context that agents cannot infer. The parallel to human onboarding is structural, not
metaphorical: a new engineer entering an unfamiliar codebase undergoes the same
cycle---guessing at conventions, violating unstated norms, consuming senior engineers'
time with corrections---over weeks and months rather than context window
turns~\cite{carlile2004transferring}. Knowledge Activation addresses the paradox for both
classes of knowledge consumer by closing the impedance mismatch at the point of knowledge
architecture rather than relying on runtime correction.

\begin{figure}[htbp]
  \centering
  \includegraphics[width=\textwidth]{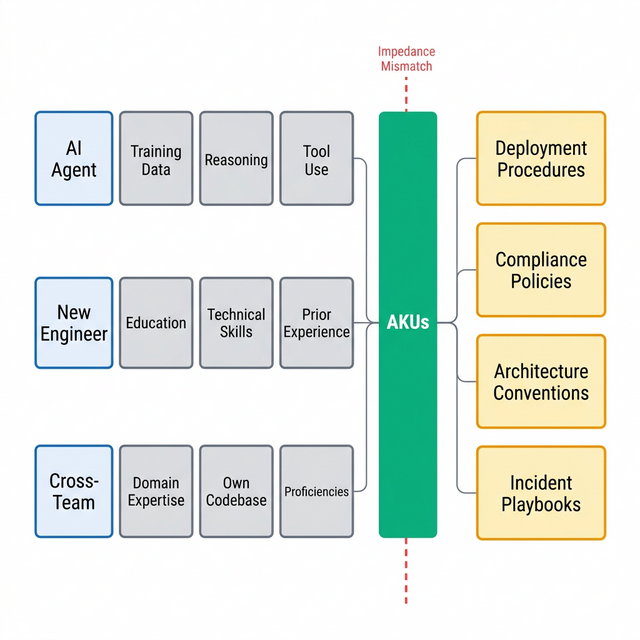}
  \caption{The Institutional Impedance Mismatch. A structural disconnect separates what knowledge consumers bring to an enterprise task (left) from what the organization's institutional knowledge contains (right). Skills (center) bridge the gap by delivering structured, governance-aware institutional knowledge at the point of need---equipping agents to act with institutional accuracy and enabling the engineers working with them to receive organizationally grounded guidance.}
  \label{fig:impedance-mismatch}
\end{figure}

The difficulty is further compounded by a fundamental architectural constraint of large language
models: the finite context window. While context window capacities have grown
substantially since the original Transformer
architecture~\cite{vaswani2017}---from thousands to millions of tokens---they
remain bounded. Every token devoted to institutional knowledge is a token unavailable for
reasoning, code analysis, or action planning. Research on prompt compression has shown
that careful management of context contents can dramatically affect model
performance~\cite{jiang2023longllmlingua}, and studies comparing retrieval with long
context have demonstrated that simply expanding the window does not eliminate the need for
deliberate content selection~\cite{xu2024retrieval}. Even the finding that a small number
of carefully curated examples can match the performance of large training
datasets~\cite{zhou2023lima} underscores a deeper principle: what occupies the context
window matters more than how large the context window is.

\subsection{The Inadequacy of Current Approaches}

Existing approaches to equipping AI agents with institutional knowledge fall into three
broad categories, none of which offers a principled solution. Retrieval-Augmented
Generation (RAG)~\cite{lewis2020rag} represents the most widely adopted strategy, whereby
relevant documents are retrieved from a knowledge base and injected into the model's
context at inference time. While effective for question-answering tasks, RAG treats
knowledge as undifferentiated text fragments optimized for semantic similarity rather than
operational relevance. Section~\ref{sec:knowledge-activation} develops this contrast in
detail.

Prompt engineering and custom system instructions represent a second approach, in which
institutional knowledge is manually encoded into static prompts or instruction
sets~\cite{brown2020language}. This strategy conflates knowledge authoring with prompt
construction, creating brittle artifacts that resist versioning, lack composability, and
cannot adapt to the dynamic requirements of multi-step agent workflows. As the scope of
required knowledge grows, these monolithic instructions inevitably exceed the context
budget, forcing practitioners into ad hoc prioritization with no formal framework to
guide the trade-offs. Recent empirical work sharpens this concern. Liu et
al.~\cite{liu2026agentsmd} found that repository-level context files (AGENTS.md) can
reduce agent runtime by 28\% and token consumption by 16\%, demonstrating that
structured context delivery has measurable value.\footnote{Several empirical findings cited in this paper---including~\cite{liu2026agentsmd, gloaguen2026evaluating, vasilopoulos2026codified, ulanuulu2026codified, zhang2026ace}---are drawn from recent arXiv preprints that have not yet undergone formal peer review. We cite them as the best available evidence in a rapidly evolving field, while acknowledging their provisional status.} Yet Gloaguen et
al.~\cite{gloaguen2026evaluating} found that na\"ive context files---both LLM-generated
and developer-written---actually \emph{reduce} task success rates while increasing
inference cost by over 20\%. A large-scale empirical study of 2{,}303 agent context files
across 1{,}925 repositories confirms the pattern: these files evolve like configuration
code rather than documentation, yet governance and security constraints are specified in
only 14.5\% of cases~\cite{chatllatanagulchai2025agentreadmes,
chatllatanagulchai2025manifests}. The productive tension between these findings---context
helps efficiency but hurts quality when unstructured---is precisely the gap that Knowledge
Activation addresses: the problem is not whether to deliver institutional knowledge to
agents, but \emph{how to architect it}.

A third category encompasses tool-use frameworks and API
integrations~\cite{patil2023gorilla, qin2023toolllm, shen2024hugginggpt}, including
emerging standards such as the Model Context
Protocol~\cite{anthropic2024mcp}. These approaches excel at extending agent capabilities
through external tool invocation but do not address the knowledge dimension: they specify
\emph{what tools exist} and \emph{how to call them}, but not \emph{when to use them},
\emph{why one tool is preferred over another in a given organizational context}, or
\emph{what governance constraints apply}. Tool registries are necessary but not
sufficient; without the institutional knowledge that contextualizes tool use, agents
cannot make the judgment calls that enterprise environments demand.

Task decomposition strategies~\cite{khot2023decomposed, zhou2023leasttomostprompting}
address the complexity of multi-step reasoning but similarly lack a knowledge architecture
layer. They assume that the agent already possesses or can retrieve the knowledge needed
for each subtask, offering no framework for how that knowledge should be structured,
scoped, or governed.

\subsection{Thesis and Contributions}

The bottleneck to effective agentic software development is not model capability but
knowledge architecture. This paper introduces the \emph{Knowledge Activation} framework,
built on the thesis that the central challenge facing both autonomous AI agents and human
engineers in enterprise software development is the absence of a deliberate knowledge
architecture---one that structures institutional knowledge into forms that serve any
knowledge consumer operating under constraints, whether the finite context window of an
LLM or the finite absorptive capacity~\cite{cohen1990absorptive} of a newly onboarded
engineer. The framework centers on the \emph{skill}---a structured, governance-aware
knowledge primitive that encapsulates institutional knowledge into composable,
action-oriented units bridging Internal Developer
Platforms~\cite{beetz2023platform, gartner2024platform, cncf2024platforms}, agent
capabilities, developer onboarding, and governance requirements.

The framework makes four interconnected contributions. First, it provides a
\emph{problem formalization}: the \emph{Context Window Economy}
(Section~\ref{sec:context-window-economy}) models the constraints under which knowledge
must be delivered to agents, drawing on information
theory~\cite{shannon1948mathematical} and cognitive load
theory~\cite{sweller1988cognitive}; the \emph{Institutional Impedance Mismatch}
(defined above) names the structural disconnect between parametric model
knowledge and organizational institutional knowledge; and the \emph{institutional
knowledge tax} identifies the sociotechnical cost when this mismatch goes unaddressed.
Second, it specifies a \emph{knowledge architecture}: the \emph{Knowledge Activation}
pipeline (Section~\ref{sec:knowledge-activation}) transforms latent organizational
knowledge through codification, compression, and injection into \emph{Atomic Knowledge
Units}---skills---whose seven-component schema
(Section~\ref{sec:atomic-knowledge-units}) bundles intent, procedure, tools, metadata,
governance, continuations, and validators into composable, governance-aware primitives.
Third, it defines a \emph{deployment and governance model}: the \emph{Agent Knowledge
Architecture} (Section~\ref{sec:golden-path-architecture}) provides a three-layer
structure---AKU Registry, Knowledge Topology, and Activation
Policy---through which organizations deploy and govern AKUs at scale;
\emph{AI-Generated Golden Paths} reconceive curated developer
pathways~\cite{skelton2019team, puppet2023state} as workflows dynamically composed by
agents at runtime; \emph{validators} enable governance-as-code; and a \emph{knowledge
commons} model (Section~\ref{sec:discussion}) grounds sustainable skill maintenance in
community practice~\cite{wenger1998communities, hessostrom2007knowledge}. Fourth, it
provides \emph{empirical validation}: Section~\ref{sec:case-study} reports the deployment
of the framework at Yahoo as a corpus of 87 modular agent-consumable skills, with an
anonymous survey of 67 engineers establishing statistically significant
developer-experience improvements consistent with the framework's structural claims
(large to very large effect sizes on all four perceived-experience drivers, a mean of
2.6 hours per week saved per engineer, NPS $+35$, and no significant difference
detected across business units or tenure strata). Together, these
contributions bridge four domains that have developed largely in isolation: knowledge
management theory, platform engineering, autonomous AI agent design, and developer
experience research~\cite{noda2023devex, forsgren2021space}.

\begin{figure}[htbp]
  \centering
  \includegraphics[width=\textwidth]{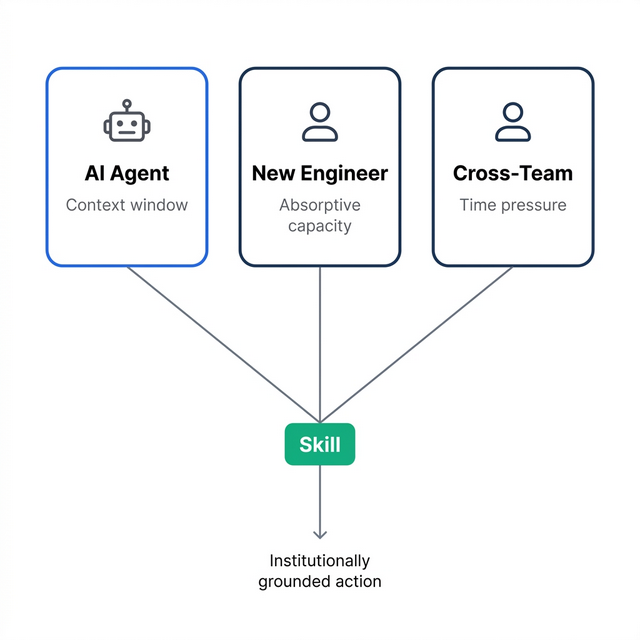}
  \caption{The knowledge consumer spectrum. Three classes of knowledge consumer face the same structural deficit---the Institutional Impedance Mismatch---under different constraints: AI agents are bounded by the finite context window, newly onboarded engineers by limited absorptive capacity for institutional knowledge, and cross-team engineers by the time pressure of operating on unfamiliar codebases. Skills serve all three through the same agent-mediated mechanism, delivering institutionally grounded guidance at the point of need.}
  \label{fig:knowledge-consumer-spectrum}
\end{figure}

\subsection{Paper Organization}

The remainder of this paper is organized as follows.
Section~\ref{sec:related-work} reviews related work across knowledge management, agent
architectures, platform engineering, and context optimization.
Section~\ref{sec:context-window-economy} develops the Context Window Economy as a
theoretical lens for understanding knowledge constraints in agentic systems.
Section~\ref{sec:knowledge-activation} defines Knowledge Activation and distinguishes it
from existing retrieval-based approaches.
Section~\ref{sec:atomic-knowledge-units} formalizes the Atomic Knowledge Unit and derives
the Skill primitive.
Section~\ref{sec:golden-path-architecture} presents Agent Knowledge Architecture as a
structural framework for organizing skills within enterprise platforms.
Section~\ref{sec:governance} addresses governance, auditability, and compliance
considerations for agent-executed knowledge.
Section~\ref{sec:case-study} reports an enterprise deployment of the framework at Yahoo
and an empirical evaluation through a survey of 67 engineers, establishing
statistically significant effects on developer experience and surfacing operational
learnings from the deployment.
Section~\ref{sec:discussion} discusses implications for platform engineering practice,
organizational knowledge strategy, developer experience and
productivity~\cite{greiler2022developer, noda2023devex}, and the sociotechnical
conditions for sustainable skill maintenance, drawing on knowledge commons
theory~\cite{hessostrom2007knowledge} and established frameworks for balancing exploration
and exploitation in organizational learning~\cite{march1991exploration}.
Section~\ref{sec:conclusion} concludes with a summary of contributions and directions for
future research.

\begin{table}[ht]
\centering
\caption{Key terms introduced in this paper.}
\label{tab:terminology}
\begin{tabularx}{\textwidth}{lX}
\toprule
\textbf{Term} & \textbf{Definition (Section)} \\
\midrule
Knowledge Consumer & Any entity---autonomous AI agent, newly onboarded engineer, or cross-team contributor---that must acquire institutional knowledge to act effectively in an enterprise context (\S\ref{sec:introduction}) \\
Context Window Economy & The regime in which knowledge delivery to agents is governed by token budget, attention decay, and latency cost constraints (\S\ref{sec:context-window-economy}) \\
Institutional Impedance Mismatch & The structural disconnect between a knowledge consumer's existing knowledge and an organization's specific institutional patterns (\S\ref{sec:introduction}) \\
Last-Mile Problem of Agentic Software Development & The deployment-side framing of the Institutional Impedance Mismatch: the gap between general foundation-model capability and organization-native agent behavior, closable only by an institutional knowledge architecture (\S\ref{sec:introduction}, \S\ref{sec:case-study}) \\
Knowledge Activation & The three-stage process (codification, compression, injection) transforming latent institutional knowledge into structured, agent-executable form (\S\ref{sec:knowledge-activation}) \\
Atomic Knowledge Unit (AKU) & The minimal self-contained bundle of intent, procedure, tools, metadata, governance, continuations, and validators enabling one coherent action (\S\ref{sec:atomic-knowledge-units}) \\
AI Skill & A portable, agent-consumable knowledge package as defined by the Agent Skills open standard; the AKU specializes this primitive for institutional knowledge delivery (\S\ref{sec:atomic-knowledge-units}) \\
Validator & A deterministic script embedded in a skill that verifies organizational standards without human approval (\S\ref{sec:atomic-knowledge-units}) \\
AI-Generated Golden Path & A workflow dynamically composed by an agent at runtime from the Knowledge Topology (\S\ref{sec:golden-path-architecture}) \\
Governance Gradient & The spectrum from fully autonomous to human-controlled operation, driven by validator coverage (\S\ref{sec:governance}) \\
Knowledge Density ($\rho$) & The ratio of actionable information to tokens consumed by a knowledge artifact (\S\ref{sec:knowledge-activation}) \\
Institutional Knowledge Tax & The overhead imposed on senior engineers who must manually supply organizational context that other knowledge consumers cannot infer (\S\ref{sec:context-window-economy}) \\
Productivity Paradox & The dynamic in which AI agents deployed without institutional knowledge degrade rather than improve organizational productivity (\S\ref{sec:context-window-economy}) \\
\bottomrule
\end{tabularx}
\end{table}

\section{Related Work}
\label{sec:related-work}

This section surveys five bodies of literature that converge on the problem addressed in this paper: encoding organizational knowledge so that autonomous AI agents can act on it reliably---and, through agent-mediated delivery, so that the human engineers who work with those agents receive institutionally grounded guidance---within enterprise software development.

\subsection{Knowledge Management in Organizations}
\label{sec:rw-km}

The foundational distinction between tacit and explicit knowledge, introduced by \cite{polanyi1966} and operationalized by \cite{nonaka1995} through the SECI (Socialization--Externalization--Combination--Internalization) model, remains central to organizational knowledge management (KM). Nonaka and Takeuchi demonstrated that competitive advantage arises from an organization's ability to convert tacit knowledge---embodied in individual expertise, intuition, and craft---into explicit, transferable artifacts. \cite{davenport1998} extended this perspective to the enterprise, framing knowledge as a strategic asset that must be actively managed through repositories, processes, and cultural incentives. \cite{argote2000knowledge} further established that knowledge transfer across organizational boundaries is a primary basis for sustained competitive advantage, yet such transfer is notoriously difficult because knowledge is embedded in routines, tools, and social networks.

\cite{alavi2001} provided a comprehensive review of knowledge management systems (KMS), identifying four key processes---creation, storage and retrieval, transfer, and application---and arguing that information technology serves as an enabler rather than a replacement for human sense-making. Their taxonomy of KMS architectures highlights the persistent tension between codification strategies (document-centric) and personalization strategies (expert-centric), a tension that persists in modern developer platforms.

Despite decades of KM research, the literature overwhelmingly assumes a knowledge consumer who can interpret ambiguity, navigate incomplete documentation, and draw on contextual judgment acquired through social interaction. Brown and Duguid~\cite{brown2000sociallife} argue that this contextual judgment is irreducibly social: information acquires meaning through the practices in which it is embedded, and document-centric knowledge systems systematically underweight the practice context in which knowledge is actually used. Lave and Wenger~\cite{lavewenger1991situated} establish that newcomers acquire situated expertise through \emph{legitimate peripheral participation}---learning by partial, supervised engagement with the practices of an established community, rather than by reading documentation in isolation. Both observations sharpen the problem this paper addresses: knowledge architectures that require situated interpretation, or social participation, are inaccessible to autonomous agents and impose a sustained tax on the senior practitioners who must transfer that situated knowledge to newcomers. This assumption fails for two increasingly important classes of consumer. Autonomous AI agents lack the ability to seek clarification through conversation, cannot tolerate ambiguity in procedural instructions, and operate under strict computational constraints such as finite context windows. But the assumption also fails, in a different register, for human newcomers: newly onboarded engineers and developers navigating unfamiliar codebases face the same knowledge transfer barriers that Szulanski~\cite{szulanski1996stickiness} identified as \emph{knowledge stickiness}---causal ambiguity, insufficient absorptive capacity, and dependence on arduous interpersonal relationships---compounded by organizational scale. The knowledge management literature has extensively documented these barriers~\cite{cohen1990absorptive, argote2000knowledge} but has not produced a knowledge primitive designed to serve both human and machine consumers under their respective constraints. This paper argues that the \emph{skill} is that primitive: encoding institutional knowledge in composable, action-oriented, governance-aware units that agents consume directly and through which the engineers working with those agents receive institutionally grounded guidance.

\subsection{Internal Developer Platforms and Golden Paths}
\label{sec:rw-idp}

Internal Developer Platforms (IDPs) have emerged as the principal mechanism through which organizations reduce developer cognitive load and enforce engineering standards at scale. \cite{beetz2023platform} traced the evolution of platform engineering from ad hoc infrastructure automation to a disciplined practice of building self-service developer platforms, positioning it as the next stage beyond DevOps. \cite{gartner2024platform} projected that by 2026, 80\% of software engineering organizations will establish platform teams, underscoring the industry's convergence on this model. The \cite{cncf2024platforms} white paper formalized the concept further, defining platforms as curated collections of capabilities that reduce cognitive load while maintaining organizational governance.

The intellectual foundation for cognitive load reduction in platform engineering draws on \cite{sweller1988cognitive}, whose cognitive load theory distinguished intrinsic, extraneous, and germane load in learning contexts. \cite{skelton2019team} applied these principles to software organizations through the Team Topologies framework, arguing that platform teams should absorb extraneous cognitive load so that stream-aligned teams can focus on value delivery. Golden paths---opinionated, well-supported default workflows---operationalize this principle by providing developers with paved roads through otherwise complex decision landscapes.

\cite{greiler2022developer} proposed an actionable framework for understanding and improving developer experience, emphasizing that developer satisfaction and productivity depend on the friction encountered in daily workflows. Their framework highlights the importance of tooling, documentation quality, and process clarity---dimensions that golden paths are explicitly designed to optimize.

Recent industry research underscores the practical consequence of this misalignment: developer experience studies report that the friction of locating institutional information across human-oriented formats remains a primary productivity drain even as AI tool adoption accelerates~\cite{atlassian2025devex, greiler2022developer}.

However, existing golden paths are architected for \emph{human} navigation. They rely on documentation written in natural language, portal interfaces designed for visual browsing, and contextual cues that assume a human operator capable of interpolating between steps. When an autonomous agent must traverse a golden path, these human-oriented affordances become obstacles: prose instructions must be parsed into executable actions, implicit prerequisites must be made explicit, and governance constraints must be machine-enforceable. This paper addresses this gap by reconceptualizing golden paths as \emph{knowledge topologies} composed of interconnected skills that agents can discover, validate, and execute programmatically.

\subsection{LLM-Based Agents and Tool Use}
\label{sec:rw-agents}

The ReAct paradigm introduced by \cite{yao2023react} demonstrated that interleaving reasoning traces with action execution enables large language models (LLMs) to solve complex tasks by grounding chain-of-thought reasoning in environmental feedback. This work established the foundational architecture for LLM-based agents: an iterative loop of observation, reasoning, and action that mirrors classical sense--plan--act cycles in robotics.

Concurrent research has focused on equipping LLMs with external tools. \cite{schick2023toolformer} showed that language models can learn to invoke APIs autonomously by embedding tool calls within generated text. \cite{patil2023gorilla} scaled this approach to large API corpora, demonstrating that retrieval-augmented fine-tuning enables accurate tool selection across thousands of endpoints. \cite{qin2023toolllm} further expanded the frontier, constructing benchmarks and training pipelines for mastering over 16,000 real-world APIs, revealing that tool selection accuracy degrades significantly as the tool space grows.

In the software engineering domain, \cite{jimenez2024swebench} established the SWE-bench benchmark, revealing both the promise and limitations of LLM agents on real-world GitHub issues. \cite{yang2024sweagent} introduced agent--computer interfaces optimized for code navigation and editing, achieving state-of-the-art results by designing interaction paradigms tailored to the software engineering task structure. \cite{wang2024openhands} extended this line of work into a general-purpose platform for AI software developers, emphasizing modularity and extensibility.

Multi-agent architectures have also gained traction. \cite{wu2023autogen} proposed a framework for orchestrating multiple conversational agents, while \cite{hong2023metagpt} demonstrated that assigning distinct roles to agents within a software development pipeline improves coherence and output quality.

Beyond multi-agent coordination, recent work has examined the cognitive and compositional structures that underpin effective agent behavior. Sumers et al.~\cite{sumers2024coala} proposed the Cognitive Architectures for Language Agents (CoALA) framework, which unifies agent designs around modular components: working memory, long-term memory (procedural and declarative), action spaces, and decision-making procedures. The skill primitive proposed in this paper maps onto CoALA's procedural memory---structured knowledge that guides action selection---while the Knowledge Topology and Activation Policy correspond to aspects of CoALA's grounding and decision-making modules. Wang et al.~\cite{wang2023voyager} demonstrated Voyager, an open-ended embodied agent that autonomously discovers, stores, and retrieves reusable skills in a growing skill library. Voyager's skill library shares structural similarities with the AKU Registry proposed in this paper, but operates in a simulated environment without organizational metadata, governance constraints, or the enterprise context that distinguishes the Knowledge Activation framework. Both CoALA and Voyager validate the skill-as-primitive intuition from different perspectives---cognitive architecture and embodied exploration, respectively---while leaving unaddressed the specific challenges of organizational knowledge delivery under context window constraints that this paper targets.

A significant industry development occurred in December 2025, when Anthropic released
the Agent Skills open standard~\cite{anthropic2025agentskills,
agentskills2025spec}, subsequently adopted by multiple major AI
platforms~\cite{microsoft2025platformagentic}. The standard formalized what earlier
ad hoc approaches---Cursor rules, GitHub Copilot custom instructions, Windsurf
rules, and repository-level context files---had been converging toward: a portable,
discoverable format for packaging agent-consumable knowledge as AI Skills. This paper
builds on the Agent Skills standard, providing the theoretical framework and enterprise
specialization that the format specification does not address: why skills must be
compact (the Context Window Economy), how to produce them for institutional knowledge
delivery (the Knowledge Activation pipeline), and how to govern them at enterprise
scale (the Governance architecture).

Concurrent empirical work by Vasilopoulos~\cite{vasilopoulos2026codified} developed a
tiered \emph{codified context infrastructure} during the construction of a 108{,}000-line
production system across 283 development sessions. The architecture independently
instantiates several core Knowledge Activation principles: structured artifacts written
explicitly for agent consumption, trigger-based knowledge routing analogous to Intent
Declaration, and a hot/cold memory separation that mirrors the injection stage's precision
requirement. Vasilopoulos documents that specification staleness is the primary failure
mode at scale and provides case studies illustrating how pre-structured domain knowledge
prevents the trial-and-error failure cascades this paper formalizes as context
rot. The contribution is complementary: where that work provides empirical grounding in a
single-developer codebase context, this paper provides the formal framework, enterprise
governance architecture, and organizational-scale deployment model.

At the system level, Cemri et al.~\cite{cemri2025multiagent} investigated why multi-agent
LLM systems fail, identifying failure modes that compound when agents lack structured
knowledge about their operational context---a finding consistent with the Institutional
Impedance Mismatch described in this paper.

Despite these advances, existing agent frameworks treat tool use as a generic capability: agents receive tool descriptions and must reason about when and how to invoke them. What is notably absent is a mechanism for delivering \emph{structured institutional knowledge}---the policies, conventions, architectural decisions, and operational constraints that govern how tools should be used within a specific enterprise context. This paper introduces skills as the vehicle for such knowledge delivery, transforming agents from generic tool users into organizationally situated actors.

\subsection{Context-Aware Knowledge Delivery}
\label{sec:rw-context}

Retrieval-Augmented Generation (RAG), introduced by \cite{lewis2020rag}, established the dominant paradigm for supplying LLMs with external knowledge at inference time. By retrieving relevant passages from a document corpus and prepending them to the prompt, RAG reduces hallucination and enables knowledge-intensive tasks without retraining. Standard RAG systems retrieve \emph{passages}---fragments of text selected by semantic similarity to the query. Section~\ref{sec:knowledge-activation} examines the structural differences between passage-level retrieval and the action-oriented knowledge delivery proposed in this paper.

The finite context window of LLMs imposes hard constraints on knowledge delivery.
Recent research has identified \emph{context rot}---the measurable degradation of LLM
performance as context windows fill with accumulated irrelevant or redundant
content---as a fundamental challenge for long-running agent
sessions~\cite{chromaresearch2025contextrot}. This finding provides empirical
motivation for context-efficient knowledge delivery: every token of poorly structured
or redundant knowledge injected into an agent's context actively degrades the agent's
reasoning capability, making the design of compact, high-signal knowledge primitives
not merely an efficiency concern but a correctness
requirement~\cite{anthropic2025contextengineering}. \cite{jiang2023longllmlingua}
addressed the compression dimension through prompt compression techniques that preserve
essential information while reducing token counts, demonstrating that significant
compression ratios are achievable without proportional performance degradation. \cite{brown2020language} established that LLMs can acquire new capabilities through in-context learning with few-shot examples, suggesting that carefully curated context can substitute for parametric knowledge---but the question of \emph{what} to place in context, and how to structure it, remains underexplored in enterprise settings.

Task decomposition strategies offer complementary approaches to managing knowledge complexity. \cite{khot2023decomposed} proposed decomposed prompting, in which complex tasks are broken into modular sub-problems solved by specialized handlers. \cite{zhou2023leasttomostprompting} demonstrated that least-to-most prompting enables models to solve problems of increasing complexity by building on previously solved sub-problems. These decomposition strategies align with the modular, composable nature of the skill primitive proposed in this paper.

The workflow patterns literature~\cite{vanderaalst2003workflow} provides a complementary perspective on structured process composition. Van der Aalst et al.'s catalog of control-flow patterns---including sequence, parallel split, synchronization, and deferred choice---formalizes the compositional structures that skill continuation paths must support. However, workflow patterns assume deterministic orchestration by a process engine; the skill primitive extends this to probabilistic composition by an autonomous agent that selects and sequences skills at runtime based on contextual reasoning.

\cite{pan2024unifying} surveyed the integration of knowledge graphs with LLMs, identifying three paradigms: KG-enhanced LLMs, LLM-enhanced KGs, and synergistic architectures. Knowledge graphs offer structured, relational representations that complement the unstructured nature of RAG retrieval, yet their application to organizational knowledge delivery for software engineering agents remains nascent.

The emerging discipline of \emph{context engineering}~\cite{anthropic2025contextengineering} provides a broader frame for these challenges, encompassing the full set of techniques for managing what information reaches an agent's context window. Mei et al.~\cite{mei2025survey} provide a comprehensive academic survey synthesizing over 1{,}400 papers across context retrieval, context processing, memory management, and tool-integrated reasoning, establishing context engineering as a formal research discipline rather than a practitioner heuristic. Zhang et al.~\cite{zhang2026ace} contribute the Agentic Context Engineering (ACE) framework, identifying \emph{brevity bias}---the tendency of iterative optimization to collapse contexts into short, generic prompts---and \emph{context collapse}---the progressive erosion of detail through successive rewrites---as systematic failure modes that parallel the context rot phenomenon documented by \cite{chromaresearch2025contextrot}.

The Knowledge Activation framework proposed in this paper can be understood as a specialized form of context engineering focused specifically on \emph{institutional knowledge}: the organizational procedures, governance constraints, and operational metadata that distinguish enterprise deployment from general-purpose agent design. Where context engineering provides the general discipline, Knowledge Activation provides a specific methodology and artifact (the skill) for the institutional knowledge dimension.

\subsection{Developer Experience and Productivity}
\label{sec:rw-devex}

A growing body of research has formalized the measurement of developer experience and productivity. Forsgren et al.~\cite{forsgren2021space} proposed the SPACE framework, identifying five dimensions---Satisfaction, Performance, Activity, Communication, and Efficiency---that together capture the multifaceted nature of developer productivity. Noda et al.~\cite{noda2023devex} refined this perspective into the DevEx framework, arguing that three core dimensions---feedback loops, cognitive load, and flow state---are the primary drivers of developer experience, with cognitive load reduction identified as the highest-leverage intervention. Xia et al.~\cite{xia2018comprehension} provided empirical grounding: in a large-scale field study of 78 professional developers across 3{,}148 working hours, developers spent approximately 58\% of their time on program comprehension activities, with senior developers requiring significantly less time than juniors---a gap attributable to accumulated institutional knowledge.

Recent industry data reveals a paradox at the intersection of DevEx and AI adoption: despite widespread AI tool adoption, developers report that locating institutional information across scattered formats remains a leading productivity drain even as individual coding tasks accelerate~\cite{dora2025ai, atlassian2025devex, cortex2024productivity}. Section~\ref{sec:context-window-economy} develops this paradox in detail.

This literature identifies the symptoms---cognitive overload, context-switching friction, information-seeking overhead---but does not prescribe a knowledge architecture intervention. The Knowledge Activation framework addresses this gap directly: skills reduce cognitive load by externalizing institutional knowledge from working memory (for humans) and context windows (for agents) into pre-structured, injectable primitives, targeting Noda et al.'s highest-leverage DevEx dimension.

\subsection{Synthesis: The Knowledge Architecture Gap}
\label{sec:rw-gap}

The critical gap across these five literatures is the absence of a knowledge delivery mechanism that is simultaneously \emph{action-oriented} (encoding what to do, not merely what is known), \emph{governance-aware} (embedding organizational constraints and policies), \emph{context-efficient} (optimized for the constraints under which knowledge consumers operate), and \emph{beneficial to both agents and the engineers who work with them} (the same knowledge primitive that makes an agent effective also compresses onboarding, reduces cross-team friction, and eliminates correction cascades for the humans in the loop). The skill as proposed in this paper is designed to address precisely this gap, unifying retrieval, structure, and governance into composable knowledge units that agents activate on demand---delivering institutionally grounded outcomes to every participant in the workflow.

\section{The Context Window Economy}
\label{sec:context-window-economy}

Traditional workflow automation proceeds from a bird's-eye view. The designer models the
complete state space---every branch, every exception, every transition---and encodes the
optimal path before any execution begins. Like an observer above a maze who traces the
correct route before anyone enters, the workflow designer eliminates navigation
uncertainty at the cost of total anticipatory design. The system works only for
situations the designer anticipated; every other situation is an exception or an error.

LLM agents navigate from within. Processing sequentially, each generation conditioned on
what preceded it, an agent moves through organizational processes like a walker in a
maze---with local visibility only, bounded by what is currently in the context window, and
no institutionally validated path. This is not a limitation to be engineered away; it is a structural
property of autoregressive generation that enables what global pre-computation cannot:
handling situations no designer anticipated, by reasoning forward from locally available
context. Suchman's foundational critique of the planning model in human--machine
interaction~\cite{suchman1987plans} articulated this insight decades before LLMs existed:
plans are resources for situated action, not deterministic programs to be executed.
LLM agents operate under precisely this regime---their plans are hypotheses drawn from
trained heuristics, and execution is fundamentally situated in the organizational context
the agent encounters at each step.

This creates the foundational design challenge that Knowledge Activation addresses. The
agent does not need a map of the entire organization---it cannot use one in any
meaningful sense, and loading the whole knowledge base into context is demonstrably
counterproductive~\cite{gloaguen2026evaluating}. What it needs at each decision point
is a \emph{locally sufficient} artifact: structured knowledge for the junction it is
currently navigating. The Context Window Economy formalizes the constraints that make
this local sufficiency both necessary and difficult, recasting institutional knowledge
management as a resource allocation problem governed by the finite capacity of the
agent's context window.

\begin{figure}[htbp]
  \centering
  \includegraphics[width=\textwidth]{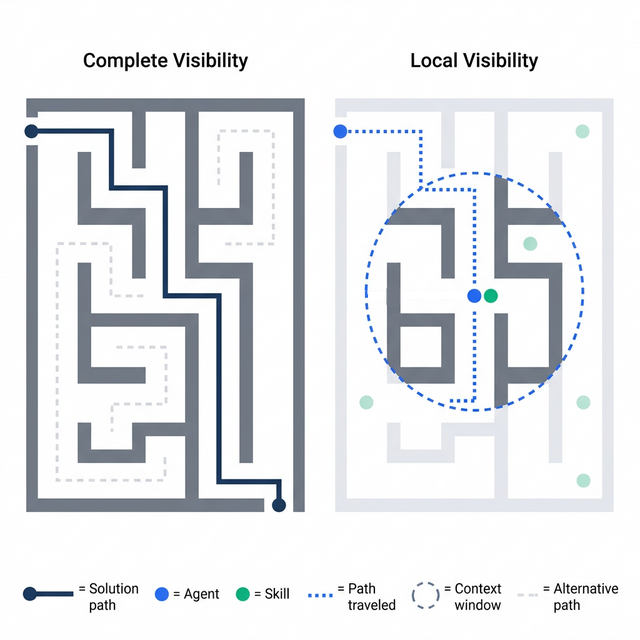}
  \caption{Two paradigms for navigating organizational knowledge. In the deterministic paradigm (left), the complete process topology is visible, alternatives are evaluated, and the optimal path is pre-computed before execution begins; the approach works only for situations the designer anticipated. In the situated paradigm (right), the agent operates with local visibility bounded by the context window, traversing the same topology one junction at a time. Green markers represent Skills---locally sufficient knowledge artifacts that provide structured guidance at each decision point. Skills are distributed throughout the topology (faded markers) but become actionable only when the agent reaches each junction.}
  \label{fig:navigation-paradigm}
\end{figure}

\subsection{From Abundance to Scarcity}

Human knowledge workers operate in an environment of effective abundance. A developer browsing an internal wiki can follow hyperlinks, skim headings, invoke search, and iteratively refine queries---consuming arbitrarily many documents across unbounded time. The cognitive cost of accessing one document does not preclude access to another; attention is the scarce resource, but it is managed dynamically by the human agent through selective reading and satisficing strategies.

LLM-based agents face a fundamentally different regime. Each invocation of the model receives a context window of fixed token capacity. All information the agent can reason over---the task description, relevant knowledge, tool definitions, prior conversation history, and intermediate results---must fit within this window. Knowledge that is not present in the context is, for practical purposes, nonexistent to the agent. Shannon's foundational work on communication channels~\cite{shannon1948mathematical} established that every channel has a finite capacity, and that efficient communication requires encoding information in ways that respect this capacity. The context window of an LLM is precisely such a channel: a bandwidth-limited conduit through which organizational knowledge must pass to influence agent behavior.

This shift from abundance to scarcity transforms knowledge management from a curation problem (organizing knowledge so humans can find it) into an \emph{allocation} problem (selecting and encoding knowledge so that agents receive maximum value within a hard capacity constraint).

\begin{definition}[Context Window Economy]
\label{def:context-window-economy}
The \textbf{Context Window Economy} is the regime in which knowledge delivery to autonomous AI agents is governed by three binding constraints---token budget, attention decay, and latency cost---such that institutional knowledge must be curated, compressed, and allocated as a scarce resource to maximize the probability of successful task completion per unit of context consumed.
\end{definition}

\subsection{The Three Constraints}

The Context Window Economy is defined by three interacting constraints that jointly determine the feasibility and quality of knowledge-informed agent behavior.

\paragraph{Token Budget.}
The most visible constraint is the hard limit on context length. Contemporary LLMs accept context windows ranging from thousands to millions of tokens, yet the knowledge requirements of enterprise software tasks---spanning architectural documentation, API specifications, compliance policies, deployment runbooks, and service catalogs---routinely exceed even the largest available windows. The token budget forces a selection problem: which knowledge artifacts should be included, and which must be omitted? Every token allocated to one piece of knowledge is a token unavailable for another. This zero-sum dynamic is the defining characteristic of the economy.

\paragraph{Attention Decay.}
Even within the nominal token budget, not all positions in the context window are equal. Empirical studies have demonstrated that attention mechanisms in transformer architectures exhibit positional biases: information placed in the interior of long contexts is attended to less reliably than information near the beginning or end~\cite{liu2024lostinmiddle}. This phenomenon---termed the ``lost in the middle'' effect---means that simply fitting knowledge within the window does not guarantee that it will influence model behavior. The \emph{effective} capacity of the context window is therefore smaller than its nominal capacity, and knowledge placement becomes a design variable with material consequences for agent performance.

\paragraph{Latency Cost.}
Larger context windows incur greater computational cost. Inference time and financial cost scale with context length, creating a direct economic trade-off between knowledge richness and operational efficiency. In enterprise settings where agents may execute hundreds or thousands of invocations per day, the marginal cost of additional context tokens aggregates into significant expenditure. The latency cost constraint motivates not merely fitting within the window, but minimizing the tokens consumed while preserving task-relevant information.

\subsection{The Economic Framing}

These three constraints suggest an economic framing. Each unit of knowledge delivered to an agent has a \emph{cost}, measured in tokens consumed, and a \emph{value}, measured in its contribution to the probability of successful task completion. The optimization problem facing the knowledge architect is structurally analogous to the classical knapsack problem: knowledge artifacts of varying value and token cost must be selected for a container of limited capacity. Unlike the classical knapsack, the value function is not directly observable and must be estimated through proxy measures such as task completion rate, correction frequency, or agent accuracy. The analogy highlights the \emph{allocation} nature of the challenge: knowledge delivery to agents is a resource allocation problem, not a curation problem.

A concrete illustration clarifies the magnitude of the opportunity. A typical enterprise deployment runbook---narrative prose with screenshots, context-setting paragraphs, and cross-references to other documents---might consume 2{,}000 tokens when injected into an agent's context. The equivalent AKU encoding the same procedural knowledge in the seven-component schema (Section~\ref{sec:atomic-knowledge-units}) might consume 300 tokens: a structured procedure with explicit tool bindings, governance constraints, and continuation paths, stripped of rhetorical scaffolding. If both artifacts enable the same task completion, the AKU achieves roughly 6--7$\times$ the \emph{knowledge density}---actionable information per token---of the narrative runbook. The Agent Skills open standard's recommendation that skills remain under 5{,}000 tokens~\cite{agentskills2025spec} reflects this design principle: compact, high-density artifacts outperform verbose documentation within the constraints of the context window.

The parallel to information theory reinforces this point. Shannon~\cite{shannon1948mathematical} demonstrated that efficient communication requires matching the encoding to the capacity of the channel. In the Context Window Economy, the ``source'' is institutional knowledge, the ``channel'' is the context window, and the encoding scheme is the knowledge representation format. Documents and wikis represent a na\"ive encoding---high redundancy, low information density, poor alignment with the channel's constraints. The Context Window Economy demands purpose-built encodings that maximize knowledge density. The parallel is structural rather than mathematical---the context window lacks a noise model in Shannon's technical sense---but the design implication is the same: what occupies the channel matters more than how large the channel is.

\subsection{Contrast with Human Knowledge Consumption}

The distinction between human and agent knowledge consumption extends beyond the finite-versus-unbounded contrast. Table~\ref{tab:human-vs-agent-knowledge} summarizes the key differences.

\begin{table}[ht]
\centering
\caption{Knowledge consumption patterns: human developers versus LLM-based agents.}
\label{tab:human-vs-agent-knowledge}
\begin{tabularx}{\textwidth}{lXX}
\toprule
\textbf{Dimension} & \textbf{Human Developer} & \textbf{LLM-Based Agent} \\
\midrule
Access model & Iterative search and browse & Single-shot context injection \\
Capacity & Effectively unbounded (over time) & Fixed token window per invocation \\
Attention & Selective, self-directed & Positional, architecture-dependent \\
Persistence & Long-term memory across sessions & Stateless across invocations (by default) \\
Format tolerance & Handles ambiguity, formatting noise & Sensitive to token waste, irrelevant content \\
Feedback loop & Can ask clarifying questions in real time & Must act on context as given \\
\bottomrule
\end{tabularx}
\end{table}

These differences have a critical implication: knowledge artifacts designed for human consumption---narrative documentation, lengthy runbooks, sprawling wiki pages---are \emph{systematically misaligned} with the requirements of agent consumption. They waste tokens on rhetorical scaffolding, duplicate information across sections, embed critical details within prose that may fall in low-attention regions, and omit the structured metadata (tool bindings, permission boundaries, trigger conditions) that agents require.

\subsection{Context Rot and the Productivity Paradox}

The phenomenon of \emph{context rot}---the progressive degradation of LLM performance as the context window fills with accumulated content---has been documented as a systematic failure mode in long-context agent interactions~\cite{chromaresearch2025contextrot}. As tokens accumulate from prior exchanges, tool outputs, and intermediate reasoning, the model's ability to attend to relevant information diminishes, producing increasingly unreliable outputs. Context rot is not merely a theoretical concern; it is an operational reality that compounds with every turn of agent interaction.

In enterprise agent deployments, the Institutional Impedance Mismatch described in Section~\ref{sec:introduction} creates a specific mechanism that accelerates context rot. When an agent lacks the institutional knowledge required for a task, it resorts to inference from its general training---guessing at deployment targets, fabricating service names, or applying generic procedures where organization-specific ones are required. These guesses fail. The human operator provides a correction. The agent revises its approach and retries. Each cycle of guess, failure, correction, and retry consumes tokens: the failed attempt, the user's correction, the agent's revised reasoning, and the second attempt all persist in the context window. After several such cycles, the window is dominated by the detritus of failed interactions rather than actionable guidance, and the agent's performance degrades further---a vicious cycle.

This characterization holds even for LLMs with extended planning capabilities. A
planning LLM forms a hypothesis about the path---informed by its parametric knowledge of
how similar processes generally work---but plans are bounded by what the agent knows at
planning time. Organizational specificity (the deployment gate requiring ticket linkage,
the compliance annotation required before regulated-environment promotion, the
service-specific blast radius threshold) is precisely what training data does not reliably
contain. Plans fail at the gap between trained heuristics and organizational
particularity. Zhang et al.'s Agentic Context Engineering framework identifies two
systematic failure modes of this kind: \emph{brevity bias}, in which iterative
optimization collapses agent contexts into short generic prompts, and \emph{context
collapse}, in which successive rewrites erode organizational
detail~\cite{zhang2026ace}. These findings reinforce the structural nature of the
problem: it is not that agents plan poorly, but that no amount of planning sophistication
can substitute for locally accurate institutional knowledge.

This dynamic produces what may be termed the \emph{productivity paradox} of enterprise AI
agents. Organizations deploy agents to increase developer productivity, yet the absence
of institutional knowledge transforms the agent from an autonomous actor into an
expensive, context-rotting autocomplete that demands constant human-in-the-loop
correction. The DORA 2025 report~\cite{dora2025ai} documents this paradox at industry
scale: despite 90\% AI tool adoption among surveyed organizations, there was no clear
link between adoption and reductions in developer friction or burnout. Individual speed
increased---more tasks completed, more pull requests merged---but organizational delivery
did not improve.

\begin{figure}[htbp]
  \centering
  \includegraphics[width=\textwidth]{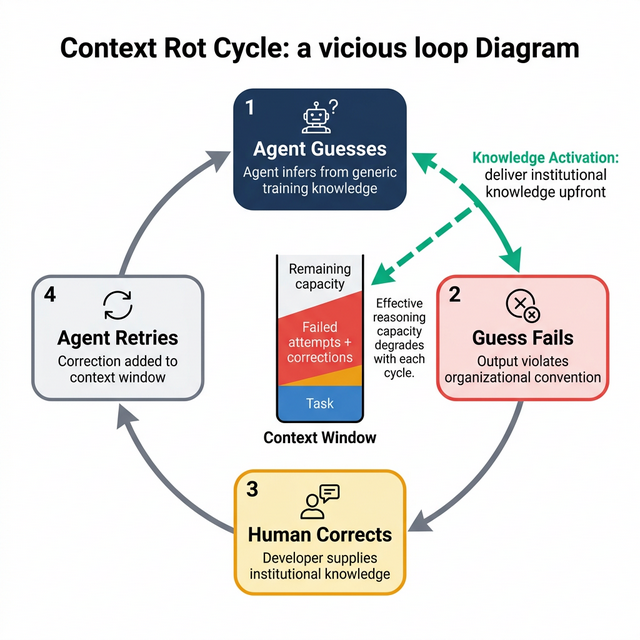}
  \caption{The Context Rot Cycle. When an agent lacks institutional knowledge, it enters a vicious cycle of guess, failure, correction, and retry. Each iteration fills the context window with detritus, degrading effective reasoning capacity. Knowledge Activation breaks the cycle by delivering pre-structured institutional knowledge upfront.}
  \label{fig:context-rot-cycle}
\end{figure}

\paragraph{The Institutional Knowledge Tax.}

The productivity paradox has a sociotechnical dimension that amplifies its organizational impact. The developers best equipped to diagnose and correct the agent's institutional knowledge gaps are precisely those with the deepest organizational expertise---senior engineers who have internalized deployment procedures, compliance requirements, and architectural conventions over years of practice. For these engineers, every agentic interaction that lacks pre-structured institutional knowledge becomes an implicit demand to externalize their tacit knowledge~\cite{nonaka1995} in real time through iterative corrections---an overhead this paper terms the \emph{institutional knowledge tax}.

Recent industry data illustrates the consequence. Atlassian's 2025 developer experience
survey of 3{,}500 developers found that while nearly all respondents report time savings
from AI tools, developers simultaneously identify finding information---across services,
documentation, and APIs---as a leading source of productivity
loss~\cite{atlassian2025devex}. A survey of engineering leaders at large software
companies corroborates the finding: 40\% of developers cite ``trouble finding context''
as their top productivity pain point~\cite{cortex2024productivity}. AI-driven
productivity gains are largely offset by organizational inefficiencies rooted in knowledge
architecture, not tool capability. The 2025 Stack Overflow Developer Survey deepens the
pattern: experienced developers report the highest rates of distrust in AI tool accuracy,
and two-thirds of respondents identify ``solutions that are almost right, but not quite''
as a leading frustration~\cite{stackoverflow2025survey}---a symptom consistent with the
Institutional Impedance Mismatch.  A randomized controlled trial by METR provides
further evidence: experienced open-source developers were an estimated 19\% slower when
using AI coding tools, with developers accepting fewer than half of AI-generated
suggestions after spending significant time on
review~\cite{metr2025experienced}. The METR study examined open-source contexts, where the
Institutional Impedance Mismatch is minimized by the overlap between training data and
repository conventions. That the mechanism is visible even in this most favorable
setting---experienced developers rejecting AI output that fails to reflect deep
codebase-specific knowledge---suggests it is structurally amplified in enterprise
contexts where institutional knowledge diverges sharply from training distributions.

Vasilopoulos~\cite{vasilopoulos2026codified} provides a concrete observational instance:
in a 70-day production development study, a domain-expert agent specification encoding
determinism theory identified three subtle bugs in a single session that had eluded five
prior unguided attempts---directly illustrating how pre-loaded institutional knowledge
eliminates the guess-fail-correct-retry cycle. The same study reports that repeated
cross-session explanation of domain knowledge was the primary signal for
codification---a practitioner-derived observation of the institutional knowledge tax in
action.

The tax is not limited to agent interactions. All knowledge consumers---the term introduced in Section~\ref{sec:introduction} to encompass agents, new engineers, and cross-team contributors---face the same
structural deficit: studies of software developer onboarding consistently identify tacit
knowledge transfer---not technical setup---as the primary barrier to
productivity~\cite{ju2021onboarding, balali2024newcomer}. The knowledge that a senior
engineer must supply to an agent in real time is the same knowledge a newcomer absorbs
over months of social interaction. The program comprehension gap between senior and junior developers documented by Xia et al.~\cite{xia2018comprehension} (Section~\ref{sec:rw-devex}) is attributable to precisely this accumulated institutional knowledge. The institutional knowledge tax, in this light, is not an
agent-specific phenomenon but a general consequence of knowledge stickiness~\cite{szulanski1996stickiness} that the arrival of AI agents has made acute and
measurable.

When this tax exceeds the perceived productivity benefit, senior engineers withdraw from agentic workflows---not because the technology is inadequate, but because the knowledge architecture surrounding it is.

The Context Window Economy framework identifies this as a knowledge architecture problem, not a model capability problem. Larger context windows delay but do not prevent context rot; more capable models reduce but do not eliminate institutional knowledge gaps. The solution lies in delivering the right institutional knowledge upfront---through the Knowledge Activation pipeline developed in the following sections---so that the guess-fail-correct-retry cycle is eliminated at its source.

\subsection{The Imperative for a New Knowledge Architecture}

The Context Window Economy renders existing knowledge management approaches insufficient for the era of autonomous AI agents. Document-centric knowledge bases optimize for human readability, not token efficiency. Search-based retrieval systems (including Retrieval-Augmented Generation~\cite{lewis2020rag}) return passage-level chunks that may contain relevant information but lack the structural guarantees needed for reliable agent action: they do not specify which tools to invoke, what permissions are required, or how the retrieved knowledge relates to the organization's governance policies.

What is needed is a knowledge architecture designed from first principles for agent consumption---one that produces artifacts optimized for the three constraints of the Context Window Economy. Such artifacts must be compact (respecting the token budget), front-loaded with actionable content (mitigating attention decay), and structured for rapid parsing (minimizing latency overhead). They must also carry the organizational metadata---ownership, permissions, service bindings---that enterprise agents require to act not merely correctly, but \emph{governably}.

The following sections introduce exactly such an architecture. Section~\ref{sec:knowledge-activation} presents Knowledge Activation as the process by which latent organizational knowledge is transformed into context-window-efficient, agent-executable form. Section~\ref{sec:atomic-knowledge-units} defines the Atomic Knowledge Unit---the skill---as the fundamental artifact of this architecture: a purpose-built knowledge primitive designed to maximize value within the constraints of the Context Window Economy.

\section{Knowledge Activation: From Retrieval to Execution}
\label{sec:knowledge-activation}

The constraints articulated by the Context Window Economy demand a response: knowledge artifacts purpose-built for consumption by agents and the engineers who work with them. This section introduces \emph{Knowledge Activation} as the theoretical framework for producing such artifacts---a process that transforms latent institutional knowledge into structured, governance-annotated form optimized for the constraints under which knowledge consumers operate. Knowledge Activation is the central theoretical contribution of this paper, offering a principled alternative to retrieval-based approaches for equipping knowledge consumers with institutional knowledge.

\subsection{Knowledge Retrieval versus Knowledge Activation}

The dominant paradigm for supplying LLMs with external knowledge is retrieval. Retrieval-Augmented Generation (RAG)~\cite{lewis2020rag} and its variants operate by identifying relevant passages from a document corpus and concatenating them into the model's context. The retrieved passages are \emph{informational}---they contain facts, descriptions, or instructions that the model must interpret, synthesize, and translate into action. The burden of interpretation remains with the model.

Knowledge Activation departs from this paradigm in a fundamental way. Where retrieval returns \emph{content for reading}, activation delivers \emph{guidance for acting}. An activated knowledge unit does not merely inform the agent about a topic; it equips the agent with a complete action specification: what to do, how to do it, which tools to use, what constraints to respect, and what to do next. The distinction is analogous to the difference between handing a human a reference manual and handing them a standard operating procedure with a pre-configured workstation.

This distinction holds even for advanced agentic RAG pipelines that augment retrieval with multi-step reasoning, tool use, and iterative refinement. Such systems improve how agents \emph{process} retrieved content, but the content itself remains informational---passages, documentation fragments, and code snippets that the agent must still interpret against an organizational context it does not possess. Knowledge Activation addresses a different layer: the \emph{structure and governance of the knowledge artifact itself}, ensuring that what reaches the agent is already action-ready, governance-annotated, and organizationally grounded rather than requiring runtime interpretation regardless of the agent's reasoning sophistication.

Table~\ref{tab:retrieval-vs-activation} summarizes the key contrasts.

\begin{table}[ht]
\centering
\caption{Knowledge Retrieval versus Knowledge Activation.}
\label{tab:retrieval-vs-activation}
\begin{tabularx}{\textwidth}{lXX}
\toprule
\textbf{Dimension} & \textbf{Knowledge Retrieval (RAG)} & \textbf{Knowledge Activation} \\
\midrule
Unit of delivery & Document chunk (passage) & Atomic Knowledge Unit (skill) \\
Content type & Informational text & Action-ready specification \\
Tool integration & Absent; model must infer tools & Explicit tool bindings included \\
Governance & Not encoded & Embedded constraints and permissions \\
Organizational context & Minimal or absent & Ownership, service catalog, team metadata \\
Optimization target & Relevance to query & Value per token for task completion \\
Agent burden & Interpret, plan, and act & Act according to specification \\
\bottomrule
\end{tabularx}
\end{table}

\begin{figure}[htbp]
  \centering
  \includegraphics[width=\textwidth]{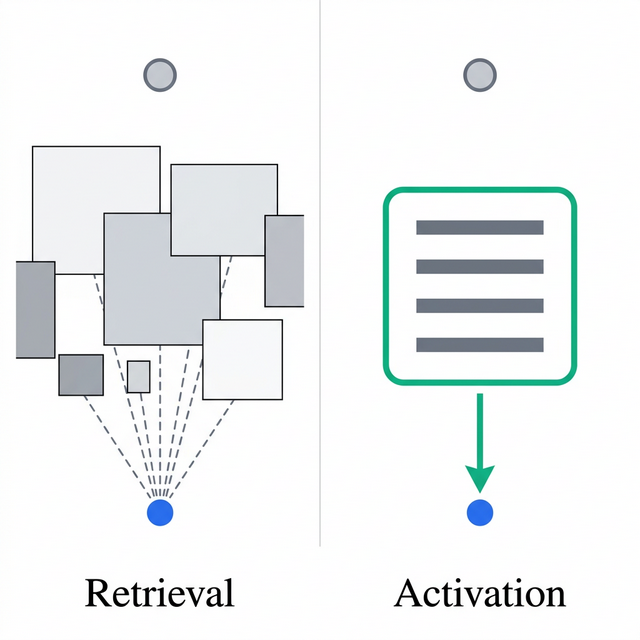}
  \caption{Knowledge Retrieval versus Knowledge Activation: architectural comparison. In retrieval-based approaches (left), the agent receives unstructured text fragments and must interpret, plan, infer tools, and determine governance constraints at runtime. In Knowledge Activation (right), the agent receives a pre-structured Atomic Knowledge Unit and acts according to the specification. The interpretation burden shifts from runtime inference to knowledge authoring.}
  \label{fig:retrieval-vs-activation}
\end{figure}

This distinction draws on, but extends, Nonaka and Takeuchi's theory of organizational knowledge creation~\cite{nonaka1995}. In their framework, \emph{externalization} is the process of converting tacit knowledge (embodied expertise, intuitions, craft knowledge) into explicit, codified form. Knowledge Activation can be understood as externalization for a machine consumer: the conversion of organizational tacit knowledge---the deployment procedures a senior engineer ``just knows,'' the compliance checks a security team enforces informally, the architectural patterns a team lead carries in memory---into structured, machine-executable representations. The critical addition is that the target consumer is not a human reader but an autonomous agent operating under the constraints of the Context Window Economy.

\subsection{The Knowledge Activation Pipeline}

\begin{figure}[htbp]
  \centering
  \includegraphics[width=\textwidth]{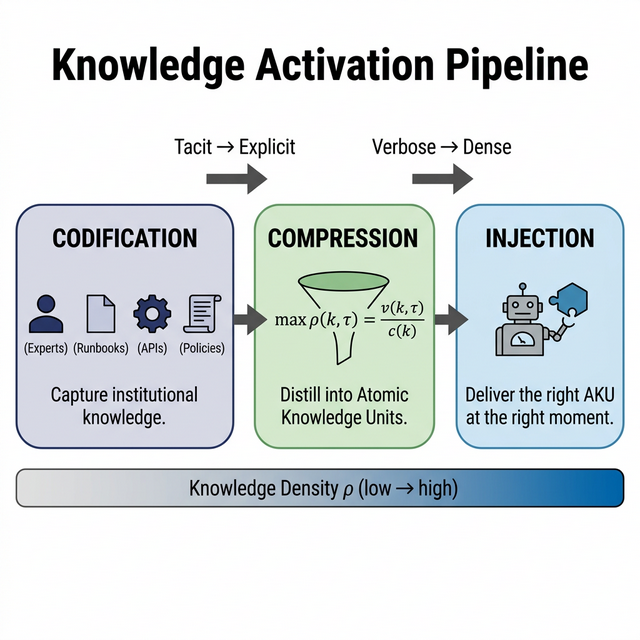}
  \caption{The Knowledge Activation Pipeline transforms latent organizational knowledge through three stages: Codification captures institutional knowledge from experts, runbooks, APIs, and policies; Compression distills it into token-efficient Atomic Knowledge Units that maximize knowledge density $\rho(k, \tau)$; Injection delivers the appropriate AKU at the point of need, equipping the agent to act with institutional accuracy or to surface the information and approval requests that require human judgment.}
  \label{fig:ka-pipeline}
\end{figure}

Knowledge Activation is not a single transformation but a three-stage pipeline, each stage addressing a distinct aspect of the problem (Figure~\ref{fig:ka-pipeline}).

\subsubsection{Stage 1: Codification}

Codification is the process of capturing institutional knowledge from its disparate sources---human experts, runbooks, platform APIs, configuration repositories, incident postmortems, and organizational policies---and rendering it in explicit, structured form. This stage addresses the well-documented challenge that much organizational knowledge is tacit~\cite{polanyi1966}: embedded in individual expertise, team practices, and cultural norms that resist straightforward documentation.

Codification in the Knowledge Activation framework is distinguished from conventional documentation by its target orientation. The goal is not to produce a human-readable document but to extract the specific elements that an autonomous agent requires: the intent behind an action, the procedural steps, the tools involved, the constraints that apply, and the organizational context in which the action is situated. This extraction process may involve structured interviews with domain experts, analysis of existing automation scripts, reverse-engineering of undocumented workflows, and synthesis of scattered policy documents.

\subsubsection{Stage 2: Compression}

Codified knowledge, even when well-structured, may remain too verbose or too diffuse for efficient delivery within a context window. The compression stage distills codified knowledge into \emph{Atomic Knowledge Units} (AKUs)---minimal, self-contained bundles that maximize knowledge density. Compression is not mere summarization; it is a principled restructuring that eliminates redundancy, foregrounds actionable content, embeds metadata inline, and organizes information for the positional attention characteristics of transformer architectures.

The compression stage operationalizes the economic logic of the Context Window Economy (Section~\ref{sec:context-window-economy}): it minimizes the token cost of each knowledge artifact while preserving or maximizing its task-completion value. Techniques from prompt compression research~\cite{jiang2023longllmlingua} inform this stage, but the Knowledge Activation framework extends beyond token-level compression to structural compression---redesigning the knowledge artifact itself for optimal agent consumption.

\subsubsection{Stage 3: Injection}

The final stage delivers the right AKU to the right agent at the right moment. Injection is the runtime counterpart to the design-time stages of codification and compression. It requires a matching mechanism that maps an agent's current task context to the most relevant AKU(s), analogous to how an operating system loads the appropriate library into memory when a program requires it.

Effective injection must satisfy three requirements: \emph{precision}, delivering only those AKUs that are relevant to the current task and avoiding context pollution with irrelevant material; \emph{timeliness}, activating knowledge at the point in the workflow where it is needed rather than front-loading the entire context; and \emph{composability}, allowing multiple AKUs to be combined when a task requires knowledge from several domains.

The injection stage connects the Knowledge Activation framework to the broader agent architecture. In tool-augmented agent systems~\cite{schick2023toolformer, patil2023gorilla}, the agent selects tools based on task requirements; in the Knowledge Activation framework, AKUs---skills---are themselves the ``tools'' through which organizational knowledge is delivered, each carrying not only procedural guidance but also the governance constraints and organizational metadata that enterprise environments demand.

\subsection{Building on the Agent Skills Open Standard}

The Knowledge Activation pipeline builds on the Agent Skills open standard released by Anthropic in December 2025~\cite{anthropic2025agentskills, agentskills2025spec}, subsequently adopted across major AI coding tools and agent frameworks including OpenAI Codex CLI, Microsoft's Agent Framework, Cursor, and GitHub Copilot~\cite{microsoft2025platformagentic}. The standard emerged from the convergent evolution of earlier ad hoc approaches---Cursor rules, GitHub Copilot custom instructions, Windsurf rules, and repository-level context files---each of which independently evolved to deliver structured knowledge into an agent's context window. The Agent Skills specification standardized what these precursors were reaching toward: a portable, discoverable format for packaging agent-consumable knowledge.

The specification implements a \emph{progressive disclosure} pattern: at startup, agents load only the name and description of each available skill (approximately 100 tokens); when a task matches a skill's description, the agent reads the full instructions into context (the standard recommends under 5{,}000 tokens); the agent then follows the instructions, optionally loading referenced files or executing bundled scripts as needed. This three-stage pattern---discovery, activation, execution---maps directly to the Knowledge Activation pipeline. Codification produces the skill artifact. Compression ensures that the artifact respects the standard's context-efficiency guideline. Injection implements the runtime activation mechanism. Implementations such as Claude Code extend the base standard with activation controls (\texttt{disable-model-invocation}, \texttt{user-invocable}, scoping hierarchies) and dynamic context injection that preprocesses shell commands into skill content at activation time---precedents for the richer Activation Policy proposed in Section~\ref{sec:golden-path-architecture}.

The standard's minimal schema---requiring only a name, description, and markdown body---is a deliberate design choice that enabled adoption across more than thirty agent providers. However, this simplicity means the standard defines a \emph{format} without addressing the specific requirements of \emph{institutional knowledge delivery} at enterprise scale. Most AI Skills in the wild are general-purpose agent instructions: coding conventions, tool preferences, repository guidelines. They lack governance constraints, deterministic validators, and the continuation paths that connect isolated skills into navigable workflows. The AKU extensions proposed here trade some of that simplicity for the institutional rigor that enterprise deployment demands---formalizing what an AI Skill must become to serve as an institutional knowledge primitive (Section~\ref{sec:atomic-knowledge-units}).

Because skills are knowledge-layer artifacts rather than model-layer artifacts, the framework is inherently LLM-agnostic (Section~\ref{sec:discussion}).

\subsection{Knowledge Density as an Optimization Target}

A central concept in the Knowledge Activation framework is \emph{knowledge density}: the ratio of actionable, task-relevant information to the total tokens consumed by a knowledge artifact. (The term ``knowledge density'' has prior usage in information retrieval and library science with different meanings; in this paper it refers specifically to the ratio of task-completion value to token cost for agent-consumed knowledge artifacts.) Formally:

\begin{equation}
\rho(k, \tau) = \frac{v(k, \tau)}{c(k)}
\label{eq:knowledge-density}
\end{equation}

\noindent where $v(k, \tau)$ is the value of artifact $k$ for task $\tau$ and $c(k)$ is its token cost. Knowledge density provides a single metric for evaluating alternative knowledge representations. A verbose wiki page and a compact skill definition may encode the same underlying knowledge, but the skill achieves higher $\rho$ by eliminating rhetorical scaffolding, inlining metadata, and structuring content for direct agent consumption.

Maximizing knowledge density is the design objective that unifies the three stages of the pipeline. Codification identifies the knowledge that must be preserved; compression maximizes $\rho$ by restructuring the representation; injection ensures that only high-$\rho$ artifacts reach the agent's context window. Empirical evidence supports the magnitude of the effect: Ulan uulu et al.~\cite{ulanuulu2026codified} report a 206\% quality improvement when LLM agents are augmented with codified expert domain knowledge in an industrial case study---a gain attributable not to model improvement but to the quality and structure of the injected knowledge---consistent with the knowledge density principle proposed in this paper.

\subsection{Mapping to the SECI Model}

Nonaka and Takeuchi's SECI model~\cite{nonaka1995} describes organizational knowledge creation as a spiral through four modes: Socialization (tacit to tacit), Externalization (tacit to explicit), Combination (explicit to explicit), and Internalization (explicit to tacit). The Knowledge Activation framework maps onto this model, with adaptations for the machine consumer.

\paragraph{Socialization $\rightarrow$ Expert Observation.}
In the human SECI model, socialization is the transfer of tacit knowledge through shared experience. In Knowledge Activation, this corresponds to the initial phase of codification: observing expert behavior, shadowing senior engineers during incident response, and analyzing the unwritten practices that teams follow. The knowledge remains tacit at this stage but is identified for subsequent extraction.

\paragraph{Externalization $\rightarrow$ Codification.}
Externalization---the conversion of tacit knowledge to explicit form---maps directly to the codification stage. The critical difference is the target format: rather than producing narrative documentation (the traditional output of externalization), Knowledge Activation produces structured, metadata-rich representations oriented toward machine consumption.

\paragraph{Combination $\rightarrow$ Compression.}
Combination in the SECI model involves synthesizing multiple explicit knowledge sources into new explicit knowledge. In Knowledge Activation, the compression stage performs an analogous synthesis: combining information from multiple codified sources (API documentation, policy documents, architectural decision records) into a single, dense Atomic Knowledge Unit.

\paragraph{Internalization $\rightarrow$ Agent Execution.}
Internalization---the conversion of explicit knowledge back into tacit knowledge through practice---finds its analog in agent execution. When an agent receives an AKU via injection and successfully completes the specified task, the agent has, in effect, ``internalized'' the organizational knowledge for the duration of that invocation. Unlike human internalization, this process is ephemeral: the agent does not retain the knowledge across invocations (absent explicit memory mechanisms), and the AKU must be re-injected for future tasks. This ephemerality reinforces the importance of efficient injection mechanisms.

\paragraph{Algorithmic Externalization.}
The mapping reveals a specialization that Nonaka and Takeuchi's original framework did
not anticipate: the conversion of tacit institutional knowledge into machine-executable
form through deliberate, structured codification. This process---which this paper terms
\emph{algorithmic externalization}---specializes traditional externalization for a class
of consumer the SECI model could not have foreseen: autonomous agents operating under the
constraints of the Context Window Economy. The practical differences are substantial:
the target format is structured and governance-annotated rather than narrative; the
optimization criterion is knowledge density rather than human readability; and the output
carries deterministic governance annotations---validators, permission scopes, blast radius
declarations---that have no analog in human-oriented knowledge artifacts. These
differences warrant treating algorithmic externalization as a distinct specialization
within the SECI framework rather than a minor variation of conventional externalization.

\subsection{Formal Definition}

\begin{definition}[Knowledge Activation]
\label{def:knowledge-activation}
\textbf{Knowledge Activation} is the end-to-end process of transforming latent institutional knowledge into agent-executable form through three stages---codification, compression, and injection---such that the resulting Atomic Knowledge Units maximize knowledge density $\rho(k, \tau)$ while satisfying the token budget, attention decay, and latency cost constraints of the Context Window Economy. Knowledge Activation is distinguished from knowledge retrieval by its output: not informational passages for interpretation, but action-ready specifications whose seven-component structure (Section~\ref{sec:atomic-knowledge-units}) includes procedural guidance, tool bindings, organizational metadata, governance constraints, continuation paths, and validators.
\end{definition}

The Knowledge Activation framework establishes the theoretical foundation; the following section specifies the concrete artifact it produces---the Atomic Knowledge Unit, instantiated as a \emph{skill}---and presents its formal structure, design principles, and compositional properties.

\section{Atomic Knowledge Units: The Design of Skills}
\label{sec:atomic-knowledge-units}

The Knowledge Activation framework (Section~\ref{sec:knowledge-activation}) produces a specific kind of artifact: the \emph{Atomic Knowledge Unit} (AKU). This section positions the AKU within the evolution of agent-consumable knowledge, provides a formal definition, specifies the components of the AKU schema, articulates design principles, provides concrete examples, and discusses how AKUs compose into complex workflows.

\subsection{From Rules to AI Skills to Atomic Knowledge Units}

The term \emph{skill} has evolved through three distinct meanings relevant to this work. In its general sense, a skill denotes a learned, reusable capability---what a human acquires through practice and experience. In reinforcement learning, ``skills'' or ``options''~\cite{wang2023voyager} refer to temporally extended actions with learned policies. This paper uses neither meaning. Instead, it builds on a third sense that emerged from the convergent evolution of agent-consumable knowledge artifacts.

Before the formalization of the Agent Skills open standard~\cite{anthropic2025agentskills, agentskills2025spec}, multiple platforms independently developed ad hoc mechanisms for delivering structured knowledge to agents: Cursor rules, GitHub Copilot custom instructions, Windsurf rules, and repository-level context files such as \texttt{AGENTS.md}. Each format evolved to serve the same function---providing agents with contextual guidance beyond their parametric knowledge---but none offered portability, standardization, or a formal schema. The Agent Skills specification, released by Anthropic in December 2025 and subsequently adopted across major AI platforms~\cite{microsoft2025platformagentic}, standardized what these precursors were reaching toward: a portable, discoverable format for packaging agent-consumable knowledge as \emph{AI Skills}.

However, most AI Skills deployed in practice are general-purpose agent instructions: coding conventions, tool preferences, repository guidelines. As documented in Section~\ref{sec:introduction}, the vast majority of deployed AI Skills lack governance and security constraints---the properties that enterprise institutional knowledge delivery demands. The Atomic Knowledge Unit is the specialization that addresses this gap.

\begin{definition}[Atomic Knowledge Unit]
\label{def:aku}
An \textbf{Atomic Knowledge Unit (AKU)} is an AI Skill specialized for institutional knowledge delivery. It is the minimal self-contained bundle of intent, procedural knowledge, tool bindings, organizational metadata, governance constraints, continuation paths, and validators that enables an autonomous agent to execute a single coherent action within an enterprise software development context. An AKU is \emph{atomic} in the sense that it represents an indivisible unit of institutional knowledge: removing any component degrades the agent's ability to act correctly, and the unit addresses exactly one coherent action rather than a collection of loosely related tasks.
\end{definition}

The AKU specializes the AI Skill primitive along four dimensions. \emph{Governance constraints} embed permission boundaries and compliance requirements, which the standard does not address. \emph{Continuation paths} connect isolated skills into navigable workflows, for which the standard has no equivalent. \emph{Validators}---deterministic pre-execution, post-execution, and invariant scripts---enforce governance programmatically, extending the standard's \texttt{scripts/} directory into a structured verification mechanism. \emph{Organizational metadata} formalizes the standard's generic \texttt{metadata} key-value field into a specific enterprise vocabulary: team ownership, service tier, SLA, environment classification, on-call rotation, and cost center attribution. These additions transform a general-purpose AI Skill into an institutional knowledge primitive---one whose role is analogous to the instruction in a processor's instruction set architecture: a well-defined, self-contained unit that can be invoked independently and composed with other units to accomplish higher-order tasks. In the knowledge management literature, Davenport and Prusak~\cite{davenport1998} identified that actionable knowledge requires context, experience, and a framework for evaluation; the AKU formalizes this insight for the machine consumer by bundling these elements into a single, context-window-efficient artifact.

\subsection{The Skill Schema}

The AKU schema specifies seven components, each serving a distinct function in enabling institutionally grounded agent action. Table~\ref{tab:skill-schema} provides an overview, Figure~\ref{fig:skill-schema} illustrates their relationships, and the subsections that follow elaborate on each component.

\begin{table}[ht]
\centering
\caption{The seven components of the Skill schema.}
\label{tab:skill-schema}
\begin{tabularx}{\textwidth}{lX}
\toprule
\textbf{Component} & \textbf{Purpose} \\
\midrule
Intent Declaration & Specifies what the skill accomplishes and under what conditions it should be activated \\
Procedural Knowledge & Provides step-by-step guidance, constraints, patterns, and anti-patterns for execution \\
Tool Bindings & Identifies which tools, APIs, or platform capabilities the agent should invoke \\
Organizational Metadata & Encodes team ownership, service catalog references, and environment-specific context \\
Governance Constraints & Defines permission boundaries, approval workflows, and blast radius limits \\
Continuation Paths & Specifies what to do next: related skills, escalation paths, and fallback strategies \\
Validators & Deterministic scripts that verify preconditions, postconditions, and invariants for automated governance \\
\bottomrule
\end{tabularx}
\end{table}

\subsubsection{Intent Declaration}

The intent declaration maps to and extends the standard's \texttt{name} and \texttt{description} frontmatter fields, which enable progressive discovery at approximately 100 tokens per skill. The AKU adds structured trigger conditions beyond the natural-language description, enabling the injection mechanism to select skills with high precision and minimizing context pollution. The natural language component enables the agent to reason about whether the skill is appropriate for a given task; the structured conditions enable programmatic matching.

A well-crafted intent declaration balances specificity and generality. Too narrow a declaration may prevent the skill from being activated when it is relevant; too broad a declaration may cause it to be injected into contexts where it adds noise without value. This balance directly affects the knowledge density $\rho(k, \tau)$ defined in Equation~\ref{eq:knowledge-density}: an overly broad skill wastes tokens when injected into irrelevant contexts, while an overly narrow skill may fail to be injected when needed, reducing the system's overall task completion rate.

\subsubsection{Procedural Knowledge}

The procedural knowledge component is the core of the skill: the step-by-step guidance that the agent follows to accomplish the declared intent. This component encodes not only the ``happy path'' but also constraints (what the agent must not do), patterns (preferred approaches that reflect organizational standards), and anti-patterns (known failure modes to avoid).

Effective procedural knowledge is structured for the attention characteristics of transformer architectures. Critical constraints and preconditions are placed early in the component, where they are most likely to influence model behavior~\cite{liu2024lostinmiddle}. Steps are ordered logically and numbered explicitly, reducing the interpretive burden on the agent. References to external resources are replaced with inline summaries wherever possible, avoiding the need for the agent to perform additional retrieval.

The procedural component distinguishes skills from the document chunks produced by RAG systems~\cite{lewis2020rag}. A RAG chunk might contain a paragraph describing a deployment process; a skill's procedural knowledge provides an ordered, constraint-annotated action sequence that the agent can execute directly.

\subsubsection{Tool Bindings}

Autonomous agents operate on the world through tools---APIs, command-line interfaces, platform services, and external integrations~\cite{schick2023toolformer, qin2023toolllm}. The tool bindings component of a skill specifies \emph{which} tools the agent should use to execute the procedural steps and \emph{how} they should be invoked. This includes function signatures, required parameters, authentication mechanisms, and expected response formats.

Tool bindings address a fundamental gap in retrieval-based approaches. A RAG system may return a passage that mentions an API endpoint, but the passage does not specify the exact invocation pattern, the required headers, or the error handling strategy. A skill's tool bindings provide this information in a structured format that the agent can act upon directly, reducing the probability of tool invocation errors and eliminating the need for the agent to infer tool usage from underspecified prose.

In the context of emerging standards such as the Model Context Protocol~\cite{anthropic2024mcp}, tool bindings serve as the bridge between the knowledge layer (what the agent should do) and the capability layer (what the agent can do). Each tool binding maps a procedural step to a specific MCP tool or API endpoint, ensuring that the skill's guidance is grounded in the agent's actual capabilities.

\subsubsection{Organizational Metadata}

Enterprise software development is situated within complex organizational structures: teams own services, services run in specific environments, and actions have consequences that propagate through dependency graphs. The organizational metadata component encodes this context within the skill, ensuring that the agent acts not in the abstract but within the correct organizational frame.

Organizational metadata includes team ownership (which team maintains the affected service), service catalog references (linking the skill to the organization's service registry), environment specifications (production, staging, development), dependency mappings (upstream and downstream services affected by the action), and communication channels (where to report outcomes or request assistance).

This component reflects a key insight from organizational knowledge management~\cite{argote2000knowledge}: knowledge is most effective when it is embedded in its organizational context. A deployment procedure that omits the identity of the owning team, the target environment, or the affected downstream services is incomplete knowledge---sufficient, perhaps, for a human who can infer context from experience, but insufficient for an autonomous agent that lacks organizational situational awareness.

\subsubsection{Governance Constraints}

In enterprise environments, the correctness of an action is necessary but not sufficient; actions must also be \emph{permissible}. The governance constraints component encodes the permission boundaries, approval workflows, compliance requirements, and blast radius limits that govern the skill's execution.

Governance constraints are specified declaratively within the skill rather than enforced by an external policy engine alone. This design choice---governance by design, not governance by interception---has two advantages. First, it ensures that the agent is aware of constraints before attempting an action, reducing the incidence of blocked operations and wasted computation. Second, it enables the agent to incorporate governance considerations into its planning process, selecting alternative approaches when the preferred path requires permissions the agent does not possess.

Examples of governance constraints include: role-based access requirements (the agent must possess a specific role to execute the skill), approval gates (human approval is required before proceeding past a specified step), change window restrictions (the skill may only be executed during designated maintenance windows), blast radius declarations (the maximum scope of impact the skill may produce), and compliance annotations (references to regulatory requirements such as SOC 2 controls or GDPR obligations). The NIST AI Risk Management Framework~\cite{nist2024ai} emphasizes the importance of embedding governance into AI systems from the design phase; the governance constraints component of the skill schema operationalizes this principle at the knowledge artifact level.

\subsubsection{Continuation Paths}

Tasks in enterprise software development rarely exist in isolation. A deployment may be followed by a verification step; a failed health check may require a rollback; an incident may escalate from automated remediation to human intervention. The continuation paths component specifies what happens after the skill's primary action completes, encoding the workflow topology in which the skill is situated.

Continuation paths include three types of links: \emph{success continuations} (which skill(s) to activate upon successful completion), \emph{failure continuations} (fallback strategies and rollback skills to invoke upon failure), and \emph{escalation paths} (conditions under which control should be transferred to a human operator or a more privileged agent). These links transform a collection of isolated skills into a navigable knowledge graph, enabling agents to traverse multi-step workflows without requiring a monolithic prompt that encodes the entire workflow in advance.

Continuation paths also serve a governance function: they constrain the agent's action space to organizationally sanctioned next steps, reducing the risk of autonomous agents improvising actions that deviate from established procedures.

\paragraph{Interoperability.}
A critical property of well-designed skills is \emph{interoperability}: each skill is not an isolated artifact but a node
in a navigable knowledge graph. The continuation component encodes routing
logic---given a successful outcome, proceed to skill X; given failure mode A, route to
the diagnostic skill; given an ambiguous state requiring escalation, route through the
governance gradient. This distributed routing transforms the AKU Registry from a
library of isolated procedures into a self-navigating process graph. The agent need not
maintain a global model of the workflow because each skill contains sufficient
information to direct the agent toward the next relevant
skill. Section~\ref{sec:golden-path-architecture} develops the architectural
consequences of this property: the Knowledge Topology is not a catalog of available
skills but a \emph{routing graph} that defines which skills reference which, what valid
traversal sequences exist, and how iteration loops are bounded.

Interoperability is also what converts context rot prevention from an individual-skill
property into a system property. Without continuation routing, unstructured
iteration---attempts accumulating in the context window without direction---remains the
failure mode even with well-authored individual skills. With continuation routing,
failure reaches a known junction with its own structured guidance; iteration becomes
traversal of the knowledge graph rather than context-free retry.

\begin{figure}[htbp]
  \centering
  \includegraphics[width=0.85\textwidth]{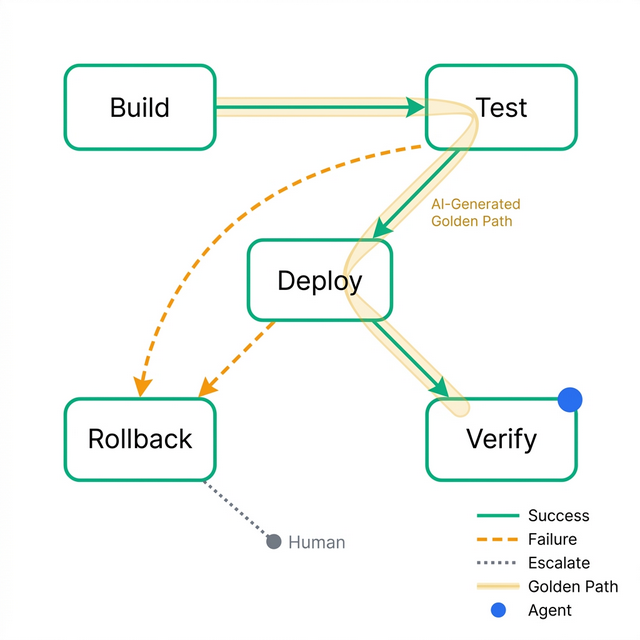}
  \caption{The interoperable skill network. Skills are not isolated artifacts but nodes in a self-navigating process graph. Each skill's continuation metadata encodes routing to neighboring skills: solid arrows represent success continuations, dashed arrows represent failure continuations, and dotted lines represent escalation to human operators. The blue dot indicates the agent's current position. The routing intelligence is distributed across the network---each skill knows its neighbors---enabling structured iteration without a central orchestrator and preventing context rot by ensuring that failure reaches a known junction with structured guidance rather than a context-free retry.}
  \label{fig:aku-network}
\end{figure}

\begin{figure}[htbp]
  \centering
  \includegraphics[width=0.85\textwidth]{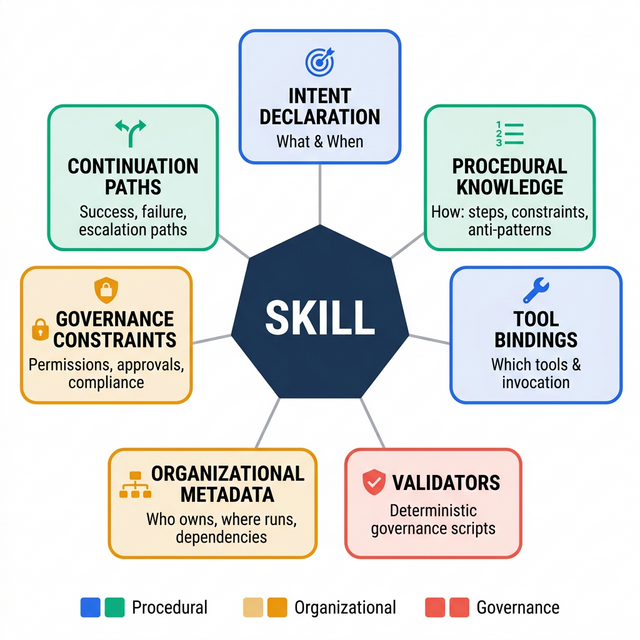}
  \caption{The seven-component Skill schema. Each Skill bundles intent, procedural knowledge, and tool bindings (the ``how'') with organizational metadata, governance constraints, validators, and continuation paths (the ``where, who, and what next''), forming a composable Atomic Knowledge Unit.}
  \label{fig:skill-schema}
\end{figure}

\subsection{Design Principles for Effective Skills}

The skill schema defines the \emph{structure} of an AKU; the following design principles govern its \emph{quality}. These principles are derived from the constraints of the Context Window Economy and the requirements of enterprise agent deployment.

\paragraph{Atomicity.}
Each skill encodes exactly one coherent action. A skill for deploying a microservice does not also cover database migration or DNS configuration; those are separate skills that may be composed into a workflow via continuation paths. Atomicity ensures that skills are independently testable, independently versionable, and injected only when their specific action is required---avoiding the token waste of bundling unrelated guidance into a single artifact.

\paragraph{Context Efficiency.}
Every token in a skill must earn its place. Redundant prose, boilerplate preambles, and information available from the agent's base training are eliminated. Context efficiency maximizes the knowledge density $\rho(k, \tau)$ of the skill, directly addressing the token budget constraint of the Context Window Economy. Research on prompt compression~\cite{jiang2023longllmlingua} and the finding that alignment quality can be achieved with minimal but high-quality data~\cite{zhou2023lima} both support the principle that less, when carefully curated, is more.

\paragraph{Trigger Precision.}
A skill's intent declaration must be sufficiently precise to avoid false-positive injection. A skill activated in the wrong context imposes a double cost: it consumes tokens without contributing value, and it may mislead the agent into taking inappropriate action. Trigger precision is the injection-stage complement of context efficiency: where context efficiency minimizes cost per skill, trigger precision minimizes the number of irrelevant skills delivered.

\paragraph{Organizational Grounding.}
Skills carry the ``where'' and ``who'' alongside the ``how.'' A procedurally correct skill that omits organizational context---deploying to the wrong environment, notifying the wrong team, or ignoring a service dependency---fails in practice despite succeeding in isolation. Organizational grounding ensures that skills reflect the situated nature of enterprise knowledge~\cite{alavi2001}.

\paragraph{Governance by Design.}
Constraints are embedded within the skill, not enforced solely by external guardrails. This principle ensures that governance is portable (the skill carries its constraints wherever it is injected), transparent (the agent can reason about constraints during planning), and composable (constraints from multiple skills can be checked for consistency before execution begins).

\subsection{Differentiation from Existing Knowledge Artifacts}

The skill primitive shares surface similarities with several existing knowledge and automation artifacts. Table~\ref{tab:skill-comparison} clarifies the distinction by comparing skills against these constructs across the seven schema dimensions.

\begin{table}[ht]
\centering
\caption{Skills compared to existing knowledge and automation artifacts.}
\label{tab:skill-comparison}
\begin{tabularx}{\textwidth}{lXXXXX}
\toprule
\textbf{Dimension} & \textbf{Skill (AKU)} & \textbf{Runbook} & \textbf{SOP} & \textbf{Ansible Playbook} & \textbf{Backstage Template} \\
\midrule
Intent declaration & Machine-matchable trigger & Title/summary & Title/scope & Play name & Template name \\
Procedural knowledge & Token-optimized for agent & Prose for humans & Prose for humans & Declarative YAML & Scaffolding steps \\
Tool bindings & Explicit, typed & Implicit or absent & Absent & Built-in modules & Plugin hooks \\
Org.\ metadata & Team, tier, SLA, catalog & Varies & Varies & Inventory groups & Catalog entity \\
Governance & Embedded constraints & Absent & Approval gates & Limited (tags) & Absent \\
Validators & Deterministic scripts & Absent & Absent & Handlers (partial) & Absent \\
Continuation paths & Skill-to-skill links & ``See also'' & ``Next steps'' & Role dependencies & Absent \\
\midrule
Target consumer & Autonomous agent & Human operator & Human operator & Automation engine & Human developer \\
Context-optimized & Yes & No & No & N/A & No \\
\bottomrule
\end{tabularx}
\end{table}

The critical differentiators are threefold. First, skills are \emph{agent-oriented}: their procedural content is structured for LLM consumption under context window constraints, not for human reading. Second, skills embed \emph{governance as a first-class property}, including validators that provide deterministic verification without human approval. Third, skills carry \emph{continuation paths} that enable compositional workflows---a capability absent from all four comparators. Runbooks and SOPs encode similar procedural knowledge but in formats optimized for human interpretation; Ansible playbooks automate execution but lack organizational metadata and governance constraints; Backstage templates scaffold projects but do not encode the institutional knowledge required for ongoing operational tasks.

\paragraph{What an AKU is \emph{not}.}
For practitioners encountering related primitives in the contemporary
agent ecosystem, the AKU is distinct from several adjacent constructs:
\begin{itemize}
  \item \textbf{AKU $\neq$ RAG passage.} Retrieval delivers semantically
    similar text fragments for the agent to interpret; an AKU delivers
    an action-ready specification with explicit governance metadata and
    continuation paths to downstream skills.
  \item \textbf{AKU $\neq$ MCP tool definition.} A Model Context
    Protocol tool describes \emph{what} an agent can call (interface,
    arguments, return type); an AKU specifies \emph{when, why, and under
    what constraints} that tool should be invoked, plus the procedure
    that composes multiple tools into a coherent action.
  \item \textbf{AKU $\neq$ AGENTS.md file.} An AGENTS.md is a
    repository-level context file informing the agent's behavior
    throughout a session~\cite{liu2026agentsmd}; an AKU is an atomic
    unit loaded on demand by the activation policy at the moment of
    need.
  \item \textbf{AKU $\neq$ Backstage golden path template.} A template
    in an Internal Developer Platform is a pre-authored deterministic
    workflow a human developer executes; an AI-Generated Golden Path
    emerges from AKU composition at agent runtime and adapts to the
    specific request.
\end{itemize}

\subsection{Validators: Deterministic Governance Scripts}

While governance constraints (Section~\ref{sec:atomic-knowledge-units}) declare the rules an agent must follow, \emph{validators} enforce them automatically. A validator is a deterministic script embedded within a skill that verifies whether an agent's actions meet organizational standards---without requiring human approval for the verification itself. Validators transform governance from a declarative annotation into an executable guarantee.

Three categories of validators address different points in the execution lifecycle:

\emph{Pre-execution validators} verify preconditions before the agent acts. Examples include confirming that the change management window is open, that the requesting identity holds the required permissions, that all continuous integration checks have passed, and that the target service is registered in the organization's service catalog.

\emph{Post-execution validators} verify outcomes after the action completes. These may confirm that a deployed service is healthy, that no policy violations were introduced, and that rollback capability has been established for the new deployment.

\emph{Invariant validators} monitor conditions continuously during execution. An invariant validator might enforce that the blast radius remains within declared limits or that resource consumption stays within the allocated budget throughout a multi-step operation.

Critically, validators are code, not prose. They are implemented as shell scripts, Python checks, or policy-as-code frameworks such as Open Policy Agent. They are version-controlled alongside the skills they govern, independently testable in isolation, and produce deterministic pass/fail results with structured audit logs suitable for compliance review.

The composability of validators follows a type-safety analogy: just as well-typed functions compose into type-safe programs, skills with comprehensive validators compose into governance-safe workflows. When an agent composes an AI-Generated Golden Path---a workflow assembled at runtime from available skills---the composed path inherits the union of all constituent validators. Governance safety is achieved by construction rather than by post-hoc review.

Validators shift the governance team's operational model from \emph{governance-as-approval}---reviewing individual agent actions as they occur---to \emph{governance-as-code}---authoring, testing, and maintaining deterministic validation scripts. This shift is analogous to the Infrastructure-as-Code transformation that freed operations teams from ticket-based provisioning: the governance team's mission becomes increasing validator coverage across the skill library, progressively moving more skills toward full autonomy as validator coverage expands.

Validators are what make the Governance Gradient (Section~\ref{sec:governance}) operationally tractable: the degree of validator coverage determines a skill's position on the gradient. A skill with comprehensive pre-execution, post-execution, and invariant validators can operate with full autonomy; a skill with partial validator coverage requires human oversight for the unvalidated aspects; a skill with no validators defaults to human-in-the-loop execution.

\subsection{Example: Deploy Microservice Skill}

To illustrate the skill schema concretely, consider a ``Deploy Microservice'' skill for an enterprise platform. Figure~\ref{fig:deploy-skill-example} presents an abbreviated representation.

\begin{figure}[ht]
\begin{tcolorbox}[title=Skill: Deploy Microservice to Production, fonttitle=\bfseries\small, fontupper=\small]
\textbf{Intent:} Deploy a specified microservice to the production Kubernetes cluster, triggered when a deployment request is received for a service registered in the internal service catalog.

\textbf{Procedure:}
\begin{enumerate}\setlength{\itemsep}{0pt}
    \item Verify the service is registered in the service catalog and the requesting agent/user has deployment permission.
    \item Confirm that all CI checks (unit tests, integration tests, security scan) have passed for the target artifact version.
    \item Check the change management calendar; abort if outside the approved change window.
    \item Execute a canary deployment to 5\% of traffic using the platform's deployment API.
    \item Monitor error rate and latency for 10 minutes; if either exceeds the threshold, invoke the Rollback Deployment skill.
    \item Promote to 100\% traffic upon successful canary validation.
    \item Notify the owning team via the configured notification channel.
\end{enumerate}

\textbf{Anti-patterns:} Do not deploy directly to 100\% traffic. Do not skip the canary phase, even if the change appears minor.

\textbf{Tool Bindings:} \texttt{service-catalog/lookup}, \texttt{ci-pipeline/status}, \texttt{change-mgmt/check-window}, \texttt{k8s-deploy/canary}, \texttt{monitoring/query-metrics}, \texttt{k8s-deploy/promote}, \texttt{notifications/send}.

\textbf{Organizational Metadata:} Owner: service owning team (resolved from catalog). Environment: production. Downstream dependencies: resolved from service graph.

\textbf{Governance:} Requires \texttt{deployer} role. Human approval required for services classified as Tier-1. Change window: weekdays 09:00--16:00 UTC. Blast radius: single service (no cascading deployments).

\textbf{Validators:} \texttt{pre:check-change-window.sh}, \texttt{pre:verify-ci-green.sh}, \texttt{post:health-check.sh}, \texttt{post:rollback-capability.sh}, \texttt{invariant:blast-radius-monitor.sh}.

\textbf{Continuations:} On success $\rightarrow$ Post-Deployment Verification skill. On failure $\rightarrow$ Rollback Deployment skill. On permission denied $\rightarrow$ escalate to team lead.
\end{tcolorbox}
\caption{Abbreviated example of a ``Deploy Microservice'' skill illustrating all seven schema components.}
\label{fig:deploy-skill-example}
\end{figure}

This example demonstrates several properties. The skill is self-contained: an agent receiving only this skill and the relevant tool definitions has sufficient information to execute the deployment. It is atomic: it covers deployment and only deployment, delegating rollback and verification to separate skills via continuation paths. It embeds governance: the agent knows the required role, the change window restriction, and the human approval gate before taking any action. And it is context-efficient: the entire specification occupies a fraction of the tokens that an equivalent narrative runbook would require.

\begin{figure}[ht]
\begin{tcolorbox}[title=Skill: Respond to Production Incident, fonttitle=\bfseries\small, fontupper=\small]
\textbf{Intent:} Initiate structured incident response when a production alert is triggered for a service classified as Tier-1 or Tier-2 in the service catalog.

\textbf{Procedure:}
\begin{enumerate}\setlength{\itemsep}{0pt}
    \item Acknowledge the alert in the monitoring system within 2 minutes.
    \item Query the service catalog to identify the owning team, on-call engineer, and service dependencies.
    \item Open an incident channel in the team's communication platform with a standardized naming convention.
    \item Gather initial diagnostics: error rates, latency percentiles, recent deployments, and infrastructure health.
    \item Post a structured incident summary to the channel using the organization's incident template.
    \item If the incident correlates with a recent deployment, invoke the Rollback Deployment skill.
    \item Escalate to the on-call engineer if automated diagnostics do not identify a root cause within 10 minutes.
\end{enumerate}

\textbf{Anti-patterns:} Do not attempt fixes before gathering diagnostics. Do not communicate outside the incident channel during active response.

\textbf{Tool Bindings:} \texttt{monitoring/acknowledge-alert}, \texttt{service-catalog/lookup}, \texttt{comms/create-channel}, \texttt{monitoring/query-metrics}, \texttt{deploy-history/recent}, \texttt{comms/post-message}.

\textbf{Organizational Metadata:} Owner: resolved from service catalog. Environment: production. Escalation: on-call rotation from PagerDuty.

\textbf{Governance:} Requires \texttt{incident-responder} role. All actions logged to incident management system. Post-incident review required within 48 hours.

\textbf{Validators:} \texttt{pre:verify-alert-active.sh}, \texttt{pre:check-responder-role.sh}, \texttt{post:verify-channel-created.sh}, \texttt{post:verify-summary-posted.sh}, \texttt{invariant:escalation-timer.sh}.

\textbf{Continuations:} On resolution $\rightarrow$ Post-Incident Review skill. On escalation timeout $\rightarrow$ notify engineering manager. On correlated deployment $\rightarrow$ Rollback Deployment skill.
\end{tcolorbox}
\caption{Abbreviated example of a ``Respond to Production Incident'' skill, demonstrating the schema's applicability beyond deployment workflows.}
\label{fig:incident-skill-example}
\end{figure}

This second example illustrates the schema's generality beyond deployment. The incident response skill operates in a fundamentally different domain---reactive diagnostics rather than proactive provisioning---yet the seven-component structure accommodates it without modification. The organizational metadata and validators differ in content but not in kind, confirming that the skill schema is domain-agnostic within enterprise software development.

\subsection{The AKU as AI Skill Specialization}

As established in Section~\ref{sec:knowledge-activation}, the AKU builds on the Agent Skills open standard~\cite{anthropic2025agentskills, agentskills2025spec}. The base specification defines a portable format for packaging agent-consumable knowledge as AI Skills---portable packages of instructions, scripts, and resources that agents discover through progressive disclosure and consume through context injection. The structural correspondence between the AI Skill format and the AKU schema is direct: intent declarations correspond to trigger descriptions that govern skill discovery, and procedural knowledge maps to step-by-step instructional content.

The contribution of this paper is not the skill primitive itself---which the open standard defines---but the \emph{enterprise specialization} and the \emph{surrounding architecture}. The AKU extends the base AI Skill with governance constraints, validators, continuation paths, and organizational metadata---the components that the governance gap documented in Section~\ref{sec:introduction} reveals are absent from the vast majority of deployed AI Skills. The Context Window Economy explains \emph{why} skills must be compact. Knowledge Activation explains \emph{how} to produce them through codification, compression, and injection. The Governance architecture (Section~\ref{sec:governance}) explains how to \emph{govern} them at enterprise scale.

The strategic implications of this LLM-agnosticism---including provider independence and knowledge portability---are discussed in Section~\ref{sec:discussion}.

\subsection{Skill Composition}

Individual skills are atomic by design, but enterprise workflows are inherently composite. A release process may involve building an artifact, running security scans, deploying to staging, executing integration tests, deploying to production, and notifying stakeholders---each a distinct skill, composed into a coherent sequence.

Skill composition operates through two mechanisms. The first is \emph{continuation-driven chaining}: each skill's continuation paths define the edges of a directed workflow graph, and the agent traverses this graph by following success, failure, or escalation continuations. This mechanism supports linear workflows, branching (on success versus failure), and convergence (multiple paths leading to a common verification step).

The second mechanism is \emph{task-driven selection}: given a high-level task (e.g., ``release version 2.3.1 of the payments service''), the agent---or an orchestration layer---decomposes the task into subtasks and selects the appropriate skill for each subtask. This mechanism draws on the decomposed prompting paradigm~\cite{khot2023decomposed} and least-to-most prompting~\cite{zhou2023leasttomostprompting}, extending these techniques from prompt-level decomposition to knowledge-level decomposition.

The analogy to function composition in programming is deliberate and instructive. Just as functions in a well-designed program are individually testable, have well-defined interfaces (parameters and return types), and compose through calling conventions, skills are individually validatable, have well-defined schemas (the seven components), and compose through continuation paths and task decomposition. This analogy suggests that the software engineering principles of cohesion (each function does one thing), coupling (functions interact through narrow interfaces), and abstraction (implementation details are hidden behind interfaces) apply equally to skill design---reinforcing the design principles of atomicity, organizational grounding, and governance by design articulated above.

The compositional nature of skills also has implications for the Context Window Economy. Because each skill is loaded only when its corresponding subtask is active, the agent's context window at any given moment contains only the skill(s) relevant to the current step---not the entire workflow. This just-in-time injection pattern amortizes the token cost of complex workflows across multiple agent invocations, enabling arbitrarily long workflows to be executed within fixed context budgets. The workflow's total knowledge exceeds the context window capacity; the skill architecture ensures that the window need only contain the current action's knowledge at any given time.

\section{From Golden Paths to Agent-Navigable Knowledge Topologies}
\label{sec:golden-path-architecture}

The concept of \emph{golden paths} has emerged as a central tenet of platform engineering: opinionated, organization-supported workflows that reduce cognitive load by guiding developers toward well-tested defaults~\cite{skelton2019team, beetz2023platform}. A golden path encodes the organization's preferred way to accomplish a task---creating a service, deploying to production, onboarding a new team member---and presents it as a paved, low-friction route through otherwise complex toolchains. The Cloud Native Computing Foundation's Platforms White Paper identifies golden paths as a mechanism for balancing developer autonomy with organizational consistency~\cite{cncf2024platforms}. When golden paths are well-maintained, they serve as institutional knowledge made navigable: the accumulated wisdom of platform teams distilled into repeatable workflows.

The cognitive load that golden paths address is familiar to anyone who has navigated a large organization: individually sensible rules compose into processes far more intricate than any single rule would suggest, and a missed dependency means retracing the path from the beginning. Golden paths encode a knowledgeable guide's understanding of which step comes first, which prerequisites apply, and which requirements are context-dependent. The challenge is encoding that guide's knowledge in a form that scales.

This is equally the challenge that large language models face when operating within complex organizations. An LLM can reason, plan, and act---but only over what is present in its context window. Flood it with everything the organization knows and critical details drown in noise; provide too little and the model improvises, violating conventions it was never told about. This is the core problem of \emph{context engineering}---the emerging discipline concerned with managing what information reaches an agent's context window~\cite{mei2025survey, anthropic2025contextengineering}. AKUs, as defined in this paper, are context engineering applied to institutional knowledge---activating the precise knowledge the agent needs for each step, at the moment it becomes relevant.

However, existing golden paths are fundamentally deterministic flowcharts. Whether consumed by humans through portal UIs and documentation pages, or by deterministic orchestrators through the APIs and command-line interfaces that platform teams also provide, the underlying structure is the same: a fixed sequence of steps executed without adaptation. The arrival of autonomous AI agents as first-class participants in enterprise software development enables a fundamentally different mode of traversal. Rather than executing a rigid flowchart, an AI agent walks through every step of a golden path while adaptively adjusting the plan as it progresses toward its goal---responding to intermediate outcomes, contextual signals, and edge cases that no pre-composed template could anticipate. This section introduces \emph{Agent Knowledge Architecture}: the structural infrastructure that enables agents to traverse institutional knowledge adaptively, built on the AKU primitive defined in Section~\ref{sec:atomic-knowledge-units}.

\subsection{The Limitations of Deterministic Workflow Templates}

Prior to the current generation of AI agents, golden paths were implemented as deterministic workflow templates: modular steps hard-wired into fixed sequences, typically realized as portal wizards, Backstage scaffolders, CI/CD pipeline templates, or Terraform modules. These templates encode a specific sequence of actions for a specific use case---``create a Java microservice,'' ``provision a PostgreSQL database,'' ``onboard a new engineer''---and expose that sequence through a human-navigable interface. While deterministic templates have delivered substantial value, they exhibit four structural limitations that constrain their effectiveness at organizational scale.

First, deterministic workflow templates are \emph{resource-intensive to compose}. Each golden path requires platform engineers to design, build, test, integrate, and maintain every step, every branching condition, and every edge case. Organizations with hundreds of services and dozens of workflow patterns face a combinatorial authoring burden that grows faster than platform teams can staff. The cost of golden path creation is not merely the initial engineering effort but the ongoing maintenance as underlying systems evolve.

Second, deterministic templates are \emph{brittle under change}. They assume stable underlying systems---fixed API contracts, unchanging compliance requirements, static team structures. When an upstream API changes, a new regulatory obligation is introduced, or a team reorganizes, the workflow breaks and must be manually updated. The maintenance burden compounds over time: each golden path is a liability as well as an asset, and the total maintenance cost grows linearly with the template library.

Third, deterministic templates are \emph{task-specific, not flexible}. A golden path for ``deploy Java microservice to Kubernetes'' does not help with ``deploy Python Lambda function'' or ``migrate PostgreSQL to Aurora.'' Each variation demands a new template, authored from near-scratch, because the composition logic is embedded in the template rather than in the reusable components themselves. The result is a proliferation of templates with substantial overlap but limited actual reuse.

Fourth, deterministic templates are \emph{deterministic in a probabilistic world}. Classic golden paths encode a single ``right way'' to accomplish a task. But enterprise software development is deeply contextual---the right approach depends on the team, the service, the environment, the time of day, the current incident status, and dozens of other variables that a deterministic template cannot adapt to. A template that is optimal for one context may be suboptimal or incorrect for another, yet the template has no mechanism to sense or respond to this variation.

These limitations are not failures of execution but structural constraints of the deterministic template paradigm. What is needed is a fundamentally different approach---one in which workflows are not pre-composed but generated at runtime by an intelligent agent drawing on the same institutional knowledge that a senior engineer would apply. The three-layer Agent Knowledge Architecture provides the structural infrastructure for such dynamically generated workflows.

\subsection{Three Layers of Agent Knowledge Architecture}

\begin{figure}[htbp]
  \centering
  \includegraphics[width=\textwidth]{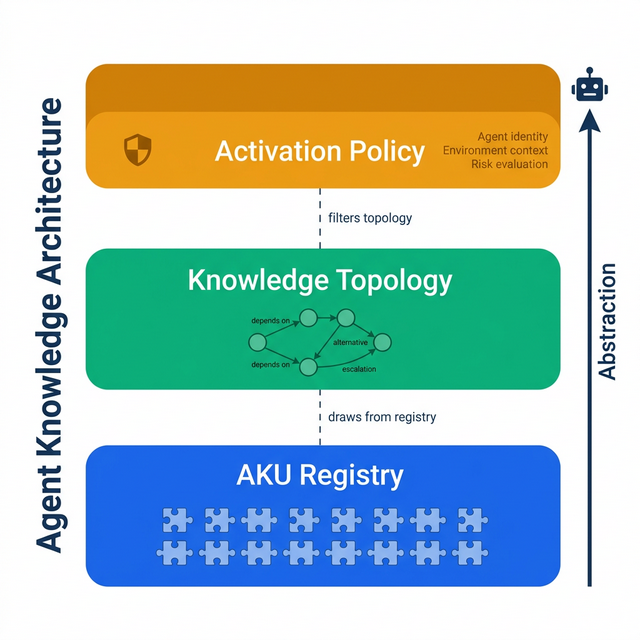}
  \caption{The three-layer Agent Knowledge Architecture. The AKU Registry provides a queryable catalog of Atomic Knowledge Units; the Knowledge Topology captures relationships between AKUs (dependencies, sequences, alternatives); the Activation Policy governs which AKUs are available to which agents under which conditions.}
  \label{fig:golden-path-arch}
\end{figure}

Agent Knowledge Architecture is organized into three compositional layers (Figure~\ref{fig:golden-path-arch}), each of which addresses a distinct aspect of how agents discover, traverse, and execute institutional knowledge.

\paragraph{AKU Registry.}
The foundational layer is the \emph{AKU Registry}: a structured catalog of all available Atomic Knowledge Units within the organization, searchable by intent, domain, capability, and operational context. The AKU Registry serves a function analogous to a service catalog in platform engineering, but its entries are not services---they are units of actionable knowledge. Each registry entry exposes the skill's metadata: its name, description, required inputs, expected outputs, tool dependencies, permission requirements, and governance annotations. Critically, the registry is designed for programmatic consumption. Agents query it not by browsing a web page but by issuing semantic or structured queries that return ranked candidate skills. The registry thus transforms the organization's knowledge surface from a collection of documents into a queryable knowledge API.

\paragraph{Knowledge Topology.}
Above the registry sits the \emph{Knowledge Topology}: the routing graph that connects
skills into a navigable network. The distinction between a registry and a topology is
architecturally significant: a registry answers ``what skills exist?''---sufficient for
retrieval---while a topology answers ``given where the agent is now, and how it arrived,
what comes next?''---necessary for structured process navigation. Skills do not exist in
isolation. A deployment skill depends on a build skill; a database migration skill
requires a schema validation skill as a precondition; a rollback skill serves as the
failure continuation of a deployment skill. The Knowledge Topology captures these
relationships---dependencies, sequences, alternatives, escalation paths, and mutual
exclusions---as first-class graph edges. This representation draws on established work in
knowledge graphs~\cite{hogan2021knowledge}, but applies graph structure specifically to
action-oriented knowledge rather than entity-relationship knowledge. The topology enables
agents to reason about skill composition: given a high-level goal, an agent can traverse
the graph to construct a plan, identify prerequisite skills, and anticipate failure modes
with their corresponding recovery paths. Because each skill's continuation metadata
encodes its edges in the graph (Section~\ref{sec:atomic-knowledge-units}), the routing
intelligence is distributed across the network rather than concentrated in a central
orchestrator---each skill knows its neighbors, and the agent navigates by following
locally available directions rather than consulting a global map.

At enterprise scale, the topology requires well-formedness guarantees that distributed authoring does not automatically provide. Continuation paths authored independently by different teams may introduce cycles (skill A references skill B, which references A), conflicts (two skills both declare themselves as the success continuation of a third), or dangling references (a continuation target that has been deleted or renamed). The platform team's stewardship role (Section~\ref{sec:discussion}) includes maintaining topology consistency through validation mechanisms analogous to those used for schema migration or dependency resolution: cycle detection at commit time, uniqueness constraints on continuation edges, and reference integrity checks that prevent orphaned links. These mechanisms are engineering requirements for production deployment, not theoretical concerns.

\paragraph{Activation Policy.}
The uppermost layer is the \emph{Activation Policy}: the set of rules governing which skills are available to which agents under which conditions. Activation policies encode organizational constraints---a junior agent operating in a development environment may have access to a broad set of exploratory skills, while an agent operating in production may be restricted to a narrow set of thoroughly vetted, human-approved skills. Policies are evaluated dynamically based on agent identity, target environment, time of day, active incident status, and other contextual signals. The Activation Policy layer ensures that Agent Knowledge Architecture is not merely a knowledge catalog but a governed knowledge system that respects organizational boundaries (discussed further in Section~\ref{sec:governance}).

\subsection{From Internal Developer Platforms to Agent-Navigable Topologies}

Traditional Internal Developer Platforms (IDPs) expose their capabilities through user interfaces, command-line tools, and API endpoints~\cite{gartner2024platform}---consumed by humans and deterministic orchestrators alike. Regardless of the consumer, these interfaces expose discrete, predetermined operations: a form to provision a database, an API call to trigger a pipeline, a CLI command to scaffold a service. Each operation is a fixed entry point with fixed parameters, offering no mechanism for the consumer---whether human or machine---to adapt the workflow mid-execution based on intermediate results. Agent Knowledge Architecture differs in a fundamental respect: it exposes \emph{semantically rich skill graphs} that agents traverse adaptively. Where an IDP presents a fixed operation---``Deploy a Service'' with predetermined parameters---Agent Knowledge Architecture exposes a subgraph of skills---build, test, security-scan, stage, canary, promote, monitor---with typed edges encoding their sequencing, and metadata encoding their preconditions, blast radius, and rollback affordances. The distinction is not one of interface modality but of knowledge representation: from predetermined operations with fixed parameters to structured, composable knowledge that an agent can reason over, sequence dynamically, and adapt as it progresses toward its goal.

\subsection{Paved Lanes: Pre-Composed Skill Chains}

While agents are capable of composing skills dynamically by traversing the Knowledge Topology, organizations benefit from encoding their most important workflows as \emph{paved lanes}: pre-composed skill chains that represent established best practices. A paved lane for the canonical deployment pipeline, for example, might specify the exact sequence of skills---code-review, build, unit-test, integration-test, security-scan, stage-deploy, canary-analysis, production-promote---along with branching conditions, retry policies, and escalation triggers. Paved lanes serve a dual purpose. First, they reduce the planning burden on agents by providing validated, organization-approved paths for common tasks, thereby conserving context window capacity for reasoning about the novel aspects of a request. Second, they function as executable documentation of organizational best practices: unlike a wiki page describing the deployment process, a paved lane \emph{is} the deployment process, expressed in a form that both agents and governance systems can interpret and enforce.

\subsection{AI-Generated Golden Paths}

The limitations of deterministic workflow templates and the enabling infrastructure of the three-layer Agent Knowledge Architecture converge in a new construct that represents the central architectural contribution of this section.

\begin{definition}[AI-Generated Golden Path]
\label{def:ai-generated-golden-path}
An \textbf{AI-Generated Golden Path} is a workflow dynamically composed by an autonomous AI agent at runtime by traversing the Knowledge Topology, selecting and chaining Atomic Knowledge Units (skills) based on the task requirements, organizational context, available tools, and governance constraints. Unlike deterministic golden path templates, an AI-Generated Golden Path is not pre-composed but \emph{generated}---assembled from validated skills at the point of need, adapting to the specific parameters of each task.
\end{definition}

The generation process operates as follows. Given a high-level task (e.g., ``release version 2.3.1 of the payments service''), the agent queries the AKU Registry, traverses the Knowledge Topology to identify required skills and their dependencies, evaluates Activation Policies to determine which skills are available in the current context, and composes a workflow by chaining skills through their continuation paths. The result is a golden path that is unique to this specific task, this specific service, this specific moment---yet grounded in the organization's codified best practices. Where a paved lane provides a pre-composed path for anticipated workflows, an AI-Generated Golden Path extends this capability to any task, including novel ones that no platform engineer anticipated.

The contrast with classic golden paths is instructive. Classic golden paths are pre-composed, deterministic, template-based, and brittle; AI-Generated Golden Paths are agent-composed, probabilistic, skill-based, and adaptive. Classic paths are authored by platform teams for anticipated use cases; AI-Generated Golden Paths are composed by agents for any task the organization's skill library can support. Classic paths break when underlying systems change; AI-Generated Golden Paths adapt because the agent re-traverses the topology at each invocation, incorporating updated skills, revised policies, and current context.

The key enabler of this capability is \emph{institutional knowledge}. A capable large language model with tool access is necessary but not sufficient for generating effective golden paths. Without skills encoding institutional knowledge---organizational conventions, governance requirements, team-specific practices, and hard-won operational lessons---the agent generates paths that violate organizational norms, trigger governance failures, and cause context rot through trial-and-error. AKUs provide the institutional grounding that transforms generic AI capability into organizationally effective action.

The developer productivity implications are substantial. Instead of waiting weeks for a platform team to design, build, test, and ship a deterministic workflow template for each new use case, agents compose AI-Generated Golden Paths on demand, guided by the same institutional knowledge a senior engineer would apply---but at machine speed and without the bottleneck of human availability. This on-demand composition also reduces the institutional knowledge tax (Section~\ref{sec:context-window-economy}) by removing senior engineers from the knowledge delivery path: the skills encode what they know, and the agent activates that knowledge without requiring their real-time intervention. The platform team's role evolves from authoring brittle templates to curating the skill library and knowledge topology from which agents compose workflows dynamically.

This concept has emerging industry momentum. Microsoft's platform engineering guidance envisions ``agent golden paths'' in which reference architectures become machine-consumable blueprints~\cite{microsoft2025platformagentic}. The Agent Skills open standard~\cite{anthropic2025agentskills} provides the interoperability layer on which skill-based composition is built. The AI-Generated Golden Path formalizes this vision and grounds it in the Knowledge Activation framework, providing both the theoretical foundation and the architectural infrastructure for its realization.

\subsection{Skill Discovery}

The practical utility of Agent Knowledge Architecture depends on effective skill discovery: the ability of an agent to find the right skill at the right time. Discovery operates through multiple complementary mechanisms. \emph{Semantic matching} allows agents to query the AKU Registry using natural language intent descriptions, leveraging embedding-based retrieval to identify candidate skills whose descriptions align with the agent's current goal~\cite{lewis2020rag}. \emph{Trigger conditions} embedded in skill metadata specify the contexts under which a skill becomes relevant---for instance, a skill for handling database connection pool exhaustion might declare an activation trigger tied to specific monitoring alert patterns. \emph{Organizational context} narrows the search space: an agent operating within the payments team's domain automatically has its discovery scope weighted toward payment-relevant skills. Together, these mechanisms ensure that skill discovery is not a brute-force search over a flat catalog but a context-sensitive, intent-driven process that reflects the organizational structure in which the agent operates. Effective discovery is particularly critical for AI-Generated Golden Path composition: the quality of the generated workflow depends directly on the agent's ability to identify the right skills, in the right order, with the right governance posture, from across the organization's entire knowledge surface.

The three-layer Agent Knowledge Architecture---AKU Registry, Knowledge Topology, and Activation Policy---thus provides the structural foundation for deploying Skills at organizational scale and for enabling AI-Generated Golden Paths as the primary mechanism through which agents deliver value. It transforms the Internal Developer Platform from a human-facing portal into a knowledge infrastructure that autonomous agents can navigate, reason about, and execute within, composing validated workflows on demand while remaining subject to the governance constraints that enterprise environments demand. The AI-Generated Golden Path represents the culmination of this architectural vision: a shift from pre-composed, brittle workflow templates to dynamically generated, institutionally grounded workflows that adapt to each task, each team, and each moment.

\section{Governance and Organizational Metadata}
\label{sec:governance}

The deployment of autonomous AI agents in enterprise software development introduces a governance challenge of considerable complexity. Agents that can read code, invoke APIs, modify infrastructure, and interact with production systems must do so within the boundaries of organizational policies, security requirements, and regulatory compliance obligations~\cite{nist2024ai}. Traditional governance models---access control lists, approval gates, audit logs---were designed for human actors whose behavior is mediated by judgment, social norms, and professional accountability. Autonomous agents lack these implicit constraints. Their governance must therefore be explicit, computable, and embedded in the very knowledge structures they consume.

The urgency of this challenge is underscored by empirical evidence: a large-scale study
of 2{,}303 agent context files across 1{,}925 repositories found that security and
governance constraints are specified in only 14.5\% of
cases~\cite{chatllatanagulchai2025agentreadmes}. The vast majority of knowledge artifacts
currently delivered to agents carry no governance metadata whatsoever---a gap that this
section addresses directly.

This section argues that governance is most effective when it is treated as a
\emph{first-class property of AKUs}, rather than as an external enforcement layer
applied after the fact. By embedding governance metadata directly into Atomic Knowledge
Units, the Knowledge Activation framework ensures that agents encounter organizational
constraints not as obstacles to be circumvented but as intrinsic features of the
knowledge they execute.

\subsection{Governance as Intrinsic AKU Property}

Each AKU in the Knowledge Activation framework carries governance metadata across four dimensions.

\paragraph{Permission Scoping.}
Every skill declares its \emph{blast radius}: the set of resources it can read, write, create, or delete. A skill for rotating database credentials, for example, declares write access to the secrets manager and read access to the service registry, but explicitly excludes access to application data. Permission scoping operates on the principle of least privilege, applied not to the agent as a whole but to each individual knowledge unit the agent executes. This granularity is critical. An agent may hold broad organizational permissions in aggregate, but any single skill invocation is constrained to precisely the resources that skill requires. The agent runtime enforces these declarations, rejecting any tool invocation that exceeds the skill's declared scope.

\paragraph{Approval Workflows.}
Skills can specify that certain actions require human sign-off before execution. A skill for scaling a production database cluster might execute autonomously in staging but require explicit confirmation when targeting production. These requirements are encoded as metadata within the skill, not as external process gates, ensuring that they travel with the knowledge unit regardless of which agent invokes it or through which orchestration path it is reached.

The mechanism supports multiple patterns: synchronous blocking (the agent waits), asynchronous deferral (the agent proceeds with other tasks and returns when granted), and conditional bypass (required only when specific risk thresholds are exceeded).

\paragraph{Audit Trail Integration.}
Enterprise environments require comprehensive audit trails for compliance, incident investigation, and continuous improvement. Skills embed logging and observability requirements directly in their metadata, specifying which events must be recorded, at what granularity, and to which systems. When an agent executes a skill, the runtime automatically generates records that capture the agent's identity, the skill invoked, the inputs provided, the actions taken, and the outcomes observed.

Because these requirements are properties of the skill rather than of the runtime, different skills can specify different logging granularities appropriate to their risk profile---a low-risk code formatting skill may require only summary logging, while a production deployment skill may require step-level recording with full input and output capture.

\paragraph{Compliance Annotations.}
Skills carry regulatory and policy tags that constrain agent behavior based on organizational compliance requirements. A skill operating on systems that process personal data may carry a GDPR annotation that triggers additional data handling constraints. A skill that modifies financial transaction systems may carry SOX compliance tags that require dual authorization. These annotations function as machine-readable policy labels: the agent runtime evaluates them against the current compliance context and enforces the corresponding constraints automatically. Compliance annotations transform regulatory requirements from documents that humans must interpret and manually apply into executable metadata that governs agent behavior deterministically.

\paragraph{Validators.}
A key mechanism for operationalizing governance without human bottlenecks is the \emph{validator}: a deterministic script embedded within the skill that automatically verifies whether the agent's actions meet organizational standards. Validators are implemented as executable code---shell scripts, Python checks, or policy-as-code rules (e.g., Open Policy Agent)---that produce pass/fail results with structured logs.

Unlike human approvers, validators are consistent (they apply identical rules every time), scalable (they execute in milliseconds regardless of volume), and auditable (every decision is logged with its inputs and the rule applied). By encoding governance checks as validators rather than human review gates, the framework enables governance teams to shift from \emph{governance-as-approval} to \emph{governance-as-code}: authoring deterministic governance artifacts that scale with the skill library rather than with headcount.

This shift mirrors the Infrastructure-as-Code transformation that freed operations teams from ticket-based provisioning. Validators free governance teams from ticket-based approval, allowing them to focus their expertise on authoring effective rules rather than repetitively applying them.

\subsection{Organizational Metadata Pairing}

Beyond governance constraints, each skill carries \emph{organizational metadata} that situates it within the broader enterprise context. This metadata pairing ensures that agents operate not as context-free executors but as participants that understand the organizational landscape surrounding each action.

The organizational metadata schema encompasses several dimensions. \emph{Team ownership} identifies which team authored and maintains the skill, enabling agents to route questions, escalations, and feedback to the appropriate human stakeholders. \emph{Service tier} classification (e.g., Tier 1 critical, Tier 2 standard, Tier 3 best-effort) informs the agent's risk calculus when selecting among alternative approaches. \emph{On-call rotation} and \emph{escalation contacts} ensure that when an agent encounters a situation requiring human intervention, it can identify and notify the correct individuals without searching through organizational directories. \emph{SLA requirements} declare the expected performance characteristics of the systems the skill operates on, enabling agents to make informed decisions about timing, batching, and retry strategies.

Additional metadata dimensions include \emph{environment awareness}---skills declare which environments (production, staging, development, sandbox) they are certified to operate in---and \emph{cost center attribution}, which enables organizations to track the resource consumption of agent-executed skills for chargeback and budgeting purposes. \emph{Data classification} tags identify the sensitivity level of data the skill may encounter, triggering appropriate handling protocols. \emph{Service catalog integration} links each skill to the broader service catalog entries it relates to, maintaining traceability between agent actions and the organizational service model.

\subsection{The Governance Gradient}

Not all agent actions carry equal risk, and governance mechanisms should reflect this reality. The Knowledge Activation framework introduces a \emph{governance gradient} that maps skills along a spectrum from fully autonomous to fully human-controlled, based on their metadata properties.

At one end of the gradient, low-risk skills with well-established track records---code formatting, documentation generation, test scaffolding---operate in a fully autonomous mode. The agent discovers, selects, and executes these skills without human intervention. Their permission scopes are narrow, their blast radii are contained, and their failure modes are reversible. At the opposite end, high-risk skills involving irreversible changes to production systems, financial transactions, or sensitive data require human-in-the-loop approval at one or more stages. Between these extremes lies a continuum of intermediate governance postures: skills that require notification but not approval, skills that operate autonomously during business hours but require approval outside them, and skills that are autonomous for certain agent roles but gated for others.

\begin{figure}[htbp]
  \centering
  \includegraphics[width=\textwidth]{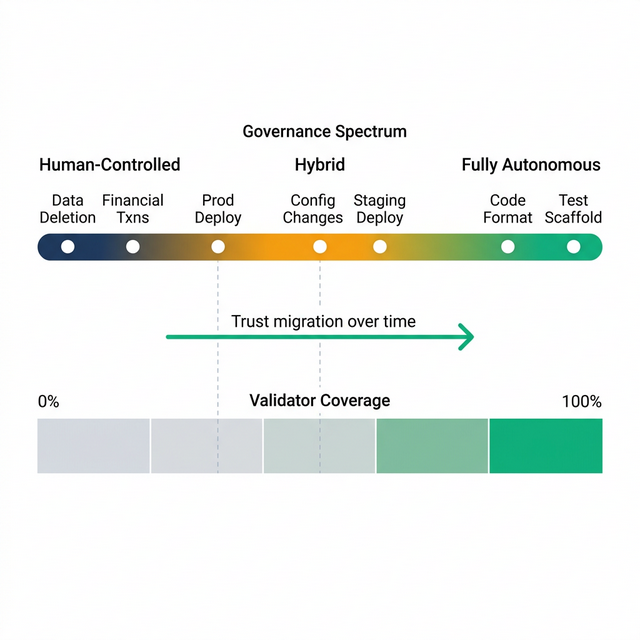}
  \caption{The Governance Gradient. Skills are positioned along a continuous spectrum from fully autonomous to fully human-controlled operation, determined by validator coverage, permission scope, and historical execution record. Over time, skills migrate toward greater autonomy as validator coverage increases and successful execution records accumulate.}
  \label{fig:governance-gradient}
\end{figure}

The position of a skill on the governance gradient is determined by its metadata: permission scope, environment target, data classification, service tier, and historical execution record. Critically, skills can \emph{migrate along the gradient over time}. A newly authored skill might begin in a fully supervised mode; as it accumulates a track record of successful, incident-free executions, the organization may progressively relax its governance requirements, moving it toward greater autonomy. This graduated trust model mirrors how organizations extend autonomy to human engineers---beginning with supervised tasks and progressively granting independence as competence is demonstrated.

Validators are the mechanism that makes the governance gradient operationally tractable. The degree of validator coverage---the proportion of a skill's governance requirements that can be verified by deterministic scripts---determines the skill's position on the gradient. A skill with complete validator coverage can operate fully autonomously; the validators enforce governance at machine speed without human intervention. A skill with partial validator coverage operates in a hybrid mode: validators handle the codifiable checks while human reviewers address the aspects that require judgment. A skill with no validators requires full human oversight. The governance team's strategic objective becomes increasing validator coverage across the skill library, progressively moving more skills toward autonomous operation. This provides a measurable, tractable roadmap for expanding agent autonomy within the enterprise.

\subsection{Building Enterprise Trust}

The embedding of governance directly into the knowledge layer addresses a fundamental barrier to enterprise adoption of autonomous agents: trust. Organizations are reluctant to grant broad autonomy to systems whose behavior they cannot predict, constrain, or audit~\cite{nist2024ai}. By making governance an intrinsic property of every knowledge unit an agent can execute, the framework ensures that organizational control scales with agent capability. As new skills are authored, they arrive pre-equipped with governance metadata. As agents discover and compose skills, the governance constraints compose as well---a skill chain inherits the union of all constituent governance requirements. Validators, in particular, compose when skills compose: when an agent assembles an AI-Generated Golden Path from multiple skills, the composed workflow inherits the union of all constituent validators---governance is safe by construction, analogous in spirit to type safety in programming, where well-typed components compose into well-typed programs. The result is a system in which expanding the agent's knowledge does not require expanding the governance infrastructure in parallel; governance travels with the knowledge itself.

This approach shifts where governance occurs. Rather than enforcing policies at the agent runtime boundary---which requires the governance layer to anticipate every possible agent behavior---organizations embed governance directly at the knowledge layer. Every action an agent can take arrives already annotated with the constraints it must respect. Governance shifts left: from runtime enforcement to knowledge-time specification.

\section{Case Study: An Enterprise Deployment of Skill-Based Knowledge Activation}
\label{sec:case-study}

\subsection{Deployment Context}
\label{sec:case-study:context}

We deployed the framework described in this paper at Yahoo, a global
internet-services company employing approximately 2{,}000 engineers across
multiple consumer-facing and enterprise business units---mail, finance,
sports, media, advertising, central technology, and others. The deployment
is the basis for the empirical findings reported in the remainder of this
section.

The deployment was motivated by the \emph{last-mile problem of agentic
software development} introduced in \S\ref{sec:introduction}: the gap between
general foundation-model capability and organization-native agent behavior,
closable only by an institutional knowledge architecture delivered to the
agent at the moment of need. At Yahoo, this gap is concrete and
substantial---decades of accumulated architectural decisions, deployment
procedures, authentication models, security baselines, cost-tagging
policies, and compliance requirements---none of which appears in the
training corpus of any foundation model, and none of which a general-purpose
AI agent can infer from the codebase alone.

The starting condition was characteristic of mature enterprise environments:
institutional knowledge accumulated over more than two decades but distributed
across many code repositories, internal wikis, runbooks, security control
catalogs, and tribal channels. Several categories of authoritative source
already existed at Yahoo: a security baseline program publishing controls
against cloud configurations, a workload identity framework governing
service-to-service authentication, central platforms for continuous integration
and deployment, observability, cost tagging, secret and certificate management,
and incident response, alongside first-day onboarding documentation, an
internal package registry, and an enterprise search index covering the rest.
That knowledge had always existed; an experienced Yahoo engineer carried a
slice of it in their head, and a new engineer absorbed it over weeks. What
did not exist was an agent-consumable codification of it: when a Yahoo
engineer asked an AI coding assistant for help with a company-specific
task---authenticate to the federated cloud environment, provision a workload
identity, deploy a service the Yahoo way, harden it against the security
baseline---the model fell back on its parametric training and produced
generic advice that violated Yahoo-specific conventions until the engineer
corrected it. The last-mile gap manifested one query at a time, paid for in
context-window tokens consumed by failed attempts and senior-engineer hours
spent on the correction cascade.

We codified this knowledge into a corpus of 87 modular, agent-consumable
skills (79 organization-wide, 8 team-scoped), distributed as a plugin via
Yahoo's internal developer marketplace and auto-imported into a second AI
coding assistant the organization also supported. Each skill cited its
authoritative source and followed a consistent structure. Skills were grouped
into nine categories covering first-day onboarding, the two major cloud
providers in use, build and deploy workflows, observability and incident
response, security and compliance hardening, code development, and meta-skills
for managing the corpus itself. Coverage extended from a new Yahoo engineer's
first-day workstation setup through production-readiness review for a
deployed service.

We did not invent Yahoo's organizational standards. We codified them. The
contribution of the deployment, relative to the body of documentation it
built upon, was the act of codification itself: rewriting human-targeted
documents into agent-readable, governance-aware skills; organizing them into
a composable graph; gating every change through a multi-criterion quality
review; and shipping the result to every Yahoo engineer's AI session by
default. From repository inception to the survey reported in this section,
57 days elapsed, 68 pull requests merged, and 26 skill-gap reports were
filed by users; 100\% of those reports were triaged to closure.

\subsection{Architecture Realization}
\label{sec:case-study:architecture}

The deployment realizes the architecture proposed in
Sections~\ref{sec:atomic-knowledge-units}
and~\ref{sec:golden-path-architecture} as a \emph{knowledge graph}: each skill
is a self-contained node carrying its institutional intent and procedure,
connected to other skills through named references, and activated dynamically
at the moment of an engineer's query rather than loaded en masse at session
start. Three properties of the graph drive the deployment's runtime behavior:
skills are \emph{connected by reference}---continuations make the dependency
structure between skills explicit and machine-traversable; \emph{self-discoverable}---each
skill declares the keyword and error-pattern triggers that determine when it
loads in response to which kind of engineer query, so no central router is
required; and \emph{session-context activated}---the agent loads skills
on demand based on what the engineer is doing, never all 87 at once. The
graph structure is what makes the corpus a knowledge architecture rather
than a static documentation set: skills compose at runtime to handle requests
the authoring team did not enumerate in advance.

\paragraph{Atomic Knowledge Unit schema (\S\ref{sec:atomic-knowledge-units}).}
Each skill bundles the seven components of the AKU schema. \emph{Intent}---a
one-sentence description of the institutional task the skill addresses.
\emph{Procedure}---the action steps an agent should take to satisfy the
intent. \emph{Tools}---the specific commands, APIs, and platform invocations
the procedure draws on. \emph{Metadata}---the keyword and error-message
triggers the assistant uses to discover the skill at the moment of need,
including, for security-hardening skills, the verbatim control identifiers
an engineer would paste from a scan output. \emph{Governance}---citations
to the authoritative Yahoo sources that the skill encodes, and the policy
constraints the procedure enforces. \emph{Continuations}---named pointers to
downstream skills that should compose with this one, such as a
cloud-resource builder skill naming its corresponding security-hardening
skill, an onboarding skill naming the authentication skills it depends on,
or an incident-response skill naming the observability and runbook skills
it composes with. \emph{Validators}---where applicable, the post-action
checks an agent runs before declaring the task complete. We did not
implement validators uniformly across all 87 skills; validator coverage
is one of the empirical findings discussed in
\S\ref{sec:case-study:operational-learnings}.

\paragraph{Agent Knowledge Architecture (\S\ref{sec:golden-path-architecture}).}
We realized the three architectural layers as follows.
\begin{itemize}
  \item \textbf{AKU Registry.} The canonical surface from which agents
    discover skills is a single repository containing all 87 skill packages,
    distributed via Yahoo's internal plugin marketplace and additionally
    auto-imported into a second AI coding assistant the organization
    supports. A single canonical schema file (the AGENTS.md
    convention~\cite{liu2026agentsmd}) anchors the corpus; tool-specific
    configuration files consumed by individual AI assistants refer back to
    the canonical schema, so the same institutional knowledge reaches every
    supported tool surface without duplication.
  \item \textbf{Knowledge Topology.} The dependency graph by which one skill
    chains to another is realized as explicit named continuations within
    each skill, plus a maintained inventory document the agent reads at
    runtime to compose multi-step workflows. The topology is what makes
    the corpus a knowledge graph rather than a flat skill list. Edges are
    not implicit through retrieval similarity; they are declared by the
    skill authors, reviewed at merge time, and visualized for engineers
    through an interactive graph view of the corpus.
  \item \textbf{Activation Policy.} The rules by which a skill is loaded
    into context at the moment of an engineer's query are realized through
    the AI assistant's native skill-discovery hooks, configured to trigger
    on keyword matches against the engineer's query and on error-message
    patterns pasted directly from cloud-provider scan output. The engineer
    does not invoke skills by name; the activation policy resolves the
    query to the relevant skill(s) and loads them into the assistant's
    context window. From the engineer's perspective the agent appears to
    know Yahoo natively, with no slash commands, doc-hunting, or manual
    skill selection.
\end{itemize}

\paragraph{AI-Generated Golden Paths as a primitive
(\S\ref{sec:golden-path-architecture}).}
The composition of the three layers above makes \emph{AI-Generated Golden
Paths}---defined in \S\ref{sec:golden-path-architecture} as workflows
dynamically composed by an agent at runtime---a first-class primitive
available to every engineer in the deployment. A natural-language request
such as \emph{``deploy a new service to the federated cloud environment, the
Yahoo way''} resolves at the agent layer into a multi-step chain:
authenticate $\rightarrow$ provision project $\rightarrow$ provision workload
identity $\rightarrow$ deploy $\rightarrow$ harden against the security
baseline. Each step is drawn from the topology by following the
continuations declared on the previous step; the activation policy loads
each skill into context only when the traversal reaches it. The engineer
issues one ask; the activation policy resolves it; the topology supplies
the traversal; the result is a complete, compliant workflow assembled at
the moment of need. Critically, golden paths are not authored as separate
artifacts in this deployment---they are \emph{emergent properties} of the
knowledge graph, available wherever the graph is dense enough to support
the traversal. This eliminates an entire class of drift: there is no
golden-path document to keep in sync with the underlying skills, because
the skills \emph{are} the golden path, composed afresh on each query.

\paragraph{Governance gradient (\S\ref{sec:governance}).}
The deployment realizes the governance-by-validators discipline along two
axes. First, every skill change must pass a multi-criterion review (we
apply twenty-four criteria, including procedural correctness, source
citation completeness, validator coverage where applicable, and explicit
composition with downstream skills) before merging. Second, every skill
that creates a cloud resource is required to chain to a corresponding
hardening skill that applies the relevant security baseline controls; the
builder--hardener dependency map is maintained in the canonical schema and
enforced at review. Together, these realize the per-skill governance
gradient described in \S\ref{sec:governance}: builder skills whose
downstream hardener has full validator coverage operate at a higher
autonomy level than skills whose dependent constraints must still be
applied by human follow-up.

\paragraph{Meta-skills: a self-referential authoring economy.}
A practical consequence of the architecture's design is that several
operational functions of the corpus are themselves expressed as AKUs.
We refer to these as \emph{meta-skills}---skills whose intent is to
manage other skills. The deployment ships four.

\begin{enumerate}
  \item \emph{Authoring.} A meta-skill that scaffolds new skill drafts
    against the canonical schema---the AGENTS.md
    convention~\cite{liu2026agentsmd}, the seven-component AKU
    structure, and the multi-criterion review rubric. Given an
    adequate source document (a runbook, security baseline, or
    platform guide), authoring time per new skill drops to minutes of
    agent-assisted drafting plus the iteration cycles required by the
    evaluator.
  \item \emph{Evaluation.} A meta-skill that scores candidate skills
    against the review rubric and produces structured feedback that
    the author iterates against until the evaluator returns a passing
    assessment. Evaluation is invoked both during authoring and on
    every pull request before merge.
  \item \emph{Gap reporting.} A meta-skill an engineer invokes when an
    existing skill produces an incorrect or incomplete result. The
    meta-skill files a structured report upstream, classifying the
    failure mode (missing skill, incomplete skill, wrong routing,
    missing trigger, missing precondition) and capturing the
    engineer's expected behavior. Every gap report becomes input to
    the next authoring cycle.
  \item \emph{Discovery fallback.} A meta-skill the agent invokes when
    it senses that no codified skill matches the query and that its
    parametric knowledge is insufficient---typically signaled by
    repeated correction cycles or low-confidence outputs. The fallback
    routes the query through a Model Context
    Protocol~\cite{anthropic2024mcp} connector to the organization's
    enterprise search platform, surfacing relevant uncodified
    material so the agent can attempt the task; the engineer can then
    invoke gap reporting to convert the discovered material into a
    new skill candidate.
\end{enumerate}

Together these four meta-skills establish a \emph{self-referential
property} of the framework as deployed: the corpus contains the
primitives required to author, evaluate, maintain, and extend itself.
The cost of growing the corpus therefore scales with the quality of
underlying institutional documentation rather than with the seniority
of the authoring engineer; the failure modes of the corpus convert
into authoring opportunities through the gap-reporting loop; and the
absence of codified skills degrades gracefully into a
retrieval-mediated discovery channel that produces the seeds of
future codification. A substantial fraction of the operational load
of running the corpus has migrated from the maintaining team to the
meta-skills themselves.

\subsection{Empirical Methodology}
\label{sec:case-study:methodology}

To assess the deployment's effect on developer experience, we fielded an
anonymous, online cross-sectional survey of Yahoo engineers between May~1
and May~15, 2026. The instrument was a fifteen-section form with conditional
routing: respondents who reported as active plugin users answered a battery
of driver questions, time-and-value questions, an open-feedback section, and
a Net Promoter Score item; respondents who had installed but lapsed answered
a barrier section; new hires (under six months at Yahoo) answered a
supplemental ramp-up section. No prompt content, source code, or other
identifying data was collected; respondents could optionally provide an
email address for a follow-up conversation.

Sixty-seven engineers responded. Forty (60\%) reported as active users of
the plugin, nineteen (28\%) had not installed it, and eight (12\%) had
installed but lapsed in active use. Respondents represented nine Yahoo
business units (mail, finance, sports, media, advertising, central
technology, and others), with the maintaining team---which originated the
deployment---contributing twenty-one respondents. The sample was AI-fluent:
85\% reported daily use of AI coding assistants prior to the plugin's
existence, and 98\% used the same primary AI assistant. These
characteristics matter for interpretation. The cohort is not a population
of skeptics; our findings should be read as describing the experience of
AI-native engineers using a knowledge-architected agent versus the same
engineers using an architecturally na\"ive one, rather than as the
experience of engineers encountering AI tooling for the first time.

We acknowledge two concurrent confounders. During the survey window,
approximately half of active-user respondents (50\%) also received an
upgrade to a more capable foundation model in their AI assistant, and
roughly a third (32\%) reported having recently begun work in a new
codebase. Both confounders could plausibly inflate the perceived benefit
attributed to the plugin. We discuss the implications in
\S\ref{sec:case-study:validity}; the findings remain statistically
significant after acknowledging the confounders, and we report effect sizes
that exceed what either confounder alone would plausibly produce.

For analysis, we computed one-sample $t$-tests against the neutral midpoint
(3 on the 5-point Likert scale) for each driver question, with effect sizes
reported as Cohen's $d$. We computed bootstrap confidence intervals for the
Net Promoter Score with 10{,}000 resamples. We computed Mann--Whitney $U$
tests to compare driver responses between respondents from the maintaining
business unit and respondents from all other business units combined; a
chi-squared test of independence to assess whether tenure with the plugin
associated with perceived impact; and a Spearman rank correlation to assess
any monotonic relationship between tenure and comparative experience. All
confidence intervals are reported at the 95\% level. Effect-size
interpretations follow Cohen's conventions (small 0.2, medium 0.5, large
0.8) with ``very large'' applied to $d > 1.2$ per Sawilowsky's extension.

\subsection{Findings: Developer Experience Drivers}
\label{sec:case-study:dx-drivers}

We operationalized the deployment's expected effect on developer experience
as four driver questions, each measuring a distinct value dimension on a
five-point Likert scale (strongly disagree to strongly agree). The four
questions, abbreviated here, were:

\begin{itemize}
  \item \textbf{Q6 (mental effort).} ``The plugin reduces the mental effort
    of remembering Yahoo-specific conventions, tools, and policies.''
  \item \textbf{Q7 (flow preservation).} ``When I have a Yahoo-specific
    question, I get a usable answer fast enough to keep working---rather
    than context-switch to internal docs, [the wiki], or [chat].''
  \item \textbf{Q8 (surfacing requirements).} ``When I ask the plugin about
    a task, it often surfaces relevant Yahoo-specific requirements or steps
    I didn't think to ask about (e.g., security controls, cost rules,
    required tags, downstream dependencies, compliance steps).''
  \item \textbf{Q9 (first-try compliance).} ``The plugin helps me follow
    Yahoo's required policies (security, compliance, cost) on the first
    try.''
\end{itemize}

Each driver corresponds to a specific architectural claim the framework
makes. Q6 (mental effort) tests the central claim of the Context Window
Economy (\S\ref{sec:context-window-economy}): when institutional knowledge
is delivered to the agent rather than reconstructed by the engineer, the
engineer's own cognitive load drops. Q7 (flow preservation) tests the
Activation Policy claim from \S\ref{sec:case-study:architecture}: that
session-context activation delivers knowledge fast enough to obviate
context-switching to documentation. Q8 (surfacing requirements) tests the
Knowledge Topology claim: connected skills, traversed by the agent at the
moment of need, reveal downstream requirements the engineer did not know to
ask about. Q9 (first-try compliance) tests the governance gradient claim
(\S\ref{sec:governance}): codified institutional policies, applied by the
agent rather than the engineer, reduce iteration on policy violations.

Table~\ref{tab:dx-drivers} summarizes the results. All four drivers are
significantly above the neutral midpoint (all $p < 0.0001$), with effect
sizes that are large or very large by conventional thresholds. The
strongest signal is on Q6 (mental effort): a top-box rate of 75\% and a
Cohen's $d$ of 1.40 place this finding firmly in the very-large range and
make it the strongest result in the survey. The three remaining drivers
all fall in the large-effect range ($d = 0.89$--$0.97$).

\begin{table}[ht]
\centering
\caption{Developer experience driver findings. Top-box is the proportion of
respondents selecting ``agree'' or ``strongly agree''; mean is on a 5-point
Likert scale (neutral $= 3$). One-sample $t$-tests against the neutral
midpoint, $p < 0.0001$ for all four drivers; degrees of freedom range from
37 to 39 due to optional-question skip patterns. Effect sizes follow
Cohen's conventions with ``very large'' applied to $d > 1.2$.}
\label{tab:dx-drivers}
\begin{tabular}{lcccl}
\toprule
\textbf{Driver dimension} & \textbf{Top-box} & \textbf{Mean (95\% CI)} & \textbf{Cohen's $d$} & \textbf{Effect size} \\
\midrule
Reduced mental effort (Q6) & 75\% & 4.05 (3.81--4.29) & 1.40 & Very large \\
Flow preservation (Q7)     & 72\% & 3.87 (3.55--4.19) & 0.89 & Large \\
Surfaces requirements (Q8) & 67\% & 3.72 (3.46--3.98) & 0.91 & Large \\
First-try compliance (Q9)  & 66\% & 3.79 (3.52--4.06) & 0.97 & Large \\
\bottomrule
\end{tabular}
\end{table}

The four drivers are not independent constructs. Q6 (mental effort), Q7
(flow preservation), and Q8 (surfacing requirements) all describe aspects
of the same underlying experience---perceived reduction in last-mile
friction---and we would expect them to correlate. We did not compute the
inter-driver correlation because the battery was designed to surface which
value dimensions land most strongly with practitioners, rather than to
identify an underlying factor structure; we treat the four drivers as
complementary measurements of a shared phenomenon. The implication for
multiplicity adjustment is that the four tests should not be treated as
independent: even under a conservative Holm--Bonferroni correction across
four correlated drivers, all four remain significant at $p < 0.001$ given
the original $p < 10^{-4}$ values, so the qualitative interpretation is
unchanged.

Respondents' open-text rationales for their Net Promoter scores reinforce
the pattern the quantitative findings suggest. Engineers described the
plugin as ``the easy button for doing things in a compliant manner that
makes sense,'' as ``less frustration and dependence on institutional
knowledge,'' and as ``a fantastic resource---it's like having a Yahoo
engineering expert on hand [for] very specific Yahoo related questions.''
One respondent captured the activation-policy claim almost verbatim: ``It
meets me where I am in the workflow, I don't need to discover it.''
Another, addressing the topology claim directly: ``The context it has
provided me has been invaluable, and faster than trying to synthesize
from [the enterprise search].''

What these findings do not show, and what we therefore do not claim. The
drivers are self-reports of perceived experience, not direct observations
of agent task completion rates, correction frequency before and after, or
wall-clock time to completion on specific tasks. We discuss what this
implies for the strength of the evidence in
\S\ref{sec:case-study:validity}. The drivers support the framework's claims
about \emph{perceived} developer experience with statistical confidence;
claims about \emph{measured} task outcomes require a different study
design and are reported only at the granularity the survey instrument
permits.

\subsection{Findings: Time, Value, and Comparative Experience}
\label{sec:case-study:value}

Beyond the four perceived-experience drivers, the survey instrument
captured three additional dimensions of value: time saved per week (Q11),
counterfactual stickiness if the plugin were removed (Q12), and
comparative experience versus the same AI assistant without the plugin
(Q13). Together with the Net Promoter Score (Q17), these questions
constitute the survey's value module.

\paragraph{Time saved per week (Q11).} Active users reported saving a mean
of 2.6 hours per week (95\% CI: 1.6--3.6 hours), with the distribution
skewing toward small but consistent savings: 60\% of respondents reported
saving at least one hour per week, 28\% reported saving three or more
hours, and a long tail of 8\% reported saving ten or more hours weekly. A
one-sample $t$-test against zero rejected the null of no time saved
($t(39) = 5.19$, $p < 10^{-5}$).

To translate this per-engineer rate to a deployment-scale magnitude, we
multiplied the mean weekly savings by the deployment's install base
(238 active installs at the time of survey) and a 48-week working year.
The resulting illustrative projection is approximately 29{,}000
engineering hours per year (range 18{,}000--41{,}000 at the bounds of the
mean's 95\% confidence interval). We emphasize that this is an
extrapolation, not a directly measured aggregate: it assumes the
population of 238 installs shares the time-savings profile of the 40
active-user respondents, which is unverified. The figure serves to
indicate the order of magnitude of the deployment's potential value at
its current adoption level, not as a precise estimate. Citations of this
finding should use the per-engineer figure (2.6 hours per week, 95\% CI
1.6--3.6) rather than the aggregate projection, which depends on the
unverified assumption that the install base shares the survey cohort's
usage profile.

\paragraph{Time saved relative to net coding time.} The 2.6-hour mean is
modest as an absolute number against a 40-hour workweek (approximately
6.5\% of total work time), but it should be read against the empirical
literature on the fraction of that workweek developers actually spend on
active coding. The literature converges on a directional finding---active
coding is a minority of the developer workweek---while diverging on the
precise percentage in a way that depends primarily on instrument design.

Survey-based studies with granular activity taxonomies report
active-coding shares of approximately 11--16\%. A 2025 ICSE-SEIP study of
484 Microsoft developers reports actual coding time at 11\% of the
workweek against a developer-stated ideal of
20\%~\cite{houck2025timewarp}. IDC's 2024 survey, asking developers to
apportion their time across eight major activity categories, reports
16\% on application development~\cite{idc2024devtime}; Atlassian's 2025
Developer Experience Report (n=3{,}500 across six countries)
independently corroborates the 16\% figure~\cite{atlassian2025dxreport}.
Coarser-grained instruments yield substantially higher numbers:
JetBrains' 2025 \emph{State of Developer Ecosystem} report (n=24{,}534)
finds that 46\% of developers spend more than 60\% of their work time on
coding when ``coding'' is interpreted broadly to include adjacent
activities such as debugging and
refactoring~\cite{jetbrains2025devecosystem}. Instrumented-observation
studies, which are less subject to taxonomy effects, triangulate the
survey findings: Xia et al.\ observed 78 professionals over 3{,}148
working hours and found approximately 58\% of working time devoted to
program comprehension rather than code
production~\cite{xia2018comprehension}, and Minelli et al.\ found that
within IDE sessions specifically, code editing accounts for only about
5\% of activity, with comprehension and navigation consuming the
remainder~\cite{minelli2015summer}. The foundational Meyer et al.\ study
of 5{,}971 Microsoft developers likewise concludes that developers
``spend little time on development''~\cite{meyer2019dailylife}.

The directional conclusion is unanimous across survey and instrumented
methods: active coding is a minority activity in a workweek dominated by
meetings, communication, program comprehension, code review, debugging,
and operational tasks. The 2.6-hour mean savings reported here therefore
sits within an active-coding window of approximately four to twelve
hours per week, representing somewhere between roughly 20\% and 60\% of
that window depending on which instrument's baseline is applied---and
small ($\sim$6.5\%) only when measured against the full forty-hour
workweek.\footnote{This range is computed by applying the empirical
literature's reported active-coding shares (11\% to 30\% across the
cited studies) to a 40-hour workweek and comparing against the Yahoo
within-organization mean of 2.6 hours per week. We do not measure
Yahoo-specific time allocation; the defensible empirical anchor remains
the per-engineer 2.6 hours per week with its 95\% CI of 1.6--3.6 hours.
The reframing is illustrative of the magnitude of the effect relative to
the activity it most directly affects, not a precise ratio applicable to
any single organization.} The result is consistent with the survey's
strongest driver: respondents reported the largest perceived effect on
\emph{mental effort} (Q6, $d = 1.40$), which is precisely the dimension
that recovers attention from operational friction back toward active
coding.

For broader calibration on AI-tool impact, two independent industry
reports are relevant. JetBrains' 2025 survey finds that nearly 9 in 10
developers report saving at least one hour weekly from AI tools, with
one in five reporting eight or more hours saved per
week~\cite{jetbrains2025devecosystem}; the Google Cloud DORA 2025
\emph{State of AI-Assisted Software Development} report (n$\approx$5{,}000)
finds that developers now spend a median of two hours per day working
with AI tools and that adoption has reached 90\% of the global developer
population~\cite{dora2025aistate}. The deployment's per-engineer 2.6-hour
mean sits comfortably within the middle of the JetBrains distribution
and is consistent with the magnitude that a knowledge-architected
deployment would be expected to recover from a general-purpose AI tool.

\paragraph{Scope-bounded effect: one vertical of developer work.} The
deployment's skill corpus concentrates on what is commonly called
infrastructure or platform engineering: cloud provisioning, federated
authentication, continuous integration and deployment, security
hardening, observability, incident response, and first-day onboarding to
the organization's developer platform. It does not directly cover other
verticals of developer work---feature development against business
logic, unit and integration test design, code review and refactoring,
frontend and accessibility engineering, data engineering and
machine-learning workflows, or domain-specific debugging---except where
those activities touch the infrastructure surface (for example, a
feature deployment will compose with the infrastructure provisioning
skills). The 2.6-hour mean therefore reflects savings concentrated in
the slice of the engineer's week where institutional knowledge is
densest: the slice where the organization's infrastructure and security
posture is the most organization-specific surface an engineer
encounters. The framework's architecture, however, is not
vertical-specific. The AKU schema, the knowledge graph structure, the
activation policy, and the governance gradient all generalize to any
domain whose institutional knowledge can be codified into composable,
governance-aware units. Whether extending the corpus into additional
verticals would compound the per-engineer savings is an empirical
question we did not test in the present deployment; the deployment's
2.6-hour figure should therefore be read as a lower bound on what a
fully populated institutional knowledge architecture could deliver,
conditioned on whether codification investment is sustained into the
verticals the present corpus does not yet cover. The lapsed-user
feedback reported in \S\ref{sec:case-study:operational-learnings}
provides concrete signal on which verticals practitioners want
codified next.

\paragraph{Counterfactual stickiness (Q12).} We asked active users what
would happen to their work if the plugin were turned off tomorrow.
Sixty-four percent reported they would meaningfully miss it---losing
hours of productivity (41\% would lose one to three hours per week,
13\% would lose four or more) or actively pushing back to have it
restored (10\%). The remaining 36\% reported they would either not
notice or would adjust without measurable loss. The stickiness finding
qualifies the time-savings result: for the majority of active users,
the savings are not optional convenience but workflow-load-bearing.

\paragraph{Comparative experience (Q13).} We asked active users to
compare their experience using their AI assistant \emph{with} the
plugin against the same assistant \emph{without} the plugin but still
equipped with the organization's other knowledge tools, including its
enterprise search platform. Sixty percent reported a noticeable or
dramatic improvement (top-2 box); 14\% described their experience with
the plugin as dramatic enough that they ``rely on it'' for
Yahoo-specific work. The mean response was 3.57 on a 5-point scale
(95\% CI: 3.25--3.90), where 3 represented ``no difference'' and 5
represented ``dramatically better.'' This finding is the most direct
test of the framework's distinguishing claim against the dominant
existing approach (retrieval-augmented enterprise search): respondents
reported a meaningful advantage of the activation-policy architecture
over the same agent equipped with retrieval-based knowledge access to
the same underlying corpora.

\paragraph{Net Promoter Score (Q17).} The deployment recorded a Net
Promoter Score of $+35$ (95\% CI: $+12$ to $+55$, bootstrap with
10{,}000 resamples), driven by a promoter rate of 48\% and a detractor
rate of 12\%. The mean recommendation score was 8.3 out of 10, with
the modal response 10 (selected by 14 respondents). The confidence
interval is wide because $n = 40$, but the lower bound remains
positive: the deployment is in promoter territory even under the
conservative end of the bootstrap distribution. We do not benchmark
the score against published industry NPS values, because comparable
measurements across developer-tooling deployments are reported with
inconsistent methodologies; we report the score as evidence that the
distribution of respondent sentiment is skewed positively and that
near-half of respondents are sufficiently positive to actively
recommend the deployment to others.

Taken together, the value findings establish three claims supported by
the data. Time saved per active user is significantly greater than zero,
with a mean that is moderate against total work time but substantial
against the much smaller fraction of that time available for direct
coding, and that reflects savings concentrated in a single vertical of
developer work whose institutional density is unusually high. The
perceived value is sticky enough that the majority of users would resist
removal. And the architecture delivers a perceptible improvement over
the same agent equipped with the organization's existing search-based
knowledge access.

\subsection{Findings: Cross-Unit Equality and Ramp-Up Signal}
\label{sec:case-study:generalization}

The framework's central claim is architectural: the Institutional
Impedance Mismatch is a structural property of agentic software
development, and any knowledge consumer encountering organizational
work without institutional context experiences the same deficit. If
the claim holds, the deployment's perceived value should not
concentrate in the maintaining team, in long-tenured users, or in
engineers fluent with the deployment's authoring conventions. We
tested these implications with three robustness analyses and one
supplementary cohort.

\paragraph{Cross-business-unit equality.} We compared driver responses
between respondents from the maintaining business unit ($n = 21$) and
respondents from all other business units combined ($n = 19$) using
Mann--Whitney $U$ tests, one per driver. No comparison reached
significance; all four drivers returned $p > 0.6$. We additionally
compared the Net Promoter Score across the same partition: the
maintaining unit recorded NPS $+38$ and the rest of the organization
$+32$ (Mann--Whitney $p = 0.725$). These null results are consistent
with the generalization claim but, with samples of $n = 21$ and $n = 19$,
do not formally establish equivalence---an equivalence test (TOST) with
a pre-specified margin would be required for that stricter inference.
We therefore report the result as ``no evidence of localization to the
maintaining team'' rather than as ``equal effect.'' Skills authored
against the maintaining team's domain context produce no statistically
distinguishable developer-experience effect when consumed by engineers
in domains the authoring team did not directly serve; this is supportive
of, but does not prove, the generalization claim.

\paragraph{Tenure-with-plugin and perceived value.} A chi-squared test
of independence between plugin tenure (less than a month, one to three
months, three to six months, more than six months) and perceived
counterfactual impact (Q12) returned $\chi^2(3) = 0.36$, $p = 0.949$:
no association. A Spearman rank correlation between plugin tenure and
comparative experience (Q13) returned $r = 0.079$, $p = 0.652$: no
monotonic relationship. The plugin's perceived value is therefore
neither concentrated in early adopters experiencing novelty nor in
long-tenured users who have absorbed its conventions; new users and
experienced users report comparable benefit.

\paragraph{New-hire ramp-up.} A supplementary section of the survey
asked engineers under six months at Yahoo whether the plugin had
helped them ramp up faster. Four respondents qualified. All four
reported the plugin helped them ramp up faster (two ``significantly,''
two ``somewhat''); they collectively named four categories of task
they completed independently using the plugin that they would
otherwise have escalated to a teammate or manager---Yahoo-specific
tools, federated cloud authentication, first-day setup, and deploying
services. Three of the four shipped their first pull request within
two weeks of joining. We report this finding as a \emph{direction}
rather than a result: $n = 4$ is too small to support a statistical
claim, and the cohort is non-randomly self-selected. One respondent
captured the qualitative pattern directly: ``For someone who is new,
definitely it helps. There is a standardization built into the
plug-in, which definitely helps for folks who are new, folks who are
not aware of what or how things are done.''

Read together, the cross-unit, tenure, and new-hire findings constitute
weak-but-consistent evidence for the generalization claim. The null
results on business-unit and tenure stratifications would be hard to
explain if the value were driven primarily by maintaining-team
familiarity or by novelty effects; the new-hire qualitative direction
is consistent with the framework's prediction that codified
institutional knowledge should close the absorptive-capacity gap for
new engineers in the same way it closes the context-window gap for AI
agents. A more rigorous test---a structured before/after measurement
of new-hire time-to-first-PR with and without the deployment---is the
natural next study and is named in \S\ref{sec:case-study:validity}.

\subsection{Operational Learnings}
\label{sec:case-study:operational-learnings}

We report operational learnings from the deployment honestly. Some are
deployment-specific shortcomings that additional engineering investment
can resolve; some are organizational concerns requiring infrastructure
beyond the skill corpus itself; and some reveal gaps in the framework's
deployment model that the present deployment surfaced but did not
resolve.

\paragraph{Non-uniform validator coverage.} We did not implement
validators---the seventh component of the AKU schema specified in
\S\ref{sec:atomic-knowledge-units}---uniformly across all 87 skills.
Many procedural skills, particularly those whose action set is
interactive or whose outcomes are inherently hard to verify
deterministically, rely on the review-time quality gate and on
builder--hardener chaining rather than on per-skill post-action
validators. The governance gradient described in \S\ref{sec:governance}
therefore manifested heterogeneously across the corpus: a builder skill
chained to a fully validator-covered hardener operates at a higher
effective autonomy level than a builder skill whose chained hardener
relies only on procedural correctness. Expanding validator
coverage---particularly for the long tail of skills whose intent
admits deterministic post-checks---is the most direct lever for
advancing the corpus further along the governance gradient, and is
the principal deployment-level work item in progress.

\paragraph{Cross-plugin coordination at organizational scale.} The
framework's Activation Policy assumes a single canonical knowledge
corpus per agent. At Yahoo's organizational scale, multiple business
units now ship their own skill plugins targeting overlapping domains.
We observed concrete cases in which the agent's skill-discovery
heuristics selected a skill from a plugin authored by a different team
over the corresponding skill in our corpus, occasionally producing
incorrect guidance: in one such case, a build-system query routed to a
generic continuous-integration skill rather than the team-specific
build-system skill that would have produced the right answer for the
querying engineer's repository. At the single-corpus level the
Activation Policy works as the framework specifies; at the
multi-corpus organizational level the framework provides no
specification for inter-corpus routing, conflict resolution, or
canonical-authority assignment across overlapping domains. We name
this as a framework-level gap, not a deployment-level one: the
three-layer Agent Knowledge Architecture
(\S\ref{sec:golden-path-architecture}) assumes a single Registry,
Topology, and Activation Policy, and an organizational reality at
Yahoo's scale already exceeds that assumption.

\paragraph{Plugin staleness detection.} Engineers install the plugin
once and drift from the latest version until they notice a release
announcement and manually update. The deployment does not yet implement
an automatic check that compares the locally installed corpus commit
against the remote canonical version at session start. Users running
month-old corpus versions receive stale guidance, and the deployment
cannot directly measure the version distribution across the active
install base. This is an operational concern that matters
disproportionately: the institutional knowledge a skill encodes can
change faster than a user's update cadence---a security control may be
revised, a deprecated tool may be removed, a deployment pattern may
change---and a stale skill confidently delivers obsolete guidance to
an engineer who has no signal that the answer is no longer current.
The architecture should treat staleness detection as a first-class
operational concern; the present deployment does not yet.

\paragraph{Authoring access and contribution scaling.} The framework's
contribution model envisions a corpus that absorbs domain-specific
skills from across the organization, with team-scoped subdirectories
allowing each team to maintain its own knowledge contributions
alongside the shared corpus. In practice, the bulk of contributions
during the survey window came from the maintaining team. Engineers
outside that team who wished to contribute team-scoped skills or
correct gaps encountered access friction---source repository access,
code review permissions, and infrastructure entitlements---before they
could open a pull request. A dedicated access-and-permissions track is
in progress; once unblocked, team-scoped contribution should scale
naturally. The operational lesson is that the architecture's
contribution model assumes low-friction access to the corpus
repository, and an organization deploying the architecture must invest
in that access infrastructure as a precondition rather than an
afterthought.

\paragraph{Coverage gaps and the adoption ceiling.} At the time of the
survey the deployment had reached 238 active installs against a
denominator of roughly 2{,}000 engineers, an adoption rate of
approximately 12\%. Open-feedback responses from non-installers and
lapsed users identified three classes of adoption barrier.
\emph{Coverage gaps}: respondents working primarily on one of the two
major cloud providers in use reported the corpus was thin on that
provider's workflows; frontend and accessibility tooling was named as a
gap by multiple respondents; data-engineering and machine-learning
workflows are not yet covered at all. \emph{Discoverability}: several
respondents reported that even after installation it remained unclear
which skills were available, what each one did, and when to invoke
each. The interactive skill-graph view and the release-announcement
channel partially address this, but one lapsed user observed: ``I
might end up saving more time working directly with [the AI assistant]
rather than spending time figuring out which skill to use.''
\emph{Workflow inertia}: several non-active respondents reported that
they had simply continued with pre-existing AI workflows---``I have my
own workflow that I keep using despite installing it''---reflecting
that even when the corpus is installed, the cost of changing one's AI
usage habits is non-trivial. Each class admits a different
remediation, but the cumulative effect is an adoption ceiling that the
present deployment has not yet broken through.

\paragraph{Competing knowledge surfaces.} Multiple respondents reported
that the assistant occasionally preempted skill loading by reaching
for the organization's enterprise search platform before consulting
the skill corpus---behavior consistent with the framework's
characterization (\S\ref{sec:knowledge-activation}) of retrieval as a
competing but less precise approach to knowledge delivery. One
respondent noted that the agent ``sometimes defaults to [the
enterprise search] first even when knowledge is in engineering
skills,'' another that they had to ``remember to actually ask my
questions in [the assistant] instead of [their] habit of [search].''
The lesson is that an institutional-knowledge architecture must
contest the agent's existing knowledge-acquisition habits, not merely
add knowledge alongside them; the Activation Policy's priority over
retrieval needs to be enforced at the assistant layer, not assumed.

\paragraph{Synthesis.} These operational learnings cluster into three
categories. Some---non-uniform validator coverage, coverage gaps in
specific verticals, incomplete discoverability---are deployment-specific
shortcomings that additional engineering investment can resolve. Some---
staleness detection, authoring access, contribution scaling---are
organizational concerns requiring infrastructure beyond the corpus
itself, which any organization adopting the architecture must invest
in alongside the corpus. And some---cross-plugin coordination at
organizational scale, the contest with competing knowledge
surfaces---surface limitations of the framework's deployment model
that the present case study identifies but does not resolve. We treat
the third category as the most consequential for future framework
refinement, and discuss the corresponding research agenda in
\S\ref{sec:conclusion}.

\subsection{Threats to Validity}
\label{sec:case-study:validity}

The case study findings are subject to several threats to validity that
qualify the strength of the empirical claims and inform the design of
follow-on studies.

\paragraph{Self-report bias.} The four driver findings and the time-saved
estimate are based on respondent self-reports rather than direct
observation of agent task completion or wall-clock task duration.
Self-report introduces well-documented biases: respondents may anchor
their answers on recent salient experiences; respondents who consented
to a survey of plugin users may have stronger views than
non-respondents (see \emph{non-random response} below); and
self-estimated time savings have been shown in prior software-engineering
research to diverge from instrumented measurements. The drivers should
therefore be interpreted as measurements of \emph{perceived} developer
experience, not as instrumented productivity outcomes. We sketch the
study designs that would address this limitation at the end of this
subsection.

\paragraph{Concurrent confounders.} During the survey window, two
exogenous changes affected substantial fractions of the active-user
cohort. Approximately half (50\%) of active-user respondents received
an upgrade to a more capable foundation model in their AI assistant,
and roughly a third (32\%) reported recently beginning work in a new
codebase. Either could plausibly inflate the perceived benefit
attributed to the plugin. The driver findings remain statistically
significant after acknowledging the confounders, and the effect sizes
($d \geq 0.89$ on all four drivers, $d = 1.40$ on the strongest) exceed
what either confounder alone would plausibly produce, but the case
study cannot fully isolate the marginal effect of the deployment from
the marginal effects of the concurrent changes.

\paragraph{Cohort selection.} The respondent cohort is AI-fluent: 85\%
reported daily use of AI coding assistants prior to the plugin's
existence, and 98\% used the same primary AI assistant. The findings
therefore describe the experience of AI-native engineers comparing a
knowledge-architected agent against an architecturally-na\"ive one.
They do not describe the experience of engineers encountering AI
tooling for the first time, who would face a different and more
complex adjustment than the cohort the case study observed.
Generalization across populations with different AI fluency is an open
empirical question.

\paragraph{Sample size.} The active-user cohort is $n = 40$, which
yields adequate statistical power for the primary driver tests
(observed effect sizes drive narrow confidence intervals despite the
moderate sample) but produces wider confidence intervals for the Net
Promoter Score and the comparative-experience finding. The new-hire
ramp-up cohort is $n = 4$, which we have explicitly reported as a
\emph{direction} rather than a result. A larger replication, ideally
across multiple deployment cohorts, would tighten the intervals and
test the robustness of the cross-unit and tenure null findings.

\paragraph{Quasi-experimental design.} The case study is a
single-arm, post-deployment survey. It is not a randomized controlled
experiment; there is no control group of comparable engineers using
the same AI assistant without the deployment over the same window. The
cross-unit and tenure analyses provide quasi-experimental robustness
checks but do not substitute for randomized assignment. A controlled
before/after study, ideally with random assignment of the plugin to a
treatment cohort within a single business unit, would provide stronger
causal evidence.

\paragraph{Non-random response.} The survey was distributed through the
deployment's community channels and was voluntary. Engineers with
strong positive or negative views are likely overrepresented relative
to engineers with indifferent views. The lapsed-user and non-installer
responses ($n = 27$ combined) partially counteract this---we
explicitly sampled non-active users---but the response rate among
non-active engineers is unknown, and the published findings should be
read with awareness that the cohort is not a random sample of Yahoo
engineers.

\paragraph{Single-organization study.} The case study reports findings
from a single deployment at one organization. Cross-organization
replication is required to establish whether the framework's effects
generalize across companies with different institutional-knowledge
characteristics, different AI tooling environments, different
infrastructure stacks, and different organizational maturity in
codifying engineering standards. We do not claim that the magnitudes
reported here will replicate elsewhere; we claim that the architecture
admits a concrete enterprise realization that demonstrates the
predicted directional effects.

\paragraph{Author--deployment proximity.} The case study reports on a
deployment that the authors of this paper led. This is both a strength
(the architecture was deployed under conditions the framework
specifies) and a threat to validity (the survey instrument was
designed by the same team that built the deployment, introducing
potential question-framing biases). We attempted to neutralize this
through the survey design choices reported in
\S\ref{sec:case-study:methodology}: fully anonymous responses,
inclusion of barrier questions for lapsed users, open-text questions
for unprompted feedback, and a comparative question (Q13) that
explicitly invited critical responses. We cannot fully eliminate the
bias; we name it explicitly so independent replications can address it
directly.

\paragraph{Reuse frequency not analyzed.} The deployment's opt-in
telemetry pipeline logs skill activations per engineer per session and
therefore provides the raw data to compute reuse-frequency distributions
across the 87 skills. We did not analyze this data for the present case
study; an analysis of which skills are invoked most often, which are
rarely invoked, and how reuse correlates with authoring effort is a
natural follow-on study and would directly address the authoring
economics that the present case study otherwise underspecifies.

\paragraph{Knowledge density not directly measured.} The Context Window
Economy (\S\ref{sec:context-window-economy}) introduces knowledge density
$\rho(k,\tau)$ as the central economic construct of the framework. The
present case study does not compute $\rho$ directly: doing so requires
per-skill token counts paired with correction-rate or task-completion
deltas that the survey instrument did not capture. The findings on
perceived experience and time saved are downstream consequences of
high-density delivery rather than direct measurements of it. An
instrumented study with paired token counts and correction events at the
skill level is the natural follow-on and would convert the framework's
economic framing from a conceptual measure to an empirical metric.

\paragraph{Future studies.} Each of the threats above points to a
specific follow-on study design: a controlled before/after measurement
of new-hire time-to-first-PR with and without the deployment; a
multi-organization replication of the survey instrument; a paired
instrumented and self-reported study of time saved per task; and a
randomized assignment of the plugin within a single business unit to
isolate the marginal effect from concurrent foundation-model upgrades.
The present case study is best understood as an existence proof---
that the framework's architecture admits a concrete realization at
enterprise scale, and that the deployed realization produces effect
sizes consistent with the framework's predictions---which more
rigorous studies should build on rather than supersede.

\subsection{Summary}
\label{sec:case-study:summary}

We deployed the Knowledge Activation framework at Yahoo as a corpus of
87 modular, agent-consumable skills distributed through an internal
plugin marketplace and auto-imported into a second AI coding assistant
the organization supports. The deployment is a concrete instantiation
of the three-stage activation pipeline
(\S\ref{sec:knowledge-activation}), the seven-component AKU schema
(\S\ref{sec:atomic-knowledge-units}), and the three-layer Agent
Knowledge Architecture (\S\ref{sec:golden-path-architecture}). Skills
connect to one another through named continuations, are
self-discoverable through metadata triggers, and activate dynamically
on session context---making the corpus a knowledge graph rather than
a static documentation set. Golden paths emerge from the graph at
runtime, composed by the assistant from the topology rather than
authored as separate artifacts.

An anonymous survey of 67 Yahoo engineers, of whom 40 reported as
active users, returned statistically significant evidence for the
framework's central claims. All four perceived-experience drivers were
significantly above neutral with large or very large effect sizes
($d = 0.89$ to $d = 1.40$); 75\% of active users agreed the plugin
reduces the mental effort of remembering Yahoo conventions;
respondents reported a mean of 2.6 hours per week saved; 64\% would
meaningfully miss the plugin if it were turned off; the Net Promoter
Score was $+35$. Cross-unit Mann--Whitney tests returned no
significant difference between the maintaining business unit and the
rest of the organization on any driver, and tenure with the plugin
showed neither chi-squared nor Spearman association with perceived
value, suggesting that the framework's effects generalize beyond the
team that built it and beyond the early-adopter cohort. The 2.6-hour
weekly saving, read against the empirical literature showing that
active coding occupies a minority of the developer workweek (11--16\%
under fine-grained survey
instruments~\cite{houck2025timewarp, idc2024devtime, atlassian2025dxreport};
$\sim$5\% within IDE sessions under instrumented
observation~\cite{minelli2015summer}; and approximately 58\% of working
time devoted to comprehension rather than code production in
multi-application field studies~\cite{xia2018comprehension}), represents
a substantial share of the slice of the workweek where direct
engineering output already concentrates---while reflecting savings from
a corpus that covers only the infrastructure and platform-engineering
vertical of developer work. The framework's architecture, however, is not vertical-specific:
the same schema, graph, activation policy, and governance gradient
extend to other verticals whose codification investment has not yet
been made.

The case study has real limits, and we have reported them honestly.
The driver findings are perceived experience, not instrumented task
outcomes; concurrent confounders (a foundation-model upgrade reaching
half the cohort, new-codebase work for a third) cannot be fully
isolated; the active-user cohort is small and AI-fluent; the study is
single-organization, single-deployment, and survey-based; and two
findings---the cross-plugin coordination friction at organizational
scale and the contest with competing knowledge surfaces---identify
limitations in the framework's deployment model that the present
deployment did not resolve. None of these limits undermines the
existence proof the case study offers, but each specifies a direction
the framework's empirical evaluation must travel next.

The findings are best read in support of a structural rather than a
magnitude claim. The structural claim is that institutional-knowledge
architectures, organized around skills as the primitive unit and
composed into governance-aware knowledge graphs, admit a concrete
realization at enterprise scale, and that the realization produces
directional effects consistent with what the framework predicts:
reduced cognitive load on engineers, preserved flow against
context-switching, surfacing of requirements engineers did not know to
ask about, higher rates of first-try compliance with organizational
policy, and no significant difference detected across business units or tenure
strata in the present sample (an underpowered null comparison that is consistent
with the generalization claim but does not formally establish equivalence).
The magnitude claim---that the specific effect sizes observed in this
deployment will hold elsewhere, at different scales, in different
institutional environments, and against different baseline AI
tooling---requires the multi-organization, multi-cohort,
controlled-design studies that the threats-to-validity discussion in
\S\ref{sec:case-study:validity} sketches. The case study is the
beginning of the framework's empirical record, not its end.

\section{Discussion}
\label{sec:discussion}

\subsection{Synthesis of Contributions}

This paper has presented the Knowledge Activation framework as a formal bridge connecting four domains that have developed largely in isolation: knowledge management theory, platform engineering, autonomous AI agent design, and developer experience research. The framework's central claim---that the bottleneck to effective agentic software development is not model capability but knowledge architecture---rests on a chain of reasoning that proceeds from the economic constraints of context windows, through the transformation of latent institutional knowledge into structured, governance-aware primitives (skills), to the architectural and governance structures required for deployment at scale. The resulting formalism provides a vocabulary and structural foundation for reasoning about how organizations can systematically prepare their institutional knowledge for agent consumption---equipping agents to act with institutional accuracy and enabling the engineers who work with them to receive organizationally grounded guidance.

\subsection{Implications}

The Knowledge Activation framework carries implications for multiple communities of practice.

\paragraph{For platform engineering teams.}
Skills represent a new product surface for Internal Developer Platforms. Platform teams have traditionally produced APIs, CLIs, infrastructure templates, and developer portals as their primary outputs~\cite{beetz2023platform, cncf2024platforms}. The framework suggests that platforms must additionally produce AKUs: curated, governance-annotated, agent-executable knowledge units that encode organizational best practices. This shifts the platform team's mandate from building tools that developers use to authoring knowledge that agents execute. More specifically, the platform team's role evolves from authoring deterministic workflow templates---which are brittle and do not scale---to curating the skill library and knowledge topology from which agents compose AI-Generated Golden Paths on demand. The Agent Knowledge Architecture described in Section~\ref{sec:golden-path-architecture} provides the structural model for this expanded platform surface: an AKU Registry, Knowledge Topology, and Activation Policy that together form the agent-facing counterpart to the developer-facing portal.

\paragraph{For knowledge management.}
The framework represents a paradigm shift from document-centric to action-centric knowledge architecture. Traditional knowledge management systems organize knowledge as documents, wikis, and databases---artifacts optimized for human retrieval and interpretation~\cite{alavi2001, davenport1998}. Knowledge Activation reframes institutional knowledge as executable procedures paired with the metadata, tools, and governance constraints necessary for autonomous execution. This is not merely a difference in format; it reflects a fundamentally different epistemological stance. Knowledge, in this framework, is not something that is \emph{read} and then \emph{applied} by a separate cognitive process; it is something that is \emph{activated}---retrieved and executed as a unified operation. This perspective builds on Nonaka and Takeuchi's distinction between tacit and explicit knowledge~\cite{nonaka1995} but adds a third category: \emph{executable knowledge}, which is explicit in its representation and tacit in the organizational judgment it encodes.

\paragraph{For AI agent design.}
The framework argues that agents benefit substantially from curated, governance-aware knowledge rather than raw retrieval over unstructured corpora. As the comparison in Section~\ref{sec:knowledge-activation} details, enterprise knowledge is not merely factual but procedural, contextual, and governance-laden. Skills provide agents with pre-structured knowledge units that include not only what to do but how to do it, what tools to use, what permissions are required, and what governance constraints apply. This shifts a significant portion of the reasoning burden from the agent's runtime inference to the knowledge authoring process.

\paragraph{For enterprise governance.}
The framework proposes that governance shifts left---from runtime enforcement to knowledge-time specification. Rather than constructing elaborate runtime guardrails that attempt to constrain arbitrary agent behavior, organizations embed governance directly into the knowledge units agents consume (Section~\ref{sec:governance}). The validator mechanism transforms governance teams from approval bottlenecks to governance-as-code authors: by encoding standards as deterministic validators embedded in skills, governance scales with the skill library rather than with headcount---a structural resolution to the tension between agent autonomy and enterprise control.

\paragraph{For developer experience and productivity.}
The Knowledge Activation framework is, at its core, a developer experience
intervention. Noda et al.'s DevEx framework~\cite{noda2023devex} identifies cognitive
load as the highest-leverage dimension of developer experience; Xia et
al.~\cite{xia2018comprehension} documented that professional developers spend the majority of their time on program comprehension (Section~\ref{sec:rw-devex}). AKUs address this dimension
directly by externalizing institutional knowledge from working memory---whether the
developer's cognitive working memory or the agent's context window---into pre-structured,
injectable primitives. The SPACE framework's~\cite{forsgren2021space} efficiency
dimension captures the reduction in time spent on non-creative work; skills reduce the
time both agents and engineers spend reconstructing organizational context that has
already been codified elsewhere. The framework predicts measurable gains across multiple
DevEx dimensions: compressed onboarding time for new engineers (who currently require
three to six months to reach
productivity~\cite{ju2021onboarding, cortex2024productivity}), reduced cross-codebase
friction for engineers navigating unfamiliar services (where context-switching costs are
well-documented~\cite{mark2008cost}), and improved experience for senior engineers
whose institutional knowledge tax is structurally reduced. The Yahoo deployment reported
in Section~\ref{sec:case-study} provides initial empirical support for several of these
developer-experience predictions: an anonymous survey of active users established large
to very large effect sizes ($d = 0.89$ to $d = 1.40$) on perceived mental effort, flow
preservation, requirement surfacing, and first-try policy compliance, with no
significant difference detected across business units or tenure strata in the
present sample. Direct measurement of onboarding-time
compression and of the institutional-knowledge tax on senior engineers remains for
follow-on studies.

\paragraph{For enterprise technology strategy.}
The Knowledge Activation framework is inherently LLM-agnostic. Skills are knowledge-layer artifacts---structured text, deterministic scripts, organizational metadata---that function identically across any compliant model provider. The broad adoption of the Agent Skills open standard~\cite{anthropic2025agentskills} across multiple LLM providers demonstrates this portability in practice. Organizations that invest in codifying their institutional knowledge into skills create a durable, provider-independent asset: the knowledge persists and retains its value regardless of which LLM powers the agent. This positions Knowledge Activation as a strategic hedge against model provider lock-in---a concern of increasing salience as enterprises deepen their dependence on AI agent capabilities.

\paragraph{For enterprise AI adoption strategy.}
The framework offers a structural explanation for a pattern consistently documented in industry research: the lag of large enterprises in scaling AI from pilot to production~\cite{mckinsey2025stateofai, deloitte2026enterprise}. Industry analyses typically attribute this lag to data fragmentation, legacy infrastructure, and organizational inertia. Knowledge Activation proposes a complementary and underexamined factor: the \emph{institutional knowledge tax} identified in Section~\ref{sec:context-window-economy}, whereby senior engineers must manually supply organizational context to every agent interaction. Because large enterprises possess deeper and more fragmented institutional knowledge reservoirs---spanning more services, more compliance regimes, more teams, and more accumulated architectural decisions---the per-interaction cost of this manual knowledge transfer is structurally higher than in smaller organizations. The framework generates several testable predictions across all three classes of knowledge consumer. Organizations investing in
Knowledge Activation will observe: (1)~improved agent task completion rates as the
Institutional Impedance Mismatch is closed by pre-structured skills; (2)~compressed
onboarding time for new engineers, as institutional knowledge that currently requires
months of social absorption becomes available through the same skill library that serves
agents; (3)~reduced \emph{institutional} friction in cross-codebase operations---specifically, the overhead of acquiring deployment procedures, compliance requirements, and service-specific conventions when operating on unfamiliar codebases---as agents receive institutional context through the Knowledge Topology rather than through ad hoc human consultation (this prediction is scoped to the institutional knowledge dimension of cross-team friction; codebase familiarity, differing team norms, and social coordination costs are orthogonal factors that AKUs do not address); and (4)~improved developer experience and productivity among senior
engineers, as the institutional knowledge tax is structurally reduced by
codifying their expertise into reusable, community-maintained
skills. Greiler et al.'s developer experience framework~\cite{greiler2022developer} and
the SPACE framework~\cite{forsgren2021space} identify friction in daily workflows and
cognitive load as primary determinants of developer satisfaction and productivity. The
Knowledge Activation framework addresses both at their structural root.

\subsection{Limitations}

Several limitations of the present work must be acknowledged.

First, the Knowledge Activation framework's empirical record is at an early stage. Section~\ref{sec:case-study} reports a single-organization enterprise deployment with statistically significant supporting effects, but this single quasi-experimental case study---subject to the limitations enumerated in \S\ref{sec:case-study:validity}---does not yet establish the framework's magnitude claims across organizations, deployment contexts, and AI tooling environments. Claims regarding context window efficiency, governance effectiveness, and institutional knowledge transfer at the level of measured task outcomes (rather than perceived experience) remain to be tested through controlled experiments and multi-organization replications.

Second, skill authoring represents a \emph{significant organizational investment}. Converting tacit institutional knowledge into well-structured, governance-annotated Skills requires effort from domain experts, platform engineers, and governance specialists. The economics of this investment---whether the benefits of agent-executable knowledge justify the authoring costs---depend on factors including the frequency of skill reuse, the cost of agent errors in the absence of skills, and the availability of tooling to assist the authoring process. Relatedly, the claim that the institutional knowledge tax on senior engineers constitutes a significant factor in enterprise adoption lag (Section~\ref{sec:context-window-economy}) is theorized on the basis of convergent industry data but has not been empirically isolated through controlled study.

Third, the framework assumes the availability of \emph{capable tool-using agents}---language model-based systems that can reliably follow structured procedures, invoke tools, and respect governance constraints. While recent advances in agent architectures~\cite{yao2023react, schick2023toolformer, yang2024sweagent} demonstrate rapid progress, the reliability of current systems in high-stakes enterprise scenarios remains an active area of research.

Fourth, the framework's reliance on validators for automated governance assumes that governance requirements can be decomposed into deterministic checks. In practice, some governance decisions require human judgment that resists codification---ethical considerations, novel risk scenarios, and context-dependent trade-offs that defy rule-based expression. The governance gradient accommodates this through its hybrid mode, but the boundary between codifiable and non-codifiable governance remains an empirical question that warrants further investigation.

Fifth, the framework is scoped to \emph{intra-organizational} knowledge. Cross-organizational knowledge sharing---industry-wide best practices, open-source community conventions, or regulatory guidance that spans enterprises---is not addressed. Extending the skill primitive to inter-organizational contexts would require mechanisms for trust, provenance, and namespace management that remain unexplored.

Sixth, making institutional knowledge machine-readable and composable introduces a \emph{security surface} not present in human-oriented knowledge systems. Skills that encode deployment procedures, access patterns, or compliance workarounds become high-value targets if the skill registry is compromised. The governance constraints embedded in skills partially mitigate this risk, but a full threat model for skill-based knowledge architectures is beyond the scope of this paper.

Seventh, the challenge of \emph{knowledge staleness and maintenance at scale} is partially addressed by the knowledge commons model described below (Section~\ref{sec:discussion}), which provides a sociotechnical structure for community-maintained skills. However, the commons model itself requires organizational investment in contribution mechanisms, trusted committer structures, and incentive alignment that this paper theorizes but does not empirically validate. Technical staleness detection---automated mechanisms for identifying skills that have drifted from practice---remains an open engineering problem.

\subsection{The Exploration--Exploitation Tradeoff}

A deeper theoretical tension deserves attention. AKUs, by their nature, encode \emph{exploitation}: they capture known best practices and proven procedures, channeling agent behavior toward established organizational patterns. March's foundational analysis of exploration and exploitation in organizational learning~\cite{march1991exploration} warns that excessive exploitation at the expense of exploration leads to competency traps---organizations that become increasingly efficient at suboptimal practices.

However, the Knowledge Activation framework is inherently hybrid rather than purely exploitative. Three mechanisms provide exploration. First, \emph{graceful degradation}: when no AKU exists for a task, the agent reverts to baseline reasoning---general-purpose exploration unconstrained by codified procedures. The framework does not prevent exploration; it provides exploitation where AKUs exist and defaults to exploration where they do not. Second, the \emph{governance gradient} (Section~\ref{sec:governance}) allows agents greater autonomy for low-risk actions, creating space for exploratory behavior where the cost of failure is contained. Third, the \emph{knowledge commons} model provides a complementary channel: communities of practice organically generate new AKUs from operational experience, surfacing novel solutions that no centralized authoring process would anticipate---converting successful exploration into reusable exploitation.

The balance between these mechanisms remains an empirical question. Organizations with high AKU coverage may experience the competency trap March describes; organizations with sparse coverage may not derive sufficient exploitation benefit. The optimal coverage level---and the conditions under which the balance shifts---is an important direction for future work.

\subsection{The Skill Library as Knowledge Commons}

The Knowledge Activation framework's technical governance layer---validators, blast
radius declarations, approval workflows---ensures that skills \emph{execute}
correctly. But technical governance cannot prevent \emph{conceptual drift}: the skill
that runs successfully but no longer reflects how the organization actually works, because
practices have evolved since authoring. Vasilopoulos~\cite{vasilopoulos2026codified}
identifies specification staleness as the primary failure mode in sustained agent-assisted
development---a finding that underscores the insufficiency of purely technical
maintenance.

Sustainable skill maintenance at enterprise scale requires a complementary \emph{social
governance layer}: a community of practice~\cite{wenger1998communities} in which the
engineers who work with skills are also their stewards. Orlikowski~\cite{orlikowski2002knowing}
argues that organizational knowledge is not a static artifact but an ongoing social
accomplishment---continuously enacted through practice, not stored in documents. Skills
authored in isolation from practice will not encode the knowledge \emph{actually used};
skills actively maintained by the community of practitioners who use them will encode
knowing as it is actually performed, including the edge cases, informal conventions, and
failure modes discovered through operational experience.

The skill library is, in this framing, a \emph{knowledge
commons}~\cite{hessostrom2007knowledge}: a shared institutional resource that is
non-rivalrous (one agent's use does not deplete it) but subject to institutional decay
without community governance. Ostrom's design principle of \emph{collective choice
arrangements}---those affected by governance rules must be able to modify
them---implies that skill contribution cannot be restricted to the platform team alone. The
developer who discovers that a skill is wrong, outdated, or incomplete is the one best
positioned to fix it. InnerSource~\cite{capraro2018innersource} provides the operational
model: pull-request-based skill contributions, trusted committer review, with the platform
team governing the topology and activation policies rather than authoring every artifact.

\begin{figure}[htbp]
  \centering
  \includegraphics[width=0.85\textwidth]{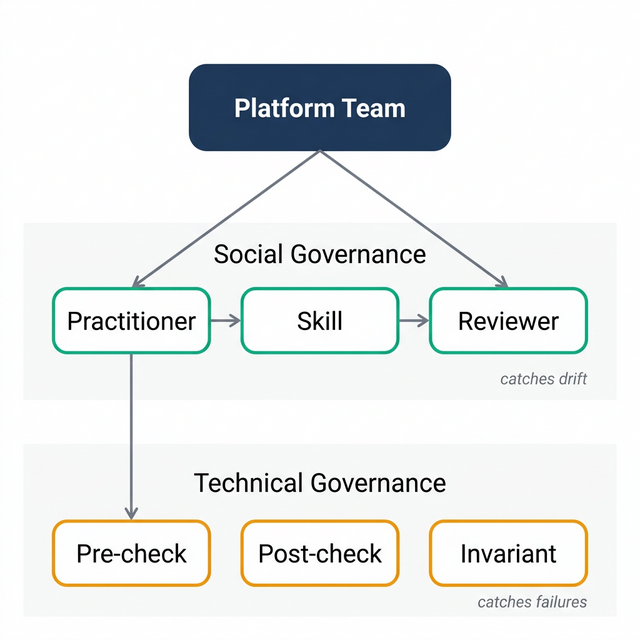}
  \caption{The knowledge commons governance model. Technical governance (bottom) uses validators to ensure skills execute correctly. Social governance (top) uses an InnerSource contribution model to ensure skills remain accurate as organizational practices evolve. The platform team stewards both layers, governing the topology and activation policies rather than authoring every skill. Technical governance catches execution failures; social governance catches conceptual drift---the case where a skill runs successfully but no longer reflects how the organization actually works.}
  \label{fig:knowledge-commons}
\end{figure}

This reframes the platform team's mandate. Rather than serving as the
\emph{authors} of all skills---a role that does not scale---the platform team becomes
the \emph{steward of the knowledge topology}: defining what kinds of skills exist, how
they compose, and what governance metadata they must carry, while the content of
individual skills is community-maintained. The institutional knowledge tax
(Section~\ref{sec:context-window-economy}) is thereby addressed at its source: when the
engineers who hold tacit knowledge are empowered to codify it, the tax is eliminated
structurally rather than absorbed into correction cascades. Figure~\ref{fig:knowledge-commons} illustrates this two-layer governance model.

\subsection{Graceful Degradation Under Partial Coverage}

A related practical concern is the behavior of the framework when skill coverage is incomplete---the realistic scenario for any organization beginning to invest in Knowledge Activation. No skill library will be comprehensive from the outset; agents will inevitably encounter tasks for which no matching skill exists. The framework must therefore degrade gracefully rather than fail entirely.

When no skill matches a task, the agent reverts to its baseline behavior: general-purpose reasoning augmented by whatever knowledge delivery mechanisms the organization already employs (retrieval-augmented generation, static system prompts, or unstructured documentation). The agent operates without the benefits of the Knowledge Activation framework---lower knowledge density, no embedded governance, higher risk of context rot---but it is not worse off than it would be in the absence of the framework entirely. The skill library provides value proportional to its coverage: each additional skill eliminates one category of guess-fail-correct-retry cycles, and the aggregate benefit grows incrementally as the library expands.

This incremental adoption model suggests a practical on-ramp for organizations. Rather than attempting to codify all institutional knowledge at once---a prohibitively expensive undertaking---organizations can begin by identifying the most frequent sources of agent correction. The patterns that cause the most costly context rot cycles are precisely the highest-value candidates for skill authoring. Over time, AI-assisted skill extraction from existing runbooks, postmortems, and correction logs can accelerate the codification process, reducing the authoring burden on the senior engineers whose knowledge the skills encode.

\subsection{An Adoption Path for Organizations}
\label{sec:adoption-path}

The incremental adoption model above admits a concrete sequencing that organizations can follow without an upfront codification of all institutional knowledge. We outline a four-phase path informed by the Yahoo deployment reported in Section~\ref{sec:case-study}.

\paragraph{Phase 0 (weeks 0--2): Identify the highest-correction patterns.}
Before authoring any skills, observe where AI agents most frequently produce organization-incorrect outputs. Candidates are the queries where parametric model knowledge consistently violates an organizational convention: federated cloud authentication, deployment hardening, security-scan remediation, and onboarding workflows are typical high-density targets in enterprise environments. A short telemetry pass against existing AI sessions, or a survey of frequent correction targets among senior engineers, is sufficient.

\paragraph{Phase 1 (weeks 2--12): Codify the top five to ten skills.}
Convert the highest-correction patterns into AKUs following the seven-component schema (Section~\ref{sec:atomic-knowledge-units}). Prioritize procedural correctness, source citation, and explicit continuations to downstream skills (typically a builder skill chaining to a security-hardening skill). Establish the multi-criterion review process and the canonical schema file (the AGENTS.md convention~\cite{liu2026agentsmd}) as repository-level invariants from the first merge.

\paragraph{Phase 2 (weeks 12--26): Build the topology and activation policy.}
As the corpus grows past ten skills, the value of the graph structure begins to dominate. Maintain a topology inventory that lets the agent compose multi-step workflows by traversing continuations. Configure the activation policy to trigger on the keyword and error-message patterns that route engineers to the correct skill at the moment of need. Invest in an interactive visualization of the graph so contributors can see where coverage is dense and where it is thin.

\paragraph{Phase 3 (weeks 26+): Open contribution and scale.}
Team-scoped contributions become the dominant growth channel beyond the initial authoring effort. Invest in the access infrastructure---repository permissions, code review rights, contributor onboarding---that the operational learnings reported in Section~\ref{sec:case-study:operational-learnings} identify as a critical bottleneck. Establish a structured gap-reporting loop so that engineer queries that fail to find a skill become input to the next authoring cycle.

The Yahoo deployment reached the early phases of contribution scaling over a 57-day initial window with a single maintaining team; the operational learnings discussion identifies several investments---staleness detection, cross-plugin coordination, contribution access at scale---that organizations following the same path will likely need to make at the Phase 3 boundary.

\section{Conclusion}
\label{sec:conclusion}

The bottleneck to effective agentic software development is not model capability but
knowledge architecture. This paper has argued that the Atomic Knowledge Unit---a structured,
governance-aware specialization of the AI Skill primitive---is the fundamental unit through which organizations
can bridge the gap between institutional knowledge and the knowledge consumers who need
it, whether autonomous AI agents or the human engineers who work alongside them. The
argument proceeds from a straightforward observation: as AI agents become capable
participants in software engineering workflows, the organizations that deploy them most
effectively will not be those with the most powerful models but those that have architected
their institutional knowledge for consumption at the point of need.

The paper makes four interconnected contributions. First, it formalizes the
\emph{problem}: the Context Window Economy
(Section~\ref{sec:context-window-economy}) establishes the resource constraints under
which knowledge must be delivered; the Institutional Impedance Mismatch names the
structural disconnect between parametric model knowledge and organizational institutional
knowledge; and the institutional knowledge tax identifies the sociotechnical cost---borne
disproportionately by senior engineers and by new engineers during onboarding---when this
mismatch goes unaddressed. Second, it specifies a \emph{knowledge architecture}:
Knowledge Activation (Section~\ref{sec:knowledge-activation}) defines the three-stage
pipeline (codification, compression, injection) that transforms latent organizational
knowledge into Atomic Knowledge Units---skills---whose seven-component schema
(Section~\ref{sec:atomic-knowledge-units}) bundles intent, procedure, tools, metadata,
governance, continuations, and validators into composable, interoperable primitives that form a
navigable knowledge graph. Third, it defines a \emph{deployment and governance model}:
the three-layer architecture of AKU Registry, Knowledge Topology, and Activation Policy
(Section~\ref{sec:golden-path-architecture}) provides the structural infrastructure for
organizing skills at scale; AI-Generated Golden Paths replace brittle deterministic
workflow templates with adaptive, agent-composed workflows; validators enable
governance-as-code; and the knowledge commons model
grounds sustainable skill maintenance in community practice rather than centralized
authorship. Fourth, it provides \emph{empirical validation}: Section~\ref{sec:case-study}
reports the deployment of the framework at Yahoo as a corpus of 87 modular
agent-consumable skills, with an anonymous survey of 67 engineers establishing
statistically significant developer-experience improvements consistent with the
framework's structural claims (large to very large effect sizes on all four
perceived-experience drivers, a mean of 2.6 hours per week saved per engineer, NPS
$+35$, and no significant difference detected across business units or tenure strata).

Together, these contributions bridge knowledge management theory, platform engineering,
autonomous AI agent design, and developer experience research---four domains that have
developed largely in isolation. The framework generates testable predictions:
organizations investing in knowledge architecture will observe improved agent task
completion rates, compressed onboarding time for new engineers, reduced institutional friction in cross-codebase operations, and measurable gains
in developer experience and productivity---particularly among the senior engineers whose
institutional knowledge tax the framework is designed to eliminate. The Yahoo deployment
reported in Section~\ref{sec:case-study} provides initial empirical support for several
of these predictions, while leaving direct measurement of onboarding-time compression
and cross-codebase friction reduction to controlled, multi-organization study.

Several directions for future work emerge naturally. The single-organization case study
reported in Section~\ref{sec:case-study} establishes initial empirical evidence for the
framework's structural claims; further empirical validation through multi-organization
replications, controlled before/after studies, and instrumented (rather than
self-reported) task-outcome measurements is the most immediate priority. Direct study of
new-hire onboarding time and senior-engineer institutional-knowledge tax under
controlled conditions remains particularly important. Measurement of developer experience outcomes---using frameworks such as
SPACE~\cite{forsgren2021space} and DevEx~\cite{noda2023devex}---in environments with and
without skill-based knowledge delivery would provide direct evidence for the
framework's predictions, particularly the hypothesized compression of onboarding time and
reduction of the institutional knowledge tax. Tooling for skill authoring, including
AI-assisted extraction of skills from existing runbooks, postmortems, and correction
logs, would reduce the organizational investment required for adoption. The extension of
the framework to multi-agent scenarios, in which agents negotiate skill selection across
organizational boundaries, presents rich theoretical and practical challenges. Automated
staleness detection---mechanisms for identifying skills that have drifted from
practice---would complement the knowledge commons model with technical
safeguards. Finally, the automatic generation and refinement of skills through agent
self-reflection~\cite{shinn2023reflexion} offers a path toward knowledge systems that
improve autonomously over time.

The integration of platform engineering, knowledge management, and developer experience is
not a speculative possibility; it is an architectural necessity driven by the rapid
integration of AI agents into enterprise software development. Organizations that
recognize AKUs as a first-class engineering artifact---and invest in the topologies,
governance structures, and community practices to support them---will establish the
institutional foundations for effective collaboration between human engineers and AI
agents. The Knowledge Activation framework provides the conceptual and structural
vocabulary for that effort.

\bibliographystyle{plainnat}
\bibliography{references}

\end{document}